\documentclass[journal]{IEEEtran}

\newcommand{\sss}[1]{{\scriptscriptstyle#1}}

\usepackage{times}
\usepackage{epsfig}
\usepackage{graphicx}
\usepackage{amsmath}
\usepackage{amssymb}
\usepackage{tikz,subfigure}
\usepackage{tabu}
\usepackage{txfonts}
\usepackage{multirow}

\usetikzlibrary{shapes,arrows,patterns,matrix,calc,positioning}
\usetikzlibrary{decorations.pathreplacing}
\tikzstyle{rect1} = [rectangle,draw, thick,fill=color1,minimum height=.55cm,minimum width=1.81cm]
\tikzstyle{rect2} = [rectangle,draw, thick,fill=color2,minimum  height=.55cm,minimum width=1.81cm]
\tikzstyle{rect3} = [rectangle,draw, thick,fill=color3,minimum  height=.55cm,minimum width=1.81cm]
\tikzstyle{rect4} = [rectangle,draw, thick,fill=color4,minimum  height=.55cm,minimum width=1.81cm]
\tikzstyle{rect5} = [rectangle,draw, thick,fill=color5,minimum  height=.55cm,minimum width=1.81cm]
\tikzstyle{rect6} = [rectangle,draw, thick,fill=color6,minimum  height=.55cm,minimum width=1.81cm]
\tikzstyle{rect7} = [rectangle,draw, thick,fill=color7,minimum  height=.55cm,minimum width=1.81cm]
\tikzstyle{rect8} = [rectangle,draw, thick,fill=color8,minimum  height=.55cm,minimum width=1.81cm]
\tikzstyle{rect9} = [rectangle,draw, thick,fill=color9,minimum  height=.55cm,minimum width=1.81cm]
\tikzstyle{rect10} = [rectangle,draw, thick,fill=color10,minimum  height=.55cm,minimum width=1.81cm]
\tikzstyle{rect11} = [rectangle,draw, thick,fill=color11,minimum  height=.55cm,minimum width=1.81cm]
\tikzstyle{rect12} = [rectangle,draw, thick,fill=color12,minimum  height=.55cm,minimum width=1.81cm]
\tikzstyle{rect13} = [rectangle,draw, thick,fill=color13,minimum  height=.55cm,minimum width=1.81cm]
\tikzstyle{rect14} = [rectangle,draw, thick,fill=color14,minimum  height=.55cm,minimum width=1.81cm]
\tikzstyle{rect21} = [rectangle,draw, thick,fill=color15,minimum  height=.55cm,minimum width=1.81cm]
\tikzstyle{rect16} = [rectangle,draw, thick,fill=color16,minimum  height=.55cm,minimum width=1.81cm]
\tikzstyle{rect17} = [rectangle,draw, thick,fill=color17,minimum  height=.55cm,minimum width=1.81cm]
\tikzstyle{rect18} = [rectangle,draw, thick,fill=color18,minimum  height=.55cm,minimum width=1.81cm]
\tikzstyle{rect19} = [rectangle,draw, thick,fill=color19,minimum  height=.55cm,minimum width=1.81cm]
\tikzstyle{rect20} = [rectangle,draw, thick,fill=color20,minimum  height=.55cm,minimum width=1.81cm]

\tikzstyle{rect15} = [rectangle,draw, thick,minimum  height=.55cm,minimum width=2.2cm]

\definecolor{Dbrown}{RGB}{127,0,0}
\definecolor{blue}{RGB}{0,0,255}
\definecolor{Dblue}{RGB}{0,0,127}
\definecolor{Dgreen}{RGB}{0,127,0}
\definecolor{brown}{RGB}{127,127,0}
\definecolor{red}{RGB}{255,0,0}
\definecolor{yellow}{RGB}{255,255,0}
\definecolor{green}{RGB}{0,255,0}
\definecolor{Lblue}{RGB}{0,255,255}
\definecolor{gray}{RGB}{127,127,127}
\definecolor{pink}{RGB}{255,0,255}
\definecolor{orange}{RGB}{255,127,0}
\definecolor{black}{RGB}{0,0,0}
\definecolor{lred}{RGB}{255,128,0}
\definecolor{white}{RGB}{255,255,255}

\definecolor{color1}{RGB}{131,153,153}
\definecolor{color2}{RGB}{254,127,0}
\definecolor{color3}{RGB}{200,50,50}
\definecolor{color4}{RGB}{127,254,127}
\definecolor{color5}{RGB}{254,0,0}
\definecolor{color6}{RGB}{80,80,80}
\definecolor{color7}{RGB}{0,127,127}
\definecolor{color8}{RGB}{254,127,127}
\definecolor{color9}{RGB}{254,254,254}
\definecolor{color10}{RGB}{254,254,127}
\definecolor{color11}{RGB}{254,127,254}
\definecolor{color12}{RGB}{254,254,0}
\definecolor{color13}{RGB}{127,254,254}
\definecolor{color14}{RGB}{127,0,254}
\definecolor{color15}{RGB}{40,40,40}
\definecolor{color16}{RGB}{183,143,143}
\definecolor{color17}{RGB}{138,43,150}
\definecolor{color18}{RGB}{85,107,47}
\definecolor{color19}{RGB}{199,21,133}
\definecolor{color20}{RGB}{123,104,238}

\definecolor{cole1}{RGB}{211,211,211}
\definecolor{cole2}{RGB}{255,0,0}
\definecolor{cole3}{RGB}{210,105,30}
\definecolor{cole4}{RGB}{205,92,92}
\definecolor{cole5}{RGB}{240,128,128}
\definecolor{cole6}{RGB}{0,238,118}
\definecolor{cole7}{RGB}{160,92,140}
\definecolor{cole8}{RGB}{154,205,50}
\definecolor{cole9}{RGB}{160,82,45}
\definecolor{cole10}{RGB}{34,139,87}
\definecolor{cole11}{RGB}{255,0,255}
\definecolor{cole12}{RGB}{138,43,226}
\definecolor{cole13}{RGB}{199,121,133}
\definecolor{cole14}{RGB}{255,255,0}
\definecolor{cole15}{RGB}{240,228,128}
\definecolor{cole16}{RGB}{160,92,240}
\definecolor{cole17}{RGB}{0,0,255}
\definecolor{cole18}{RGB}{135,206,235}
\definecolor{cole19}{RGB}{160,192,240}

\makeatletter
\newcommand{\thickhline}{%
	\noalign {\ifnum 0=`}\fi \hrule height 1pt
	\futurelet \reserved@a \@xhline
}

\newcommand*{\@rowstyle}{}
\newcommand*{\rowstyle}[1]{
	\gdef\@rowstyle{#1}%
	\leavevmode\@rowstyle
	\ignorespaces
}
\newcolumntype{=}{
	>{\gdef\@rowstyle{}\ignorespaces}%
}
\newcolumntype{+}{
	>{\leavevmode\@rowstyle\ignorespaces}%
}

\newcolumntype{C}[1]{>{\centering\arraybackslash}p{#1}}
\makeatother

\usepackage{pgfplots}
\usepackage{pgfplotstable}

\usepackage{helvet}
\usepackage[eulergreek]{sansmath}
\pgfplotsset{
	tick label style = {font=\sansmath\sffamily},
	every axis label = {font=\sansmath\sffamily},
	legend style = {font=\huge,font=\sansmath\sffamily},
}

\tikzstyle{line} = [draw, line width=.5pt, -latex]

\newcommand{\comment}[1]{}

\newcommand{\bx}{\mathbf{x}}

\newcommand{\by}{\mathbf{y}}

\newcommand{\bty}{\tilde{\mathbf{y}}}

\newcommand{\bB}{\mathbf{B}}

\newcommand{\bz}{\mathbf{z}}

\newcommand{\bA}{\mathbf{A}}

\newtheorem{claim-ap}{Claim}

\ifCLASSINFOpdf
\else
\fi

\begin{document}
	%
	\title{Soft Correspondences in Multimodal Scene Parsing}
	%
	%
	%
	
	\author{Sarah~Taghavi Namin, Mohammad~Najafi, Mathieu~Salzmann, and~Lars~Petersson
		\thanks{S. T. Namin, M. Najafi and L. Petersson are with CSIRO (DATA61), Canberra ACT 2601,
			Australia, and with the College of Engineering and Computer Science, 
			Australian National University, Acton ACT 2601, Australia. M. Salzmann is with Australian National University and CVLab, EPFL, Switzerland.
			E-mail: \{sarah.taghavi-namin, mohammad.najafi\}@anu.edu.au, lars.petersson@data61.csiro.au, mathieu.salzmann@epfl.ch.}}
	
	%
	%

	\markboth{}%
	{}
	%



	\maketitle
	
	\begin{abstract}
		Exploiting multiple modalities for semantic scene parsing has been shown to improve accuracy over the single-modality scenario. However multimodal datasets often suffer from problems such as data misalignment and label inconsistencies, where the existing methods assume that corresponding regions in two modalities must have identical labels. We propose to address this issue, by formulating multimodal semantic labeling as inference in a CRF and introducing latent nodes to explicitly model inconsistencies between two modalities. These latent nodes allow us not only to leverage information from both domains to improve their labeling, but also to cut the edges between inconsistent regions. We propose to learn intra-domain and inter-domain potential functions from training data to avoid hand-tuning of the model parameters. We evaluate our approach on two publicly available datasets containing 2D and 3D data. Thanks to our latent nodes and our learning strategy, our method outperforms the state-of-the-art in both cases. Moreover, in order to highlight the benefits of the geometric information and the potential of our method in simultaneous 2D/3D semantic and geometric inference, we performed simultaneous inference of semantic and geometric classes both in 2D and 3D that led to satisfactory improvements of the labeling results in both datasets.
	\end{abstract}
	

	%
	\IEEEpeerreviewmaketitle

	\section{Introduction}\label{sec:introduction}
	%
	%
	%
	%
	\IEEEPARstart {V}{arious} sensing modalities can be concurrently used to enhance the performance of scene understanding systems. For instance, high resolution 2D images provide useful textural information of the objects and 3D point cloud data reveal the 3D structure and size of the objects. 
	In the context of scene labeling, where the goal is to assign a class label to the elements of each modality, such as image pixels and 3D points, this has been shown to consistently yield increased accuracy over relying on a single domain~\cite{Posner:2008:city, zhang:2013:joint, Douillard:2009:LIDAR, Cadena:2014:icra,
		Munoz:2012:eccv, wacv2015}. 
	
\begin{figure}
		\centering
		
		\scalebox{.6}{
			\subfigure{
				\begin{tikzpicture}[node distance = 120cm, auto, semithick]
				
				\node [] (N1true) at (0,0) {\includegraphics[height=14cm,width=0.65\textwidth,clip]{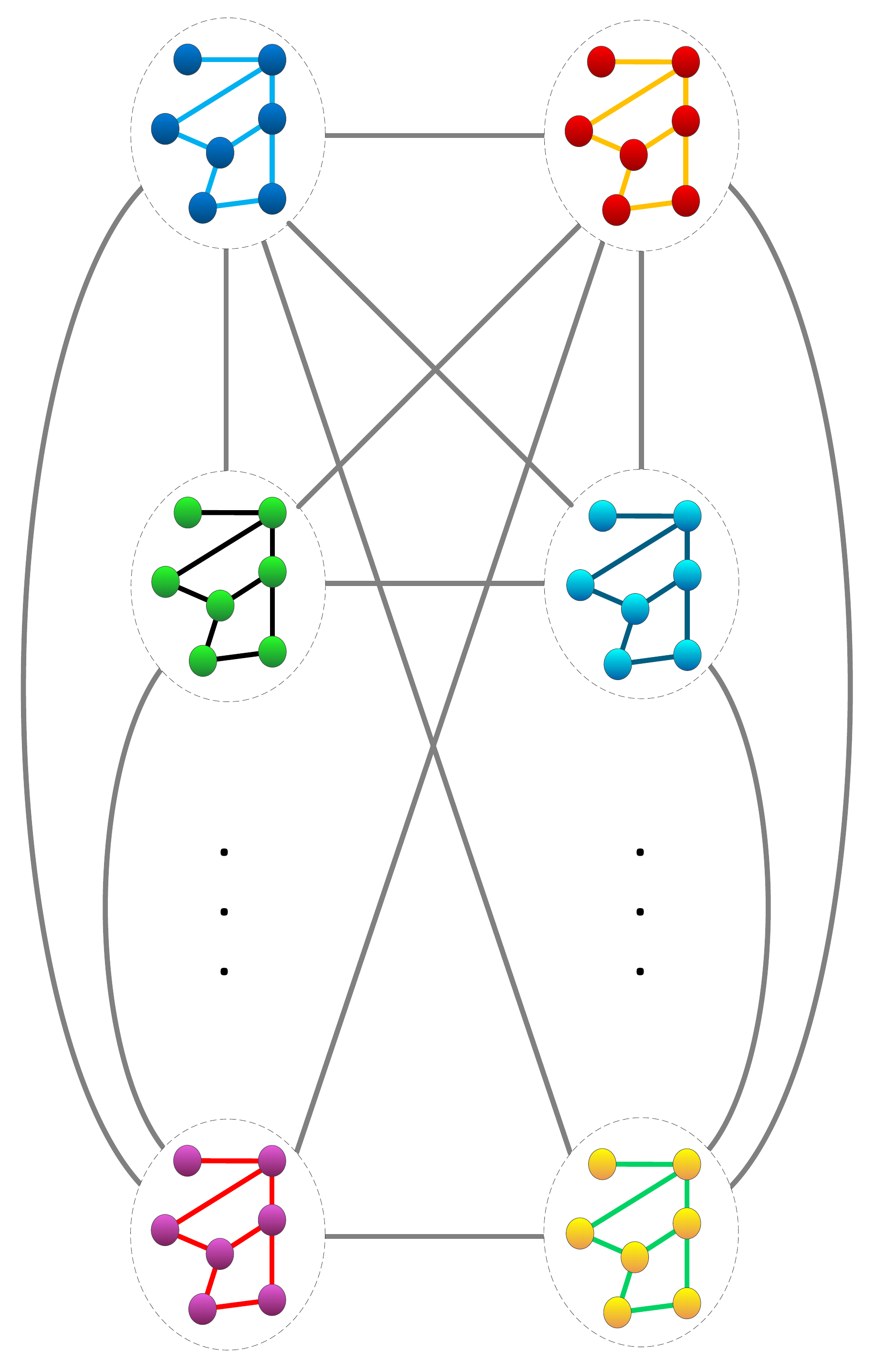}};
				
				\end{tikzpicture}
			}
		}
		\centering
		\caption{The proposed multimodal graphical model. The dots represent the nodes of more modalities, the intra-domain connections are represented by colored lines and the inter-domain connections are denoted by gray lines. The latent nodes exist between each inter-modality connection, though they have not been illustrated in this figure to avoid any confusion.}
		\label{multi_model}
	\end{figure}

	In this paper, we propose a multimodal model that can leverage the potential of various modalities simultaneously and the classification of each modality can be enhanced using the information of other sensing modalities (Figure~\ref{multi_model}).
	Taking into account multiple modalities that contain different types of information and cover different sorts of object categories is a challenging task, which will be addressed in this work. There are only a limited number of works done using multimodality sensing, where it has been generally assumed that the corresponding elements in various modalities must take identical class labels.
	This assumption is encoded either explicitly by having a single label variable for all modalities~\cite{Posner:2008:city, Douillard:2009:LIDAR, Cadena:2014:icra}, or implicitly by penalizing label differences between the domains~\cite{zhang:2013:joint, Munoz:2012:eccv, wacv2015}. This assumption, however, is very restricting if not infeasible, given different data modalities with their own specific properties and object categories. For example, Grass in 2D data may correspond to the class of horizontal plane in 3D data, or Sky which is a frequent class in 2D images of outdoor data can not be recorded using 3D data. In addition, the different modalities are typically not perfectly aligned/registered in practice. Furthermore, in dynamic scenes, moving objects may not easily be captured by some sensors, such as 3D Lidar, due to their lower acquisition speed. Note that a Lidar system captures 3D data continuously using a rotating sensor, unlike snapshot sensors where the image data are captured instantaneously. To give a concrete example, in the DATA61/2D3D dataset employed in our experiments, 17\% of the connections between the two modalities correspond to inconsistent labels. As a consequence, existing methods fail to model these inconsistencies and, hence, produce wrong labels in at least one modality.
	
\begin{figure}[tp]
		\centering
		
		\scalebox{.7}{
			\subfigure{
				\begin{tikzpicture}[node distance = 120cm, auto, semithick]
				
				\node [] (N1true) at (0,0) {\includegraphics[height=8cm,width=0.65\textwidth,clip]{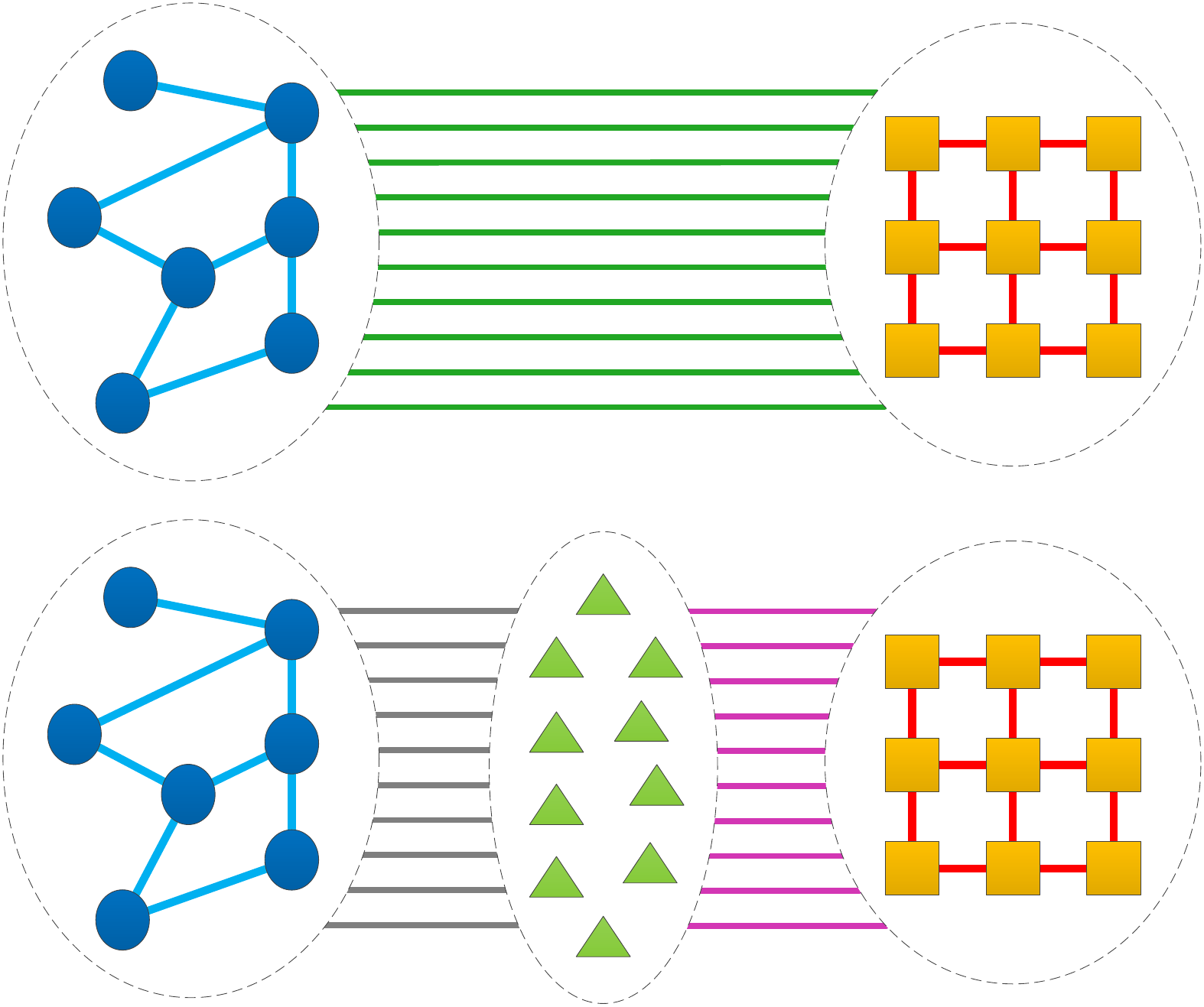}};
				
				\end{tikzpicture}
			}
		}
		\centering
		\caption{{\bf Top:} Existing approaches typically directly connect corresponding regions in different modalities and penalize these regions for taking different labels, thus producing wrong labeling in the presence of data misalignment, or other causes of label disagreement. {\bf Bottom:} Here, we introduce latent nodes that are placed between each connected pair of 2D and 3D nodes in the graph. They explicitly let us account for such inconsistencies, and potentially cut edges between the different domains. Circles denote the nodes in one domain (e.g., 3D) and squares denote the nodes in another domain (e.g., 2D). The latent nodes are depicted by triangles.}
		\label{model}
	\end{figure}

	Given the dissimilarities in the classes of different modalities and also the inherent misalignments between the domains, these modalities should be either studied separately, or connected such that each one of them could simultaneously utilize the incoming information of other modalities correctly.
	To this end, as shown in Figure~\ref{model}, we formulate multimodal scene parsing as inference in a Conditional Random Field (CRF), and introduce latent nodes to handle conflicting evidence between the different domains. The benefit of these latent nodes is twofold: First, they can leverage information from both domains to improve their respective labeling. Second, and maybe more importantly, these nodes allow us to cut the edges between regions in different modalities when the local evidence of the domains is inconsistent. As a result, our approach lets us correctly assign different labels to the modalities. In our formulation, different modalities can cover different sets of class labels and still leverage the information of other modalities to enhance the performance of the scene parsing system.
	
	More specifically, each connection between two domains is encoded by a latent node, which can take either a label from the same set as the regular nodes, or an additional label that explicitly represents a broken link. We then model the connections between the latent nodes and the different modalities with potential functions that allow us to handle inconsistencies. While many such connections exist, they come at little cost, because the only cases of interest are when the latent node and the regular node have the same label, and when the latent node indicates a broken edge. By contrast, having direct links between two modalities would require to consider potential functions for each combination of two labels (i.e., for $L$ labels, $L^2$ vs $2L$ in our model). The connections between the modalities that do not have identical label spaces are also governed by the latent nodes which have access to the features of both modalities. If these features match, the latent nodes then take the class labels that are consistent with the labels of the nodes at two ends of their respective connections. For example, the class Grass for a latent node is consistent with both horizontal plane in one modality and Grass in another one (Grass usually grows on horizontal surfaces). However, in case of a mismatch between the features of two modalities, the latent node breaks the link between them.
	
	Note that our method enables us to apply additional modalities with their own set of categories. To investigate this ability of our model, we use a 2D-3D dataset and take into account each modality twice using its corresponding \emph{geometric} and \emph{semantic} annotations and model their relationships. This in turn improves the performance of the system, with negligible impact on its run-time. Furthermore, we also model intra-domain connections with potential functions that encode some notion of label compatibility and thus let us model more accurately the relationships between different class labels. Altogether, these connections allow the information to be transferred across the domains, thus encoding the fact that some classes may be easier to recognize in one modality than in the others. Since such general potential functions cannot realistically be manually tuned, we propose to learn them from training data. To this end, we make use of the truncated tree-reweighted (TRW) learning algorithm of~\cite{Justin:2013:PAMI}. The resulting method therefore incorporates local evidence from each domain, intra-domain relationships and inter-domain compatibility via our latent nodes.
	
	We demonstrate the effectiveness of our approach on two publicly available 2D-3D scene analysis datasets: The DATA61/2D3D dataset~\cite{wacv2015} and the CMU/VMR dataset~\cite{Munoz:2012:eccv}. Our experiments evidence the benefits of the latent nodes and augmentation of the multiple modalities with their semantic and geometric annotations. It also indicates the advantage of learning the potentials for multimodal scene parsing. In particular, our approach outperforms the state-of-the-art on both datasets.

	\section{Related Work}
	Scene parsing has been an important and challenging problem in computer vision in the recent years. In particular, semantic labeling of 2D image data has been studied to a large extent, yielding increasingly accurate results~\cite{Singh,
		conf:iccv:XiaoQ09,Ladicky:2010:MCO:1888089.1888122, Girshick,Yang,ijcv:KohliLT09}. 
	With the advent of 3D depth sensors, such as laser range
	sensors (Lidar)~\cite{Hu, Triebel} and RGB-D cameras (e.g., Kinect)~\cite{Couprie, Silberman:ECCV12, Arbelaez,
		LBo:2011:iros}, it seems natural to leverage these additional sources of information to further increase the level of scene understanding~\cite{ NajafiECCV, koppula_semanticlabeling3d, Shapovalov2010, Lin_2013_ICCV}.
	
	
	In fact, more recently, several works have focussed on integrating 2D imagery and 3D point clouds for scene parsing \cite{Posner:2008:city, Douillard:2009:LIDAR, zhang:2013:joint, Cadena:2014:icra,Taghavi_iros2014,
		Munoz:2012:eccv, wacv2015}. In particular,~\cite{Posner:2008:city,Douillard:2009:LIDAR,Taghavi_iros2014} designed models based on variables corresponding to only one visual domain and then augmented them with visual cues extracted from the other modality. This approach, however, assumes that the same regions of the scene are observed in both domains, which is virtually never the case in practice. On the contrary, the model of~\cite{Cadena:2014:icra} incorporates variables for the two domains, but still relies on a single variable for the corresponding regions in both modalities. As a result, this approach still assumes that there is a perfect alignment between different visual domains. This, unfortunately, can typically not be achieved in practice, and the above-mentioned techniques will thus misclassify some regions in at least one of the domains.
	
	
	This assumption has been relaxed in some approaches by dedicating separate variables to the scene elements in the two modalities, even for matched regions. More specifically, ~\cite{Munoz:2012:eccv} came up with a hierarchical segmentation framework that performs parsing in two domains alternatively. However, since each modality transfers its labeling results to facilitate labeling in other  modality (depending on the overlap area of the 2D region and the projection of the 3D segment onto the 2D region), this method implicitly assumes that the regions that correspond with each other in two domains should take identical labels. In~\cite{zhang:2013:joint}, a framework to train a joint 2D-3D graph from unlabeled data was proposed. Similar to~\cite{Munoz:2012:eccv}, this method also propagates the labeling cues from one domain to the other thus implicitly assuming that corresponding nodes in 2D and 3D data should take the same labels. 
	Likewise, \cite{wacv2015} introduced a multimodal graphical model where each domain was represented by separate nodes. This approach, however, is designed based on Pott's model as pairwise potentials for both intra-domain and inter-domain edges. As a result, the assumption of assigning identical labels to the matched nodes in 2D and 3D domains is implicitly encoded.
	
	Here, by contrast, we propose to introduce latent nodes in a CRF to explicitly model the inconsistencies between two modalities. Furthermore, our approach lets us learn the intra-domain and inter-domain relationships from training data. Learning the parameters of CRFs for semantic labeling has been tackled by a number of works, such as~\cite{Weinman:2008,Philip:2013:icml} with mean-field inference,~\cite{Levin:2009:LCB:1487450.1487511} with TRW, and~\cite{journals/ijcv/RenFM08} with loopy belief propagation. Of more specific interest to us is the problem of learning label compatibility~\cite{Philip:2013:icml}, as studied by~\cite{Philip:2013:icml} for 2D images and by~\cite{Koppula:2011:indoor} for 3D data. Here, we consider label compatibility within and across domains. To the best of our knowledge, this is the first time such a learning approach is employed for multimodal scene parsing.
	
	There are other works that focus on semantic labeling and 3D reconstruction \cite{Hane:2013:cvpr, Ladicky:2012:ijcv, Eigen:2015:PDS}. However, none of these works deal with misalignment problem between natural 2D and 3D data. In particular, \cite{Eigen:2015:PDS} is formulated based on only a single modality (RGB image) as input, and \cite{Hane:2013:cvpr, Ladicky:2012:ijcv} reconstruct 3D data synthetically from stereo images in their framework. Zhang et al \cite{Zhang:2015:icra} also addressed the problem of multimodal 2D-3D semantic labeling by independently parsing the 2D and 3D data, and fusing their classification results. They however fail to account for misalignment issue, which is a challenging problem in natural multimodal datasets. 
	The closest work to this paper is \cite{shahin:2016:icpr}, where the authors addressed  domain mismatch problem by designing a specific cardinality loss function with an SSVM framework. However, the higher-order potentials in their model makes their approach computationally demanding, particularly when dealing with large-scale datasets. On the contrary, our graphical model is scalable and can be easily generalized to larger set of modalities and classes. Furthermore, unlike other graph-based approaches, the set of edges in our graph is flexible and can vary depending on how aligned the data modalities are in the problem.
	
	Xie et al~\cite{Xie:2016:CVPR} presented a multimodal dataset for outdoor scene understanding, though only 3D ground truth annotation information is provided with the dataset. The authors then used a dense 2D-3D graph to tansfer the 3D label information to all 2D pixels. 	
	Gould et al~\cite{Gould+al:ICCV09} integrated the semantic and geometric clues into their 2D scene understanding system and decomposed the scene into semantically and geometrically meaningful regions. Following~\cite{Gould+al:ICCV09}, Tighe and Lazebnik~\cite{ijcv/TigheL13} incorporated the geometric information into their region-wise scene parsing system ({\it Superparsing}) where they enforced coherence between the semantic labels ({\it building, car, person, etc.}) and geometric labels ({\it sky, ground, vertical surfaces}).
	
\begin{figure}[tp]
		\centering
		
		\scalebox{.7}{
			\subfigure{
				\begin{tikzpicture}[node distance = 120cm, auto, semithick]
				
				\node [] (N1true) at (0,0) {\includegraphics[height=12cm,width=0.65\textwidth,clip]{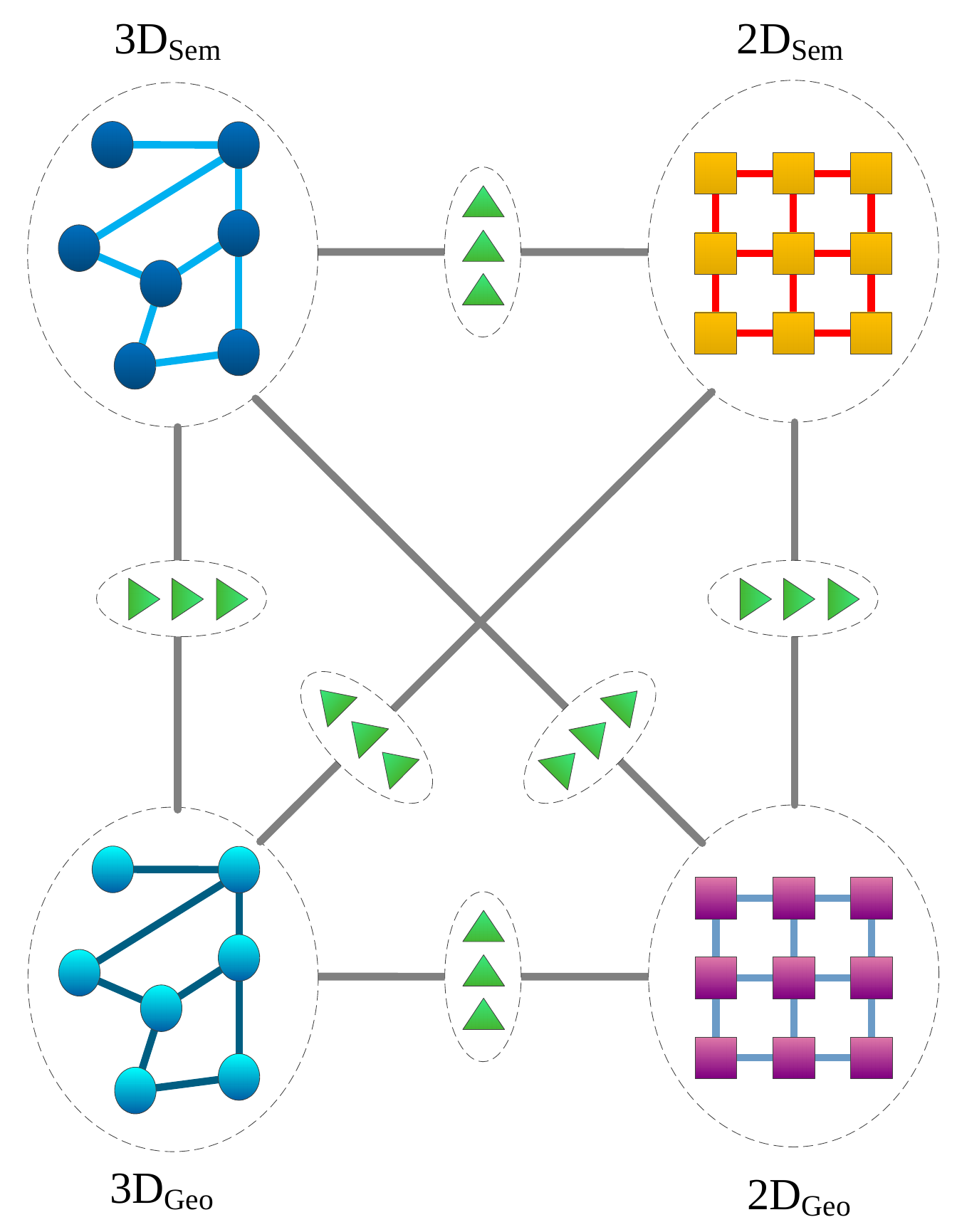}};
				
				\end{tikzpicture}
			}
		}
		\centering
		\caption{{\bf Top:} Our model which considers 2D semantic, 3D semantic, 2D geometric and 3D geometric nodes that are connected to each other via latent nodes. This model enables us to do inference on all the nodes using the semantic and geometric information simultaneously. Different colors represent different modalities. The latent nodes are represented by triangles.}
		\label{model_semgeo}
	\end{figure}

	Inspired by the above, we propose to use the semantic and geometric information of both 2D and 3D data simultaneously. To this end, we build our model upon different nodes which represent the semantic and geometric labels of each modality separately. These nodes are then linked together as seen in Figure~\ref{model_semgeo} for a simultaneous inference procedure. The evaluation results illustrate the superiority of this method over the previous work.
	
	\section{A General Multimodal CRF}
	\label{sec:method200}
	In this section, we present our multimodal graphical model.
	Let $\bx^{Mod_P} = \{\bx_{i}^{Mod_P}\}\;,\; 1 \leq i \leq N_m$, be the set of features extracted from the elements of the $p^{th}$ modality and $\by^{Mod_P} = \{y_i^{Mod_P}\}\;,\; 1 \leq i \leq N_m$, be the set of variables encoding the labels of the nodes in that modality, where each variable can take a label in the set $\mathcal{L} = \{1,\cdots,L\}$. Then the joint distribution of all modalities conditioned on the features can be expressed as 
	\begin{align}\label{eq:CRF_formula20}
	&\hspace{-1cm}P(\by^{Mod_1},\by^{Mod_2},...,\by^{Mod_P}|\bx^{Mod_1},\bx^{Mod_2},...,\bx^{Mod_P}) = \\  
	&\hspace{-1cm} \frac{1}{Z}\cdot \exp\Bigl(-\sum_{m=1}^P\Bigl(\sum_{i=1}^{N_m}\Phi_{i}^{Mod_m}+\sum_{(i,j)\in \mathcal{E}^{Mod_m}}\Psi_{ij}^{Mod_m}\nonumber \\
	&\hspace{-1cm}+\sum_{i=1}^{m-1}\sum_{(j,t)\in \mathcal{E}^{Mod_m,Mod_i}}\Psi_{jt}^{Mod_m-Mod_i}\Bigl)\Bigl), \nonumber
	\end{align}
	where $Z$ is the partition function, and
	$\Phi^{Mod_P}$ denotes the unary potentials of modality $p$. 
	$\Psi^{Mod_P}$ and $\Psi^{Mod_m-Mod_i}$ denote pairwise potentials defined over the set of edges $\mathcal{E}^{Mod_P}$ (intra-domain) and $\mathcal{E}^{Mod_m-Mod_i}$ (inter-domain), respectively. 
	The potential functions in Equation~\ref{eq:CRF_formula20} are built such that they could intuitively model the correlation between the class probabilities and local information of each node, as well as the contextual relationships between the pairs of adjacent nodes in the graph (pairwise potentials).
	
	In~\cite{wacv2015}, handcrafted potentials were used for the multimodal graphical model, where the pairwise potential function is defined in a way that penalizes dissimilar class labels for two adjacent regions if their feature vectors are very similar. The contributions of the handcrafted potentials in the inference process are determined via a set of weighting parameters. These parameters are then adjusted through a validation step, so as to produce the lowest error on the validation data. 
	
	A drawback of the handcrafted potentials that are based on a Pott's model is that they do not convey any information on the compatibility of different objects and class labels. As an example, take the scenario where a superpixel in the 2D domain is classified as \emph{Grass} and it has connections with two different 3D segments, one labeled as a flat object, e.g., \emph{Road} or \emph{Grass}, and the other one predicted to be a cylindrical object such as \emph{Powerpole}. Assigning the same weight to these pairwise links, even if they have the same amount of 2D-3D overlap, might not be a right decision because, in the former case, the predicted 2D class is compatible with the predicted class in the 3D domain. However, in the latter, the difference in shape of the predicted classes demands a more tuned and class-specific pairwise weight. This problem can be addressed by considering different weights for different class combinations of the nodes in a pairwise edge, e.g., \emph{2D:Grass-3D:Grass}, \emph{2D:Grass-3D:Road}, or \emph{2D:Grass-3D:Tree Trunk}. Therefore, we assign a set of label compatibility parameters for all possible class combinations and learn them from data.
	
	Moreover, assigning a fixed set of weights to the unary potentials of different modalities overlooks the fact that some of the classes are recognized better using one data modality and some other object classes can be described more precisely using the another modality. For instance, when it is deduced from the 3D data that the object of interest has a flat shape, the labeling algorithm should trust the 3D information more to put the object in one of the flat categories. If, in this case, the 2D data describes the object as a green entity, e.g., \emph{Grass}, \emph{Bushes}, \emph{Tree top}, the classifier should ideally pick \emph{Grass} as class label.
	
	Our goal is to construct and train our graphical model based on a set of potentials that describe: {\bf I)} the reliability of the local information of each domain per class, and {\bf II)} the cost of various intra-domain and inter-domain class neighborhoods (a.k.a. the label compatibility). To obtain a labeling, we perform inference in our CRF by making use of the truncated TRW algorithm of~\cite{Justin:2013:PAMI}.
	\subsection{Potential Definition}\label{ch:potential_definition}
	The CRF formulation in Equation~\ref{eq:CRF_formula20} includes several unary and pairwise potentials that are defined here.
	The unary potential of a node is generally computed via its local information and indicates the cost of assigning a class label to the node.
	We define the cost of assigning label $l$ to the corresponding variables as
	
	\begin{equation}\label{eq:learned_ModP_potentials}
	\Phi_{i}^{Mod_P}(y_{i}^{Mod_P} = l) = \mathbf{A}_l^{Mod_P} \bx_{i}^{Mod_P}\;,
	\end{equation}
	where $\bA^{Mod_P} \in \mathbb{R}^{L\times D_{Mod_P}}$ is the parameter matrix for the unary potential in modality $p$, with $\bA_l^{Mod_P}$ the row of $\bA^{Mod_P}$ corresponding to label $l$. Since they directly act on the local features $\bx_{i}^{Mod_P}$, this matrix encodes how much each feature dimension should be relied on to predict a specific label. Note that $D_{Mod_P}$  refers to the dimension of the feature vector in modality $p$.
	
	Pairwise potentials express the cost of all possible joint label assignments for two adjacent nodes in the graph. The handcrafted potentials are limited to simply encouraging the nodes to share the same labels. By contrast, here, we define general pairwise potentials that let us encode sophisticated label compatibilities. For the intra-domain edges, these potentials are defined as
	
	\begin{equation}\label{eq:learned_3d_pairwise}
	\Psi_{jk}^{Mod_P}(y_{j}^{Mod_P}=l,y_{k}^{Mod_P}=s) = \mathbf{B}_{ls}^{Mod_P} \mathbf{v}_{jk}^{Mod_P}\;,
	\end{equation}
	where $\mathbf{B}^{Mod_P}$ is a parameter matrix with $L^2$ rows representing all possible combinations of two labels, and $\mathbf{B}_{ls}^{Mod_P}$ is the row of $\mathbf{B}^{Mod_P}$ corresponding to the combination of label $l$ with label $s$. In this case, we set the edge features $\mathbf{v}_{jk}^{Mod_P}$ to be the $\ell_2$-norm of the difference of a subset of the original node features $x_{j}$ and $x_{k}$, which will be discussed in Section~\ref{F&P}.
	
	Similarly, the inter-domain pairwise potential between modality $i$ and modality $m$ is defined as
	
	\begin{equation}\label{eq:learned_Mod_i-Mod_m_pairwise}
	\Psi_{jt}^{Mod_i-Mod_m}(y_{j}^{Mod_i}=l,y_{t}^{Mod_m}=s) = \mathbf{B}_{ls}^{Mod_i-Mod_m} \mathbf{v}_{jt}^{Mod_i-Mod_m}\;,
	\end{equation}
	where $\mathbf{v}_{jt}^{Mod_i-Mod_m}$ is the concatenation of a subset of the original node features in ${Mod_i}$ and ${Mod_m}$.
	
	\section{General Multimodal CRF with Latent Nodes}
	\label{sec:method2}
	We now address the problem of inconsistencies across the modalities by introducing latent nodes to our model. The latent nodes are placed between the pairs of corresponding nodes in two modalities. This breaks down the between-modality edges into two edges that link the node in ${Mod_i}$ and the latent node, and also the node in ${Mod_m}$ and the latent node. In other words, no edge directly connects ${Mod_i}$ to ${Mod_m}$. Our latent nodes can either take a label from the same space as the label space of the ${Mod_i}$ or the ${Mod_m}$ nodes\footnote{When ${Mod_i}$ and ${Mod_m}$ have different label spaces, the latent node can take a label from one of them.}, or another label indicating that the link between the two modalities should be cut.

	Formally, let $\by^{Mod_P} = \{y_i^{Mod_P}\}\;,\; 1 \leq i \leq N_m$ be the set of variables encoding the node label in modality $p$. Each of these variables can take a label in the set $\mathcal{L} = \{1,\cdots,L\}$. Furthermore, let $T_{m,i}$ be the number of pairs of corresponding nodes in modality $m$ and modality $i$, found in the manner described in Section~\ref{F&P}. We then denote by $\by^{\Delta} = \{y_t^{\Delta_{Mod_i,Mod_m}}\}\;,\; 1 \leq t \leq T_{m,i}$ the latent nodes associated with these correspondences. These variables can be assigned a label from the space $\mathcal{L}' = \{0, 1, \cdots, L\}$, where label $0$ represents a broken link, which means the nodes do not influence each other.
	
	Given $\bx^{Mod_P} = \{\bx_{i}^{Mod_P}\}$ as the features extracted from the elements in modality $p$, the joint probability distribution of all data nodes and latent nodes conditioned on the features can be expressed as
	\begin{align}\label{eq:CRF_Gformula_latent}
	P(\by^{Mod_1},\by^{Mod_2},...,\by^{Mod_P},\by^{\Delta_{Mod_1,Mod_2}},\by^{\Delta_{Mod_1,Mod_3}},\by^{\Delta_{Mod_2,Mod_3}},...,\\&\hspace{-7.5cm}\by^{\Delta_{Mod_{P-1},Mod_P}}|\bx^{Mod_1},\bx^{Mod_2},...,\bx^{Mod_P}) =  \frac{1}{Z}\cdot \nonumber \\
	&\hspace{-7.5cm}\exp\Bigl(-\sum_{m=1}^P\Bigl(\sum_{i=1}^{N_m}\Phi_{i}^{Mod_m}+\sum_{(i,j)\in \mathcal{E}^{Mod_m}}\Psi_{ij}^{Mod_m}\nonumber \\
	&\hspace{-7.5cm}+\sum_{i=1}^{m-1}\Bigl(\sum_{t=1}^{T_{m,i}}\Phi_t^{\Delta_{Mod_m,Mod_i}}+\sum_{(j,t)\in \mathcal{E}^{Mod_m,\Delta_{Mod_m,Mod_i}}}\Psi_{jt}^{Mod_m-\Delta_{Mod_m,Mod_i}}\nonumber \\&\hspace{-7.5cm}+\sum_{(j,t)\in \mathcal{E}^{Mod_i,\Delta_{Mod_m,Mod_i}}}\Psi_{jt}^{Mod_i-\Delta_{Mod_m,Mod_i}}\Bigl)\Bigl)\Bigl), \nonumber
	\end{align}
	Where $\Phi^{\Delta_{Mod_m,Mod_i}}$ denotes the unary potential of the latent nodes and
	$\Psi^{Mod_m-\Delta_{Mod_m,Mod_i}}$ denotes the pairwise potentials defined over the set of edges $\mathcal{E}^{Mod_m-\Delta_{Mod_m,Mod_i}}$. To obtain a labeling, as in Section~\ref{sec:method200} we use the TRW method to perform inference in our CRF. In the remainder of this section, the latent potentials in Equation~\ref{eq:CRF_Gformula_latent} are described.
	
	\subsection{Unary Potentials of Latent Nodes}\mbox{}\\
	Similar to data modality nodes, the unary potential for the latent nodes is defined as
	\begin{equation}\label{eq:learned_2d_latent_unary}
	\Phi_t^{\Delta_{Mod_m,Mod_i}}(y_t^{\Delta_{Mod_m,Mod_i}} = l) = \mathbf{A}_l^{\Delta_{Mod_m,Mod_i}} \bx_{t}^{\Delta_{Mod_m,Mod_i}}\;,
	\end{equation}
	where $\mathbf{A}^{\Delta_{Mod_m,Mod_i}}$ is, again, a parameter matrix, which this time contains $L+1$ rows to represent the fact that a latent node can take an additional label to cut the connection between two modalities.
	The feature vector of a latent node is constructed by concatenating the features of the corresponding ${Mod_m}$ and ${Mod_i}$  nodes, i.e., $\bx_{t}^{\sss{\Delta_{Mod_m,Mod_i}}}=[(\bx_{j}^{Mod_m})^T,(\bx_{k}^{Mod_i})^T]^T$. Having access to both ${Mod_m}$ and ${Mod_i}$ features allows this unary to detect mismatches in the ${Mod_m}$ and ${Mod_i}$ observations, and in that event, favor cutting the corresponding edge.
	
	\subsection{Inter-domain Pairwise Potentials with Latent Nodes}\mbox{}\\
	The inter-domain pairwise potentials associated with the latent nodes
	that connect two modalities are defined as
	\begin{align}\label{eq:learned_modm_latent_pairwise}
	\Psi_{jt}^{Mod_m-\Delta_{Mod_m,Mod_i}}(y_{j}^{Mod_m}=l,y_{t}^{\Delta_{Mod_m,Mod_i}}=s) =\\&\hspace{-5cm} \mathbf{B}_{ls}^{Mod_m-\Delta_{Mod_m,Mod_i}} \mathbf{v}_{jt}^{Mod_m-\Delta_{Mod_m,Mod_i}}\;, \nonumber
	\end{align}
	and
	\begin{align}\label{eq:learned_modi_latent_pairwise}
	\Psi_{kt}^{Mod_i-\Delta_{Mod_m,Mod_i}}(y_{k}^{Mod_i}=l,y_{t}^{\Delta_{Mod_m,Mod_i}}=s) =\\&\hspace{-5cm} \mathbf{B}_{ls}^{Mod_i-\Delta_{Mod_m,Mod_i}} \mathbf{v}_{kt}^{Mod_i-\Delta_{Mod_m,Mod_i}}\;, \nonumber
	\end{align}
	where the parameter matrices now have $L\times(L+1)$ rows to account for the extra label of the latent nodes. In practice, we set $\mathbf{v}_{jt}^{Mod_m-\Delta_{Mod_m,Mod_i}}$ and $\mathbf{v}_{kt}^{Mod_i-\Delta_{Mod_m,Mod_i}}$ to 1, thus resulting in $L \times (L+1)$ parameters. Note, however, that the effective number of parameters corresponding to these potentials is much smaller. The reason is that the only cases of interest are when the latent node and the regular node take the same label, and when the latent node indicates a broken link. The cost of the other label combinations should be heavily penalized since they never occur in practice. This therefore truly results in $2L$ parameters for each of these potentials.
	
	\section{Training our Multimodal Latent CRF}\label{ch:training}
	Our multimodal CRF contains many parameters, which thus cannot be tuned manually. Here, we propose to learn these parameters from training data. To this end, we make use of the direct loss minimization method of~\cite{Justin:2013:PAMI}.
	
	More specifically, let $\{\bz_i\}$, $1 \leq i \leq N$ be a set of $N$ labeled training examples, such that $\bz_i = \big(\bx_i^{Mod_1}, ..., \bx_i^{Mod_P}, \bty_i^{Mod_1}, ..., \bty_i^{Mod_P}, \bty_i^{\Delta_{Mod_1-Mod_2}}, ..., \bty_i^{\Delta_{Mod_{P-1}-Mod_P}}\big)$, where, with a slight abuse of notation compared to Section~\ref{sec:method200} and Section~\ref{sec:method2}, $\bx_i^{Mod_P}$, resp. $\bty_i^{Mod_P}$, englobes the features, resp. ground-truth labels, of all the nodes in the $i^{th}$ training sample for modality ${P}$, and similarly for the other terms in $\bz_i$. In practice, to obtain the ground-truth labels of the latent nodes $\bty_i^{\Delta_{Mod_i-Mod_m}}$, we simply check if the ground-truth labels of the corresponding ${Mod_i}$ and ${Mod_m}$ nodes agree, and set the label of the latent node to the same label if they do, and to 0 otherwise\footnote{Note that our nodes are latent in the sense that they do not correspond to physical entities, not in the sense that we do not have access to their ground-truth during training.}.
	
	Learning the parameters of our model is then achieved by minimizing the empirical risk 
	\begin{equation}\label{eq:risk}
	r({\Theta}) = \sum_{i=1}^N l(\Theta,\bz_i)
	\end{equation}
	w.r.t. $\Theta=\{\bA^{Mod_1}, ..., \bA^{Mod_P}, \bA^{\Delta_{Mod_{1},Mod_2}}, ..., \bA^{\Delta_{Mod_{P-1},Mod_P}},\bB^{Mod_1}, ..., \\ \bB^{Mod_P},\bB^{Mod_1-\Delta_{Mod_{1},Mod_2}}, ..., \bB^{Mod_P-\Delta_{Mod_{P-1},Mod_P}}\}$,
	where $l(\Theta,\bz_i)$ is a loss function. 
	
	Here, we use a marginal-based loss function, which measures how well the marginals obtained via inference in the model match the
	ground-truth labels.  In particular, we rely on a loss function defined on the clique marginals~\cite{Wainwright:2008:GME}. This can be expressed as $l(\Theta,\bz_i) = -\sum_c \log \mu(\bz_{i,c};\Theta)$ 
	where $c$ sums over all the cliques in the CRF, i.e., all the inter-domain and intra-domain pairwise cliques in our case, $\bz_{i,c}$ denotes the variables of $\bz_i$ involved in a particular clique $c$, and $\mu(\bz_{i,c};\Theta)$ indicates the marginals of clique $c$ obtained by performing inference with parameters $\Theta$.
	
	We use the publicly available implementation of~\cite{Justin:2013:PAMI} with truncated TRW as inference method. This method was shown to converge to stable parameters in only a few iterations. In practice, we run a maximum of 5 iterations of this algorithm. 
	
	\section{Special Cases}
	\label{sec:method007}
	In this section, we demonstrate how our general multimodal model can be used for modeling two special cases of {\bf I)} 2D-3D multimodal data, and {\bf II)} 2D-3D semantic and geometric multimodal data, both accompanied with latent nodes.
	\subsection{2D-3D CRF with Latent Nodes}
	\label{sec:method}
	Since 2D imagery and 3D data are often the most popular modalities used for semantic labeling, here we focus the discussion on these two visual domains. Nevertheless, our approach generalizes to other modalities, such as infrared or hyper-spectral data.
	
\begin{figure}[t]
		\centering

		\subfigure{
			\begin{tikzpicture}[node distance = 120cm, auto, semithick]
			
			\node [scale=1.1] (N1true) at (0,2.3) {\includegraphics[height=2.7cm,width=0.21\textwidth,clip]{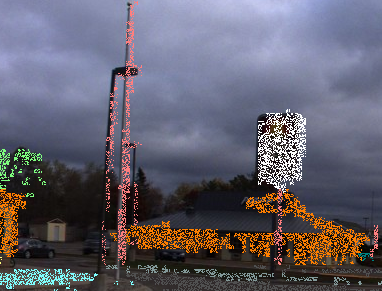}};
			\node [scale=1.1] (N2true) at (4.2,2.3) {\includegraphics[height=2.7cm,width=0.21\textwidth,clip]{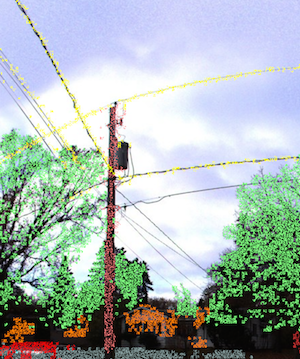}};
			\node [scale=1.2] (N1true) at (0,-.1) {\includegraphics[height=1cm,width=0.18\textwidth,clip]{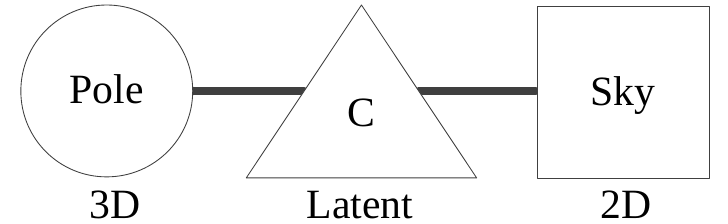}};
			\node [scale=1.2] (N2true) at (4.2,-.1) {\includegraphics[height=1cm,width=0.18\textwidth,clip]{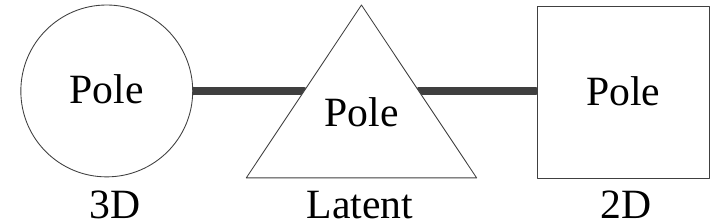}};

			\end{tikzpicture}
		}
		\centering
		\caption{{\bf Latent nodes for data misalignment.} {\bf Left:} The projection of
			\emph{pole} from 3D to 2D covers some regions of \emph{sky}, which creates a connection between the corresponding 3D and 2D nodes. Having access to both 3D and 2D features, the latent node should
			detect the mismatch and cut this connection thus allowing the nodes to take different labels. {\bf Right:} In this example, we have an accurate projection . As a result, the features of the 2D and 3D nodes are both congruent with category label {\it pole}. Hence, the latent node between them preserves an active edge between the nodes and predicts the same label.}
		\label{misalign}
	\end{figure}

\begin{figure*}[tp]
		\centering
		\def\dd{3.5}
		\def\ddy{-1.6}
		
		\scalebox{1}{
			\subfigure{
				\begin{tikzpicture}
				
				\node [scale=.8] (p1) at (0,0) [rect5] {Vehicle};
				\node [scale=.8] (p2) at (0,1) [rect6] {Road};
				
				\node [scale=1.1] (N1true) at (0+\dd,2.4+\ddy) {\includegraphics[height=2.9cm,width=0.25\textwidth,clip]{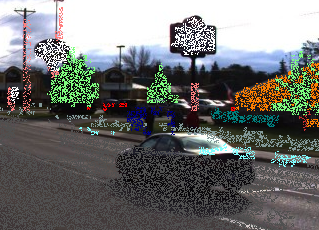}};
				\node [scale=1.1] (N2true) at (5+\dd,2.4+\ddy) {\includegraphics[height=2.9cm,width=0.25\textwidth,clip]{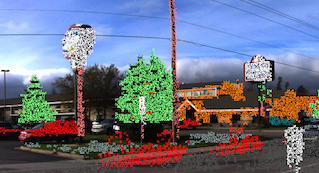}};
				\node [scale=1.1] (N3true) at (10+\dd,2.4+\ddy) {\includegraphics[height=2.9cm,width=0.25\textwidth,,clip]{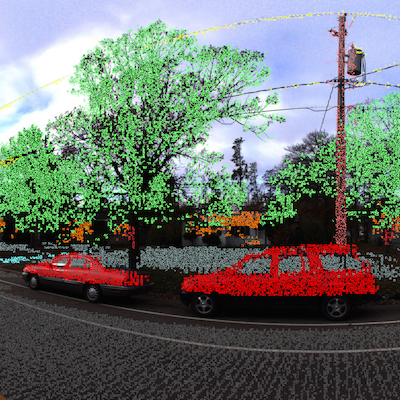}};
				
				\node [scale=1.2] (N1true) at (0+\dd,0+\ddy) {\includegraphics[height=1cm,width=0.2\textwidth,clip]{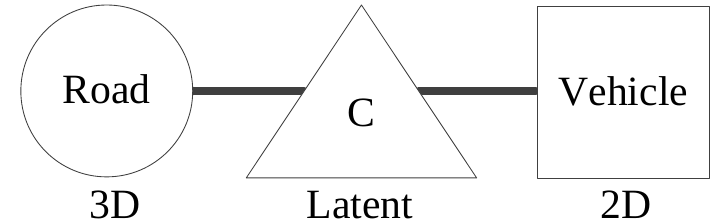}};
				\node [scale=1.2] (N2true) at (5+\dd,0+\ddy) {\includegraphics[height=1cm,width=0.2\textwidth,clip]{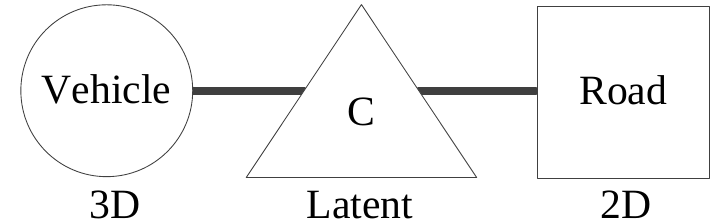}};
				\node [scale=1.2] (N3true) at (10+\dd,0+\ddy) {\includegraphics[height=1cm,width=0.2\textwidth,clip]{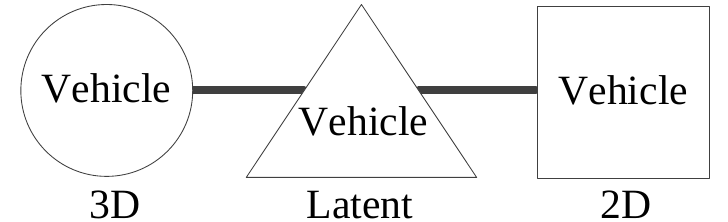}};
				
				\end{tikzpicture}
			}
		}
		\centering
		\caption{{\bf Latent nodes for moving objects.} {\bf Left:} A \emph{vehicle} was observed in the 2D image, but missing from the 3D data, since the 3D laser sensor has not covered that area when vehicle passed. As a result, the 3D points in that region are labeled as {\it road}. By relying on both 2D and 3D features, the latent node should predict that this connection must be cut. {\bf Middle:} This represents the opposite scenario where the image depicts an empty \emph{road}, while the 3D points were acquired when a \emph{vehicle} was passing. Here again, the latent node should cut the edge, thus allowing the nodes to take different labels. {\bf Right:} In contrast, here, the 2D and 3D regions belong to the same class and thus have coherent features. The latent node should therefore leverage this information to facilitate prediction of of the correct label \emph{vehicle}. }
		\label{vehicle3}
	\end{figure*}

	Our model specifies separate nodes to 2D regions (i.e., superpixels) and 3D regions (i.e., 3D segments). More details about these regions are provided in Section~\ref{F&P}. We also consider latent nodes that enable us to take into account inconsistencies between the different modalities. To this end, and as illustrated in Figure~\ref{model}, we incorporate one such latent node between each pair of corresponding 2D and 3D nodes. This results in edges between either a 2D node and a latent node, or a 3D node and a latent node, but no edges directly connecting a 2D node to a 3D node. Our latent nodes can then either take a label from the same space as the 2D and 3D nodes, or take another label indicating that the link between the two modalities should be cut (label 0).
	Figure~\ref{misalign} illustrates through an example how latent nodes
	operate in case of a misalignment between 2D and 3D data for narrow
	objects. In Figure~\ref{vehicle3} we show that multimodal
	data is prone to errors due to moving objects like a vehicle. In
	each case, latent nodes utilize the 2D and 3D information and either
	assist the linked 2D-3D regions to find their class label or cut off
	the link between them.
	
	Formally, let $\by^{2D} = \{y_{ij}^{2D}\}\;,\; 1\leq i \leq F\;,\; 1 \leq j \leq N_i$, be the set of variables encoding the labels of the 2D nodes in $F$ frames, with frame $i$ containing $N_i$ 2D regions. Similarly, let $\by^{3D} = \{y_i^{3D}\}\;,\; 1 \leq i \leq M$ be the set of variables encoding the label of $M$ 3D nodes. Each of these variables, either 2D or 3D, can take a label in the set $\mathcal{L} = \{1,\cdots,L\}$. Furthermore, let $T$ be the number of pairs of corresponding 2D and 3D nodes, found in the manner described in Section~\ref{F&P}. We then denote by $\by^{\Delta} = \{y_t^{\Delta}\}\;,\; 1 \leq t \leq T$ the latent nodes associated with these correspondences. These variables can be assigned a label from the space $\mathcal{L}' = \{0, 1, \cdots, L\}$.
	
	Given features extracted from the 2D and 3D regions, $\bx^{2D} = \{\bx_{ij}^{2D}\}$ and $\bx^{3D} = \{\bx_i^{3D}\}$, respectively, the joint distribution of the 2D, 3D and latent nodes conditioned on the features can be expressed as
	\begin{align}\label{eq:CRF_formula_latent11}
	P(\by^{2D},\by^{3D},\by^{\Delta}|\bx^{2D},\bx^{3D}) &=  \frac{1}{Z}\cdot\\
	&\hspace{-4.5cm}\exp\Bigl(-\sum_{i=1}^F\sum_{j=1}^{N_i}\Phi_{ij}^{2D}-\sum_{i=1}^M\Phi_i^{3D} - \sum_{t=1}^T\Phi_t^{\Delta} -\sum_{i=1}^F\hspace{-0.1cm}\sum_{(j,k)\in \mathcal{E}_{i}^{2D}}\hspace{-0.2cm}\Psi_{ijk}^{2D} \nonumber \\ 
	&\hspace{-3.75cm}  - \hspace{-0.3cm}\sum_{(i,j)\in \mathcal{E}^{3D}}\hspace{-0.3cm}\Psi_{ij}^{3D} - \sum_{i=1}^F\hspace{-0.1cm}\sum_{(j,t)\in\mathcal{E}^{2D-\Delta}}\hspace{-0.3cm}\Psi_{ijt}^{2D-\Delta}- \hspace{-0.3cm}\sum_{(i,t)\in\mathcal{E}^{3D-\Delta}}\hspace{-0.3cm}\Psi_{it}^{3D-\Delta}\Bigl), \nonumber
	\end{align}
	
	where
	$\Phi^{2D}$, $\Phi^{3D}$, and
	$\Phi^{\Delta}$ denote the unary potentials of the 2D, 3D and latent nodes, respectively. 
	$\Psi^{2D}$, $\Psi^{3D}$, $\Psi^{2D-\Delta}$ and $\Psi^{3D-\Delta}$ denote pairwise potentials defined over the set of edges $\mathcal{E}^{2D}$, $\mathcal{E}^{3D}$, $\mathcal{E}^{2D-\Delta}$ and $\mathcal{E}^{3D-\Delta}$, respectively. All the unary and pairwise potentials are calculated based on the formulations in Section~\ref{sec:method200} and Section~\ref{sec:method2}.
	Below, we provide some details regarding our features and potentials.
	
	\subsubsection{Features and Potentials}
	\label{F&P}
	{\bf 3D nodes: }
	We extracted the following 3D shape features from the point cloud data: fast point feature histogram (FPFH \cite{Radu:fpfh:2009}) that describes the local point distributions based on the point distances and orientations of their surface normal vectors w.r.t. each other, eigenvalue features that model the shape of the spatial distribution of the points, deviation of the surface normal vectors from the vertical axis, and also the height of the points. The 3D segments were obtained from these features by first classifying the points using an SVM classifier, partitioning them into different groups given their class labels, and then performing k-means clustering on each group of the points based on their spatial coordinates. We then further leveraged the SVM results and used the negative logarithm of the multiclass SVM probabilities as features in our unary potentials. The probabilities for a segment were obtained by averaging over the points belonging to the segment. We also used three eigenvalue descriptors and the vertical-axis deviation as additional features for the segments.
	
	{\bf 2D nodes: }
	As 2D regions, we used superpixels extracted by the mean-shift algorithm~\cite{Comaniciu:2002:MSR}. We utilized histogram of SIFT features \cite{Lowe:2004:DIF}, GLCM features (entropy, homogeneity and contrast, each computed in both horizontal and vertical directions), and RGB values to train an SVM classifier, and used the negative logarithm of the SVM probabilities as features in our unary potentials. We augmented these features with six GLCM features and three RGB features. 
	
	{\bf Latent nodes: }
	The features of the latent nodes were obtained by concatenating the features of their respective 2D and 3D nodes, described above. Furthermore, we augmented these features with the normalized overlap area of the projection of the 3D segment onto the 2D superpixel. 
	
	{\bf Edges: }
	\label{ss:edges}
	For the intra-domain potentials, we employed the $\ell_2$-norm of the difference of a subset of the local feature vectors (RGB for 2D-2D edges and vertical-axis deviation for 3D-3D edges) as pairwise features. The feature vectors of the 2D-$\Delta$ and 3D-$\Delta$ edges were set to a single value of $\mathbb{1}$. In the case of the 2D-3D CRF with no latent nodes, however, the feature vector of the 2D-3D edges was constructed by concatenating the RGB values of the 2D node with the eigenvalue features and deviation of the 3D node from the vertical-axis, as well as with the same normalized overlap area used for the unary of the latent nodes. We selected these features through an ablation study that was conducted on the validation set. As evidenced by our results, they yield better accuracies than when employing all of them, which causes overfitting. Note that we obtain the 2D-3D edges by projecting the 3D clusters onto the 2D regions and then, linking the pairs of 2D-3D elements that have a considerable projection overlap with each other, i.e., an intersection over union of more than 0.2.
	
	\subsection{Simultaneous Inference of Semantic and Geometric Classes both in 2D and 3D}
	\label{sec:method3}
	
\begin{figure}[t]
		\centering
		
		\subfigure{
			\begin{tikzpicture}[node distance = 120cm, auto, semithick]
			
			\node [scale=1.1] (N1true) at (0,2) {\includegraphics[height=2.7cm,width=0.21\textwidth,clip]{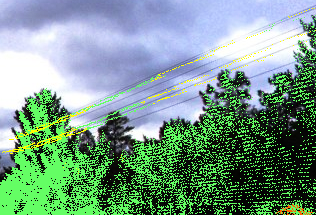}};
			\node [scale=1.1] (N2true) at (4.2,2) {\includegraphics[height=2.7cm,width=0.21\textwidth,clip]{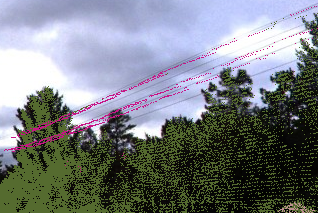}};
			\node [scale=.65] (p1) at (0,0) [rect12] {Wire-sem};
			\node [scale=.65] (p2) at (1.5,0) [rect4] {Tree leaves};
			\node [scale=.65] (p4) at (3,0) [rect19] {Wire-geo};
			\node [scale=.65] (p5) at (4.5,0) [rect18] {Scattered};

			\end{tikzpicture}
		}
		
		\centering
		\caption{{\bf Semantic labeling vs. geometric labeling.} {\bf Left:} Semantic labeling {\bf Right:} Geometric labeling. This sample image shows geometric labeling in compare with semantic labeling could distinct between wire and tree leaves.}
		\label{vehicle33}
	\end{figure}
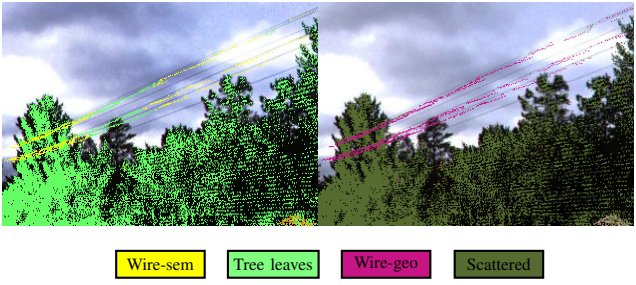

	Fusing geometric and semantic cues has shown some ability enhancing scene parsing results~\cite{Gould+al:ICCV09},~\cite{ijcv/TigheL13}. This procedure can become more promising by using 3D data geometric labeling, counter to relying on 2D data for computing geometric labels~\cite{Gould+al:ICCV09},~\cite{ijcv/TigheL13}. In Figure~\ref{vehicle33} the results of semantic and geometric labeling of wire and tree leaves are shown. In semantic labeling they were wrongly labeled as tree leaves, but using geometric labeling, they were distinct from tree leaves and correctly labeled as wire and scattered categories. This can help us improve the semantic labeling. In this paper, we use the 2D and 3D semantic labellings as well as the 2D and 3D geometric labeling collaboratively and leverage their information through a concurrent inference process to improve the labeling results in each one of them.
	~\cite{Gould+al:ICCV09},~\cite{ijcv/TigheL13} picked three categories, \emph{horizontal}, \emph{vertical} and \emph{sky}, as geometric classes in their methods. Having access to 3D point cloud data enabled us to expand this list by taking into account the \emph{cylindrical} and \emph{scattered} categories in both 2D and 3D data, which is explained in more detail in Section~\ref{subsec:expsg}. In our semantic-geometric mapping, each semantic class belongs only to one of the geometric classes, e.g., all the roads are assigned a horizontal label and all the vehicles are given a vertical label. 
	
	Let $y^{{2D}_{Sem}}$, $y^{{3D}_{Sem}}$, $y^{{2D}_{Geo}}$ and $y^{{3D}_{Geo}}$ be the variables encoding the 2D semantic, 3D semantic, 2D geometric and 3D geometric class labels, respectively.
	We can then define the joint distribution of the 2D semantic, 2D geometric, 3D semantic, 3D geometric and the latent nodes, conditioned on the node features, $P(\by^{2D_{Sem}},\by^{3D_{Sem}},\by^{2D_{Geo}},\by^{3D_{Geo}},\by^{\Delta_{2D_{Sem},2D_{Geo}}},\by^{\Delta_{3D_{Sem},3D_{Geo}}}, \by^{\Delta_{2D_{Sem},3D_{Sem}}}, \\ \by^{\Delta_{2D_{Geo},3D_{Geo}}},\by^{\Delta_{2D_{Sem},3D_{Geo}}},\by^{\Delta_{3D_{Sem},2D_{Geo}}}|\bx^{2D_{Sem}},\bx^{3D_{Sem}}, \bx^{2D_{Geo}},\bx^{3D_{Geo}}),$ similarly to the definition in Equation~\ref{eq:CRF_Gformula_latent}. 
	Note that the label set in geometric nodes and semantic nodes are different.
	
	Given that the geometric nodes represent the same set of 2D and 3D regions that were previously produced for semantic labeling, the 2D-3D geometric edges are similar to the 2D-3D semantic edges. Furthermore, note that the latent nodes which link the semantic and geometric nodes both representing one 2D region (or 3D segment), cannot cut their corresponding edges although their class labels are different. The reason behind this is that they connect two visually identical regions (segments). Instead, they try to find a coherent pair of semantic and geometric class labels that sufficiently fit the 2D and 3D features of the region (segment). The truncated TRW method is used for the inference, similar to what is described in Section~\ref{ch:training}. The inference time, however, is still quite short and satisfactory, despite the considerable increase in the size of the graph (number of nodes and edges). Table~\ref{time1} presents the training and inference times for the DATA61/2D3D and CMU/VMR datasets.
	
	Our method trains all the compatibility parameters between the semantic and geometric class labels, which contrasts with the Superparsing method~\cite{ijcv/TigheL13}, where only one parameter is embedded in the cost function to enforce consistency between these two groups of classes.
	Note that we used the same features as in Sec.~\ref{F&P} for the geometric nodes.
	
	\subsubsection{Semantic and Geometric Classes}
	\label{subsec:expsg}
	In order to best exploit the geometric cues, particularly given the 3D point cloud data, the data is  clustered into different structural classes including \emph{horizontal plane}, \emph{vertical plane}, \emph{scattered} and \emph{cylindrical} (in addition to three other groups for specifically representing \emph{sky}, \emph{person} and \emph{wire}). Table~\ref{Semantic_geometric} provides the mapping between the geometric and semantic classes.

	\begin{table*}[h]\centering\renewcommand{\arraystretch}{1.15}
		\caption{Different parameter matrices used in our experiment for two multimodal datasets. The feature sets used in each case are listed.}
		\vspace{-10pt}
		\def\d{.03cm}
		\def\r{30}
		\small\addtolength{\tabcolsep}{2.8pt}
		{
			\resizebox{\textwidth}{!}{  
				\begin{tabular}{cccccc}\\ \thickhline
					\multicolumn{3}{c}{\bf DATA61/2D3D}&\multicolumn{3}{|c}{\bf CMU/VMR}\\ \thickhline
					
					{\bf Parameter Matrices}&{\bf Rows}&{\bf Columns}&{\bf Parameter Matrices}&{\bf Rows}&{\bf Columns}\\
					\thickhline
					
					{\large $\mathbf{A}_{[14\times23]}^{2D}$}&14 classes&14 2D-probabilities + 6 GLCM + 3 RGB&{\large $\mathbf{A}_{[19\times28]}^{2D}$}&19 classes&19 2D-probabilities + 6 GLCM + 3 RGB\\
					\thickhline
					
					{\large $\mathbf{A}_{[13\times17]}^{3D}$}&13 classes&13 3D-probabilities + 3 Eigenvalues + 1 $z$-deviation&{\large $\mathbf{A}_{[19\times23]}^{3D}$}&19 classes&19 3D-probabilities + 3 Eigenvalues + 1 $z$-deviation\\
					\thickhline
					
					{\large $\mathbf{A}_{[15\times41]}^{\Delta}$}&14 classes + 1 edge-cut &23 2D-features + 17 3D-features + 1 2D-3D overlap &{\large $\mathbf{A}_{[20\times52]}^{\Delta}$}&19 classes + 1 edge-cut &28 2D-features + 23 3D-features + 1 2D-3D overlap \\
					\thickhline
					
					{\large $\mathbf{B}_{[196\times1]}^{2D}$}&14$\times$14 classes&1 &{\large $\mathbf{B}_{[361\times1]}^{2D}$}&19$\times$19 classes&1 \\
					\thickhline
					
					{\large $\mathbf{B}_{[169\times1]}^{3D}$}&13$\times$13 classes&1 &{\large $\mathbf{B}_{[361\times1]}^{3D}$}&19$\times$19 classes&1 \\
					\thickhline
					
					{\large $\mathbf{B}_{[210\times1]}^{2D-\Delta}$}&14$\times$15 classes&1 &{\large $\mathbf{B}_{[380\times1]}^{2D-\Delta}$}&19$\times$20 classes&1 \\
					\thickhline
					
					{\large $\mathbf{B}_{[195\times1]}^{3D-\Delta}$}&13$\times$15 classes&1 &{\large $\mathbf{B}_{[380\times1]}^{3D-\Delta}$}&19$\times$20 classes&1 \\
					\thickhline
					
					{\large $\mathbf{B}_{[182\times1]}^{2D-3D}$}&14$\times$13 classes&1 &{\large $\mathbf{B}_{[361\times1]}^{2D-3D}$}&19$\times$19 classes&1 \\
					\thickhline
					
					{\large $\mathbf{B}_{[182\times8]}^{2D-3D}$}&14$\times$13 classes&3 RGB + 3 Eigenvalues + 1 $z$-deviation + 1 2D-3D overlap &{\large $\mathbf{B}_{[361\times8]}^{2D-3D}$}&19$\times$19 classes&3 RGB + 3 Eigenvalues + 1 $z$-deviation + 1 2D-3D overlap \\										
					\thickhline
					
					{\large \multirow{2}{*}{$\mathbf{B}_{[182\times41]}^{2D-3D}$}}&\multirow{2}{*}{14$\times$13 classes}&3 RGB + 3 Eigenvalues + 1 $z$-deviation + 1 2D-3D overlap &{\large \multirow{2}{*}{$\mathbf{B}_{[361\times52]}^{2D-3D}$}}&\multirow{2}{*}{19$\times$19 classes}&3 RGB + 3 Eigenvalues + 1 $z$-deviation + 1 2D-3D overlap \\
					
					&&6 GLCM + 14 2D-probabilities + 13 3D-probabilities&&&6 GLCM + 19 2D-probabilities + 19 3D-probabilities \\
					\thickhline
					
				\end{tabular}
			}
		}
		
		\label{matrices_table}
	\end{table*}
	
	\section{Experiments}\label{ch:experiments}
	We evaluate our method on two publicly available 2D-3D multimodal datasets (DATA61/2D3D~\cite{wacv2015} and CMU/VMR~\cite{Munoz:2012:eccv}). KITTI~\cite{Geiger:2012:CVPR} is another well-known multimodal dataset used for outdoor scene understanding and object detection, and it has recently been adapted to be used as a benchmark for semantic labeling task \cite{Cadena:2014:icra}. However, the point cloud data in this dataset has a pretty small vertical field of view \cite{Xie:2016:CVPR} and as a result, a large portion of 2D images do not have a correspondence in the 3D data. Therefore in our application where we are interested in investigating the 2D-3D links, this dataset is less useful and thus we evaluate our method on the two above mentioned multimodal datasets. 
	
	We provide the results of 2D-3D CRF with and without latent nodes and also simultaneous inference of semantic and geometric classes both in 2D and 3D. We also compare the results to the state-of-the-art algorithms of~\cite{wacv2015} and~\cite{Munoz:2012:eccv}. The experiment on 2D-3D CRF without latent nodes is a special case study of the general multimodal CRF (Section~\ref{sec:method200}). In addition, we provide the results of the pairwise models with learned potentials acting on a single domain, either 2D or 3D. These models are referred to as \emph{Pairwise 2D (learned)} and \emph{Pairwise 3D (learned)}. 
	We followed the evaluation protocol of \cite{wacv2015} and partitioned the data into 4 non-overlapping folds. We then used three of the folds for training and the remaining fold as test set.

	\subsection{Results on DATA61/2D3D}
	The DATA61/2D3D dataset contains 12 outdoor scenes where each scene is described by a 3D point cloud block together with 10-20 panoramic images. The number of 3D points in the scenes varies from 1 to 2 millions. It comprises 14 classes (13 for 3D where {\it sky} was removed), which yields the following sizes for the parameter matrices for 2D-3D CRF with latent nodes: $\mathbf{A}_{[14\times23]}^{2D}$, $\mathbf{A}_{[13\times17]}^{3D}$, $\mathbf{A}_{[15\times41]}^{\Delta}$, $\mathbf{B}_{[196\times1]}^{2D}$, $\mathbf{B}_{[169\times1]}^{3D}$, $\mathbf{B}_{[210\times1]}^{2D-\Delta}$ and $\mathbf{B}_{[195\times1]}^{3D-\Delta}$. The 2D-3D CRF with no latent nodes involves a different parameter matrix of the form $\mathbf{B}_{[182\times1]}^{2D-3D}$, $\mathbf{B}_{[182\times8]}^{2D-3D}$ and $\mathbf{B}_{[182\times41]}^{2D-3D}$. Table~\ref{matrices_table} lists these parameters matrices and describes how their size relates with the selected set of features and class labels in the experiments.
	\vspace{2pt}
	
	Table~\ref{New_2d_table} and Table~\ref{New_3d_table} compare the results, as F1-scores, of the 2D-3D CRF model with handcrafted and learned potentials and also with latent nodes and no latent nodes. Note that no results for~\cite{Munoz:2012:eccv} are available on this dataset. The results in these tables evidence the benefits of using latent nodes, especially on the narrow classes that suffer more from misalignment. On average, our approach with latent nodes clearly outperforms the model with no latent nodes, and thus achieves state-of-the-art results on this dataset. Moreover, note that the 2D-3D CRF with no latent nodes that utilizes fewer features (selected features) for the 2D-3D edges is less likely to face overfitting and yields better results, compared with the CRF model with a full set of features. Furthermore, the results of the 2D-3D CRF with no latent nodes, where the feature vector of the 2D-3D edges were set to a single value of 1 are presented in Table~\ref{New_2d_table} and Table~\ref{New_3d_table} for comparison (no feature).
	
\begin{figure*}[t]
		\centering
		\def\dd{.75}
		\def\dh{.75}
		\subfigure{
			\begin{tikzpicture}
			
			\coordinate [] (ref0) at (0,1.7);
			\node [right =-1.4cm of ref0] (note1) {\footnotesize Image};
			\node [right =.65cm of note1] (note2) {\footnotesize 2D ground-truth};
			\node [right =.0cm of note2] (note3) {\footnotesize 3D ground-truth};
			\node [right =-.0cm of note3] (note4) {\footnotesize 3D-2D projection};
			\node [right =.3cm of note4] (note5) {\footnotesize 2D results \cite{wacv2015}};
			\node [right =.25cm of note5] (note6) {\footnotesize 3D results \cite{wacv2015}};
			\node [right =.20cm of note6] (note7) {\footnotesize Our 2D results};
			\node [right =.25cm of note7] (note8) {\footnotesize Our 3D results};
			
			\coordinate [] (ref) at (0,0);
			\node [right =-2cm of ref,xscale=\dd] (img1) {\includegraphics[height=2.7cm,width=.16\textwidth]{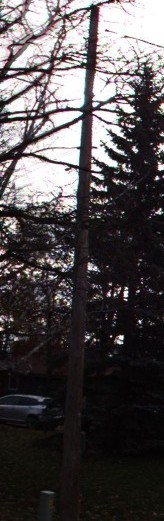}};
			\node [right =-.2cm of img1,xscale=\dd] (img2) {\includegraphics[height=2.7cm,width=.16\textwidth]{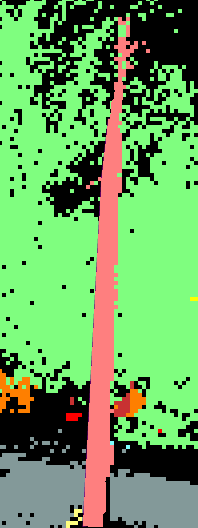}};
			\node [right =-.2cm of img2,xscale=\dd] (img3) {\includegraphics[height=2.7cm,width=.16\textwidth]{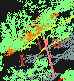}};
			\node [right =-.2cm of img3,xscale=\dd] (img4) {\includegraphics[height=2.7cm,width=.16\textwidth]{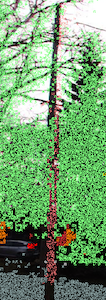}};
			\node [right =-.2cm of img4,xscale=\dd] (img5) {\includegraphics[height=2.7cm,width=.16\textwidth]{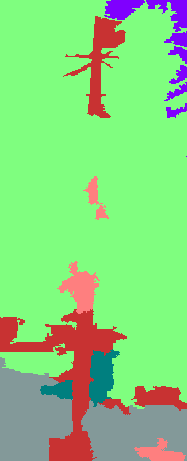}};
			\node [right =-.2cm of img5,xscale=\dd] (img6) {\includegraphics[height=2.7cm,width=.16\textwidth]{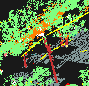}};
			\node [right =-.2cm of img6,xscale=\dd] (img7) {\includegraphics[height=2.7cm,width=.16\textwidth]{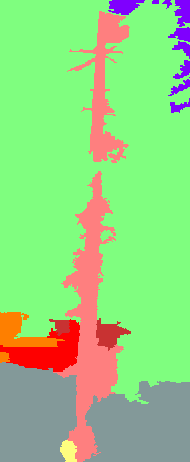}};
			\node [right =-.2cm of img7,xscale=\dd] (img8) {\includegraphics[height=2.7cm,width=.16\textwidth]{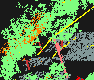}};
			
			\coordinate [] (ref2) at (0,-2.5cm);
			\node [right =-2cm of ref2,xscale=\dd,yscale=\dh] (im1) {\includegraphics[height=2.7cm,width=.16\textwidth]{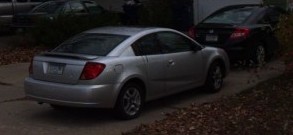}};
			\node [right =-.2cm of im1,xscale=\dd,yscale=\dh] (im2) {\includegraphics[height=2.7cm,width=.16\textwidth]{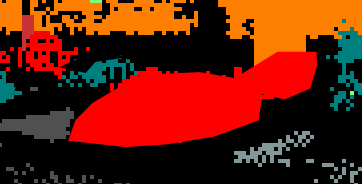}};
			\node [right =-.2cm of im2,xscale=\dd,yscale=\dh] (im3) {\includegraphics[height=2.7cm,width=.16\textwidth]{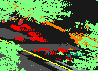}};
			\node [right =-.2cm of im3,xscale=\dd,yscale=\dh] (im4) {\includegraphics[height=2.7cm,width=.16\textwidth]{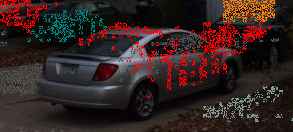}};
			\node [right =-.2cm of im4,xscale=\dd,yscale=\dh] (im5) {\includegraphics[height=2.7cm,width=.16\textwidth]{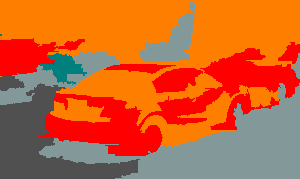}};
			\node [right =-.2cm of im5,xscale=\dd,yscale=\dh] (im6) {\includegraphics[height=2.7cm,width=.16\textwidth]{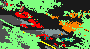}};
			\node [right =-.2cm of im6,xscale=\dd,yscale=\dh] (im7) {\includegraphics[height=2.7cm,width=.16\textwidth]{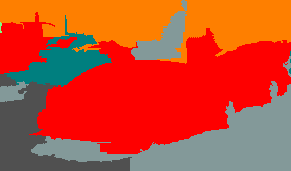}};
			\node [right =-.2cm of im7,xscale=\dd,yscale=\dh] (im8) {\includegraphics[height=2.7cm,width=.16\textwidth]{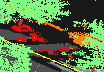}};
			
			\draw [ultra thick, white, ->] (3.8,-2.3) -- (3.8,-2.8);
			\draw [ultra thick, white, ->] (10.25,-2.4) -- (10.25,-2.9);
			\draw [ultra thick, white, ->] (14.55,-2.3) -- (14.55,-2.8);
			
			\end{tikzpicture}
		}
		\scalebox{.8}{
			\subfigure{
				\begin{tikzpicture}[baseline,decoration=brace,scale=.8]
				
				\node [] (hmn1) at (0,.8) [rect1] {Grass};
				\node [] (hmn2) at (3,.8) [rect2] {Building};
				\node [] (hmn3) at (6,.8) [rect3] {Tree trunk};
				\node [] (hmn4) at (9,.8) [rect4] {Tree leaves};
				\node [] (hmn5) at (12,.8) [rect5] {Vehicle};
				\node [] (hmn6) at (15,.8) [rect6] {Road };
				\node [] (hmn8) at (18,.8) [rect8] {Pole};
				\node [] (hmn12) at (21,.8) [rect12] {Wire};
				
				\end{tikzpicture}
			}
		}
		\centering
		\caption{{\bf Examples of how our latent nodes improve the labeling in practice.} As shown in the 3D-2D projection, the data misalignment and object motions have caused 3D points labeled as \emph{leaves} to cover the \emph{pole} (top) and 3D points labeled as \emph{road} to project onto the \emph{vehicles} (bottom). Consequently, by applying the approach in \cite{wacv2015} which encourages the corresponding nodes in two modalities to take identical class labels, the pole was segmented as \emph{leaves} in the image domain and the vehicle was labeled as \emph{road} in 3D data (highlighted by a white arrow). On the contrary, thanks to our latent nodes that can cut inconsistent edges, our method produces the correct labels.}
		\label{latentFig}
	\end{figure*}
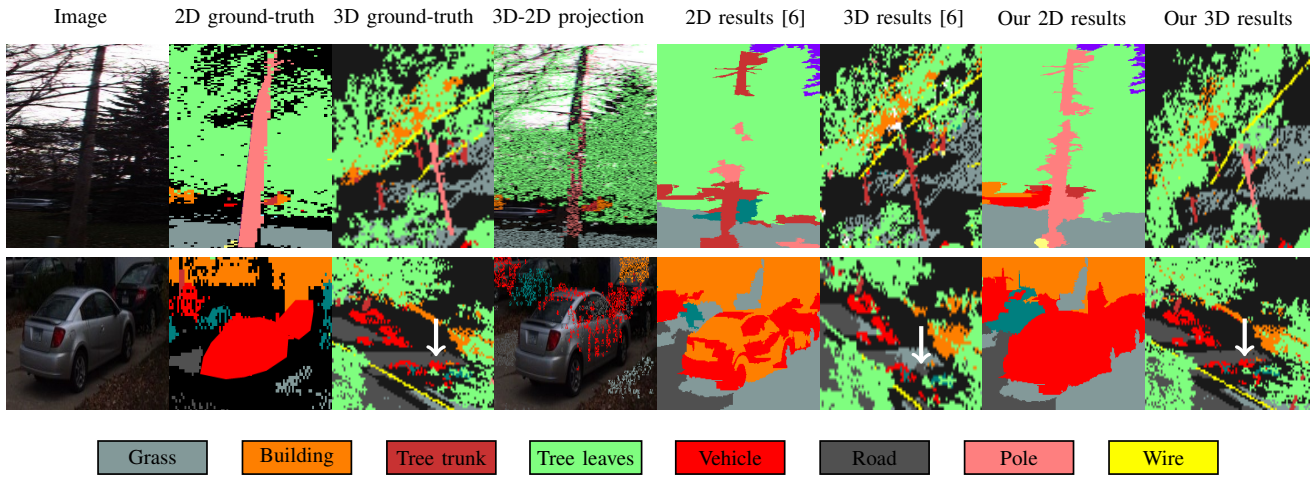

\begin{figure*}[b]
		\centering
		\def\sc{.7}
		\def\dc{-.05}
		
		\begin{tikzpicture}
		
		\coordinate [] (ref0) at (-.1cm,0);
		\node [right =-.2cm of ref0] (2D1) {\includegraphics[height=2.6cm,width=0.33\textwidth,trim={0 4.8cm 0 3.7cm},clip]{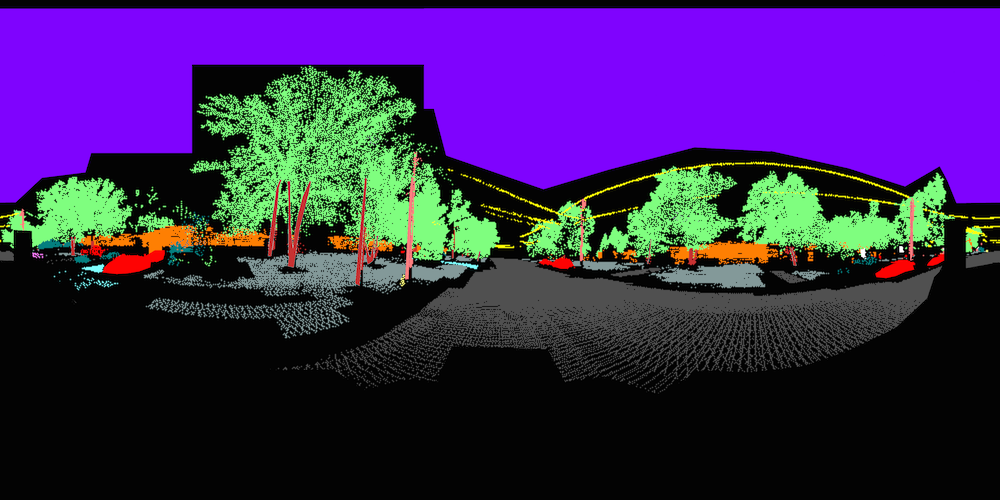}};
		
		\node [right =-.2cm of 2D1] (2D2) {\includegraphics[height=2.6cm,width=0.33\textwidth,trim={0 4.8cm 0 3.7cm},clip]{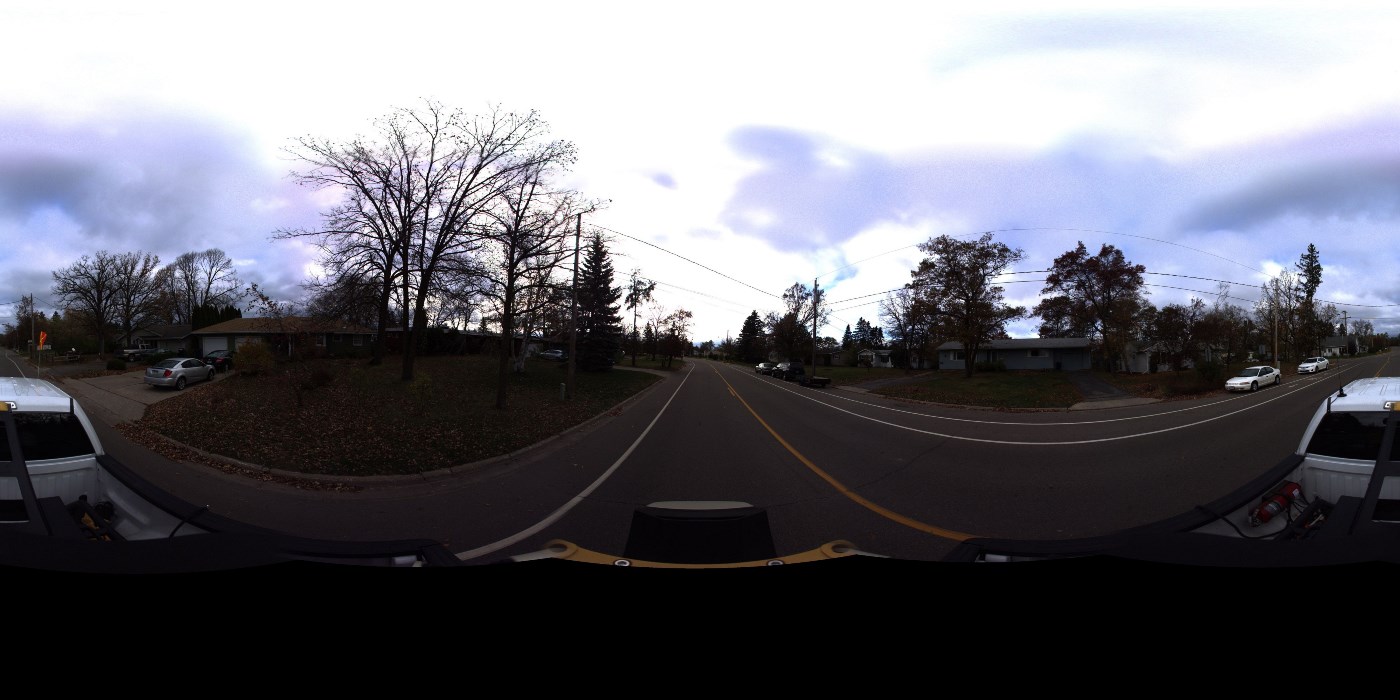}};
		\node [opacity=0.5,right =-.2cm of 2D1] (2D3) {\includegraphics[height=2.6cm,width=0.33\textwidth,trim={0 4.8cm 0 3.7cm},clip]{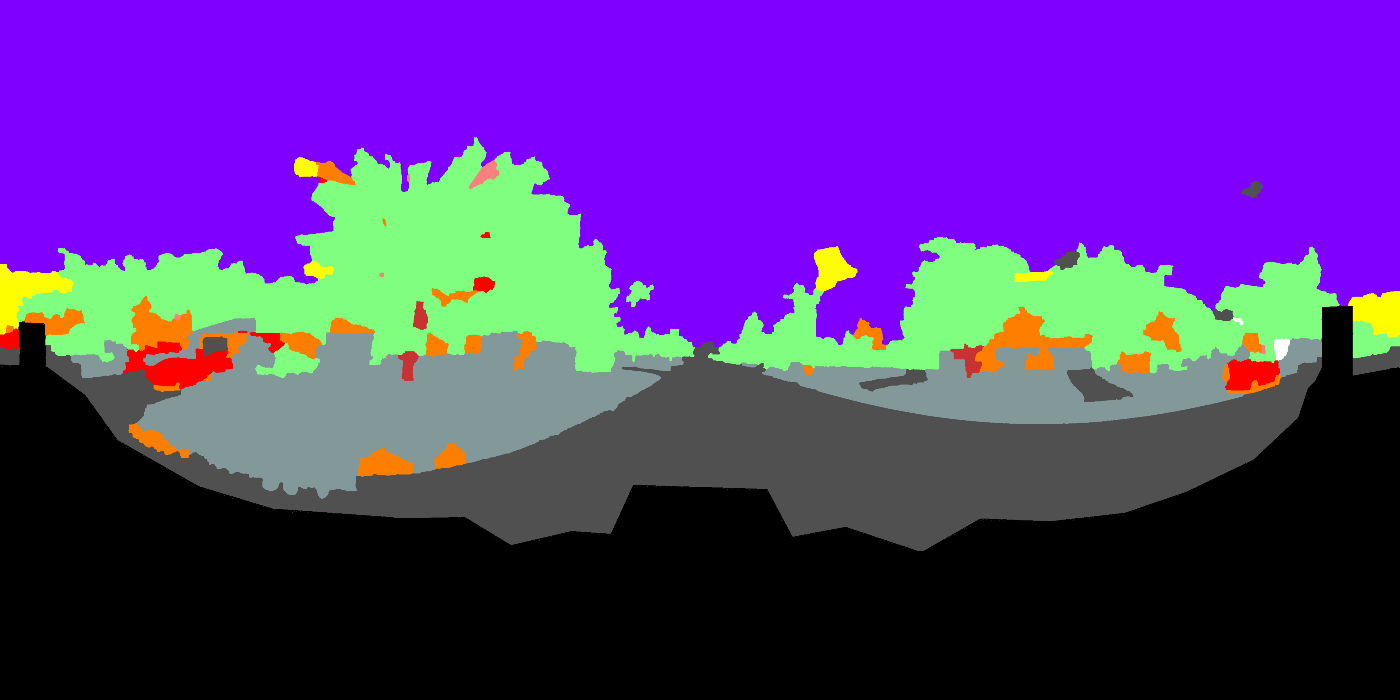}};
		
		\node [right =-.2cm of 2D2] (2D4) {\includegraphics[height=2.6cm,width=0.33\textwidth,trim={0 4.8cm 0 3.7cm},clip]{result1/frame1365.jpg}};
		\node [opacity=0.5,right =-.2cm of 2D2] (2D5) {\includegraphics[height=2.6cm,width=0.33\textwidth,trim={0 4.8cm 0 3.7cm},clip]{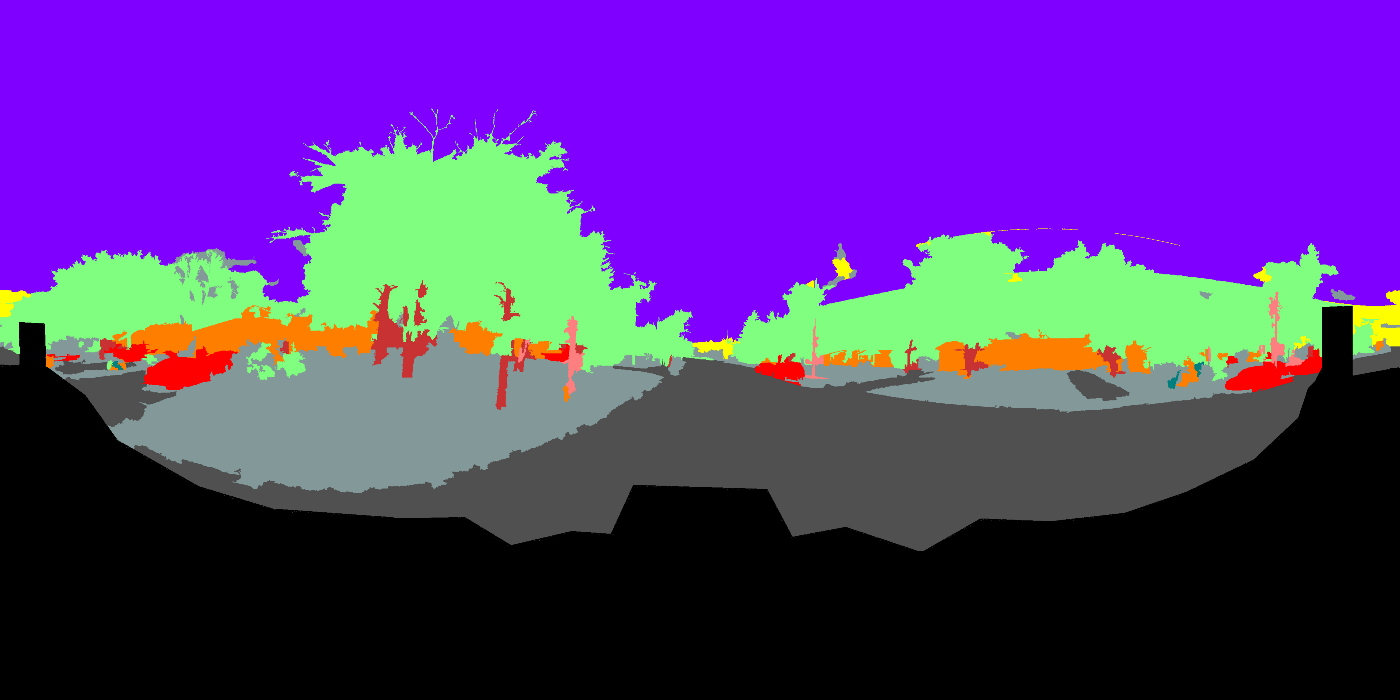}};
		
		\node [below =-.2cm of 2D1] (3D1) {\includegraphics[height=4.3cm,width=0.33\textwidth,trim={0 .2cm 0 0cm},clip]{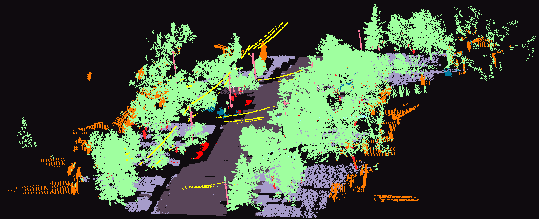}};
		\node [right =-.2cm of 3D1] (3D2) {\includegraphics[height=4.3cm,width=0.33\textwidth]{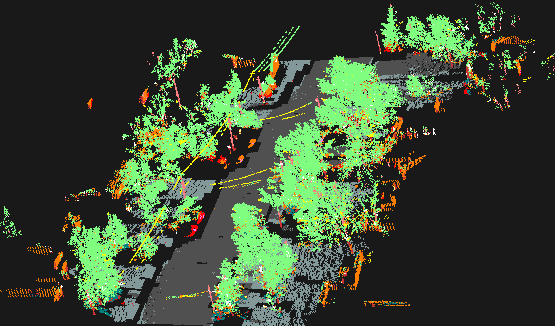}};
		\node [right =-.2cm of 3D2] (3D3) {\includegraphics[height=4.3cm,width=0.33\textwidth]{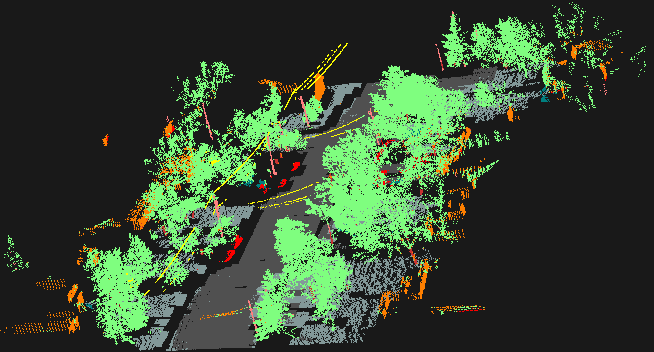}};
		
		\node  [draw,white,thick,circle,minimum size=1.2cm] (D1) at (14.3,-2.3){};
		\node  [draw,white,thick,circle,minimum size=1.2cm] (D2) at (8.6,-2.2){};
		
		\node  [draw,white,thick,circle,minimum size=1.2cm] (D3) at (14,-3.5){};
		\node  [draw,white,thick,circle,minimum size=1.2cm] (D4) at (8.2,-3.4){};
		
		\node  [draw,white,thick,circle,minimum size=1.2cm] (D5) at (13.5,-.2){};
		\node  [draw,white,thick,circle,minimum size=1.2cm] (D6) at (7.6,-.25){};
		
		\node  [draw,white,thick,circle,minimum size=1.2cm] (D7) at (15.8,-.2){};
		\node  [draw,white,thick,circle,minimum size=1.2cm] (D8) at (10,-.2){};
		
		\coordinate [] (ref1) at (.5cm,-2cm);
		\node [below =4cm of ref1,scale=\sc] (b1) [rect1] {Grass};
		\node [right =\dc cm of b1,scale=\sc] (b2) [rect2] {Building};
		\node [right =\dc cm of b2,scale=\sc] (b3) [rect3] {Tree trunk};
		\node [right =\dc cm of b3,scale=\sc] (b4) [rect4] {Tree leaves};
		\node [right =\dc cm of b4,scale=\sc] (b5) [rect5] {Vehicle};
		\node [right =\dc cm of b5,scale=\sc] (b6) [rect6] {Road};
		\node [right =\dc cm of b6,scale=\sc] (b7) [rect7] {Bush};
		\node [right =\dc cm of b7,scale=\sc] (b8) [rect8] {Pole};
		\node [right =\dc cm of b8,scale=\sc] (b9) [rect9] {Sign};
		\node [right =\dc cm of b9,scale=\sc] (b10)  [rect10] {Post};
		\node [right =\dc cm of b10,scale=\sc] (b11) [rect11] {Barrier};
		\node [right =\dc cm of b11,scale=\sc] (b12) [rect12] {Wire};
		\node [right =\dc cm of b12,scale=\sc] (b13) [rect13] {Sidewalk};
		\node [right =\dc cm of b13,scale=\sc] (b14) [rect14] {Sky};
		
		\end{tikzpicture}

		\vspace{-.0cm}
		
		\centering
		\caption{Sample results on the DATA61/2D3D dataset. {\bf1st row:} {\bf Left:} 2D ground-truth; {\bf Middle:} 2D results of \cite{wacv2015}; {\bf Right:} our 2D results. {\bf2nd row:} {\bf Left:} 3D ground-truth; {\bf Middle:} 3D results of \cite{wacv2015}; {\bf Right:} our 3D results. Our approach (the right figures) has managed to correct some of the mislabellings in the results of \cite{wacv2015}, e.g. the \emph{vehicles} and \emph{wires} in 3D domain, and \emph{poles} and \emph{tree trunks} in 2D domain. In fact, these are the categories that are most likely to be affected by misalignments.}
		\label{DATA61DatasetResultsFig}
	\end{figure*}

	In Figure~\ref{latentFig}, we illustrate the influence of our latent nodes by two examples. As shown in the figure, cutting the edge between the non-matching 2D and 3D nodes (which have been connected because of misalignment) helps predicting the correct class labels. Figure~\ref{DATA61DatasetResultsFig} shows the results of our approach in one of the scenes in this dataset, compared to the results of \cite{wacv2015}.
	
	Our results on DATA61/2D3D indicate that, while our latent nodes are in general beneficial, thanks to their ability to cut incorrect connections, they still occasionally yield lower performance than a model without such nodes. We observed that this is mainly due to the inaccurate ground-truth (which is inevitable because of the imperfect 3D-2D projection of the ground-truth labels particularly at the boundaries of the narrow objects), or to the fact that, sometimes, eventhough the 2D and 3D features seem to be inconsistent (e.g., due to challenging viewing conditions), they still belong to the same category. In these circumstances, the stronger smoothness imposed by the model without latent nodes is then able to address this problem.
	
	2D-3D multimodal scene parsing on semantic and geometric classes can be seen as a special case of our multimodal model with four modalities. 
	We considered six geometric classes in the DATA61/2D3D dataset (Table~\ref{Semantic_geometric}) and conducted similar procedures as in the semantic labeling for finding their regions and node features. 
	The 2D and 3D geometric data augment the semantic model as two separate data modalities and their simultaneous inference is carried out given the semantic and geometric cues of the 2D and 3D data. Tables~\ref{New_2d_table} and \ref{New_3d_table} demonstrate the results of the 2D and 3D semantic scene parsing using the proposed semantic and geometric 2D/3D multimodal model. As reported in this table, leveraging the geometric cues has led to 4\% and 5\% improvement in F1-scores of the 2D and 3D data, respectively. 
	The results of the geometric labeling of the 2D and 3D data are shown in Table~\ref{DATA61_geometry_table}.
	
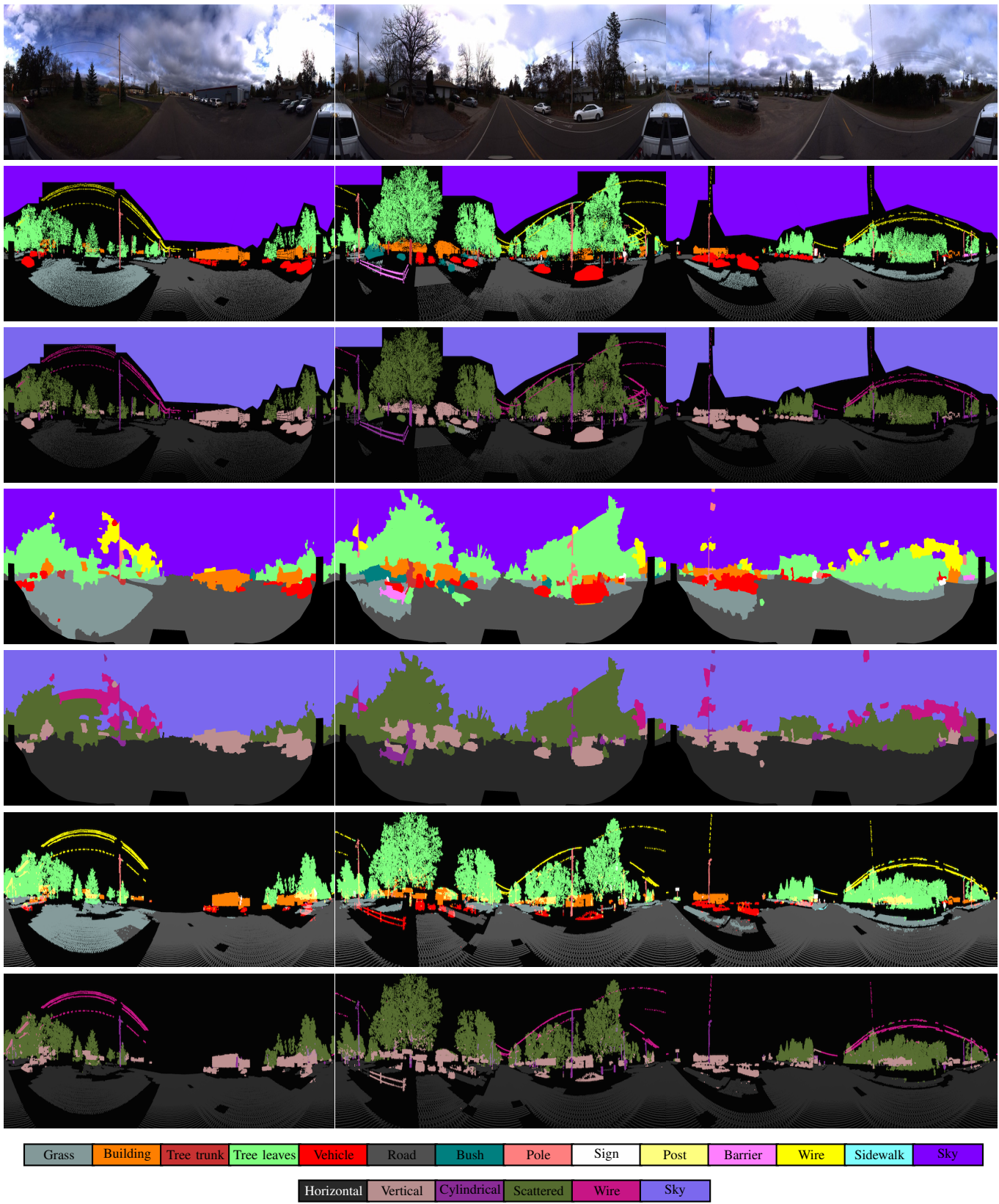
\begin{figure*}[b]
		\centering
		\def\sc{.7}
		\def\dc{-.05}
		
		\begin{tikzpicture}
		
		\coordinate [] (ref0) at (-.1cm,0);
		\node [right =-.52cm of ref0] (2D1) {\includegraphics[height=2.8cm,width=0.33\textwidth,trim={0 4.8cm 0 3.7cm},clip]{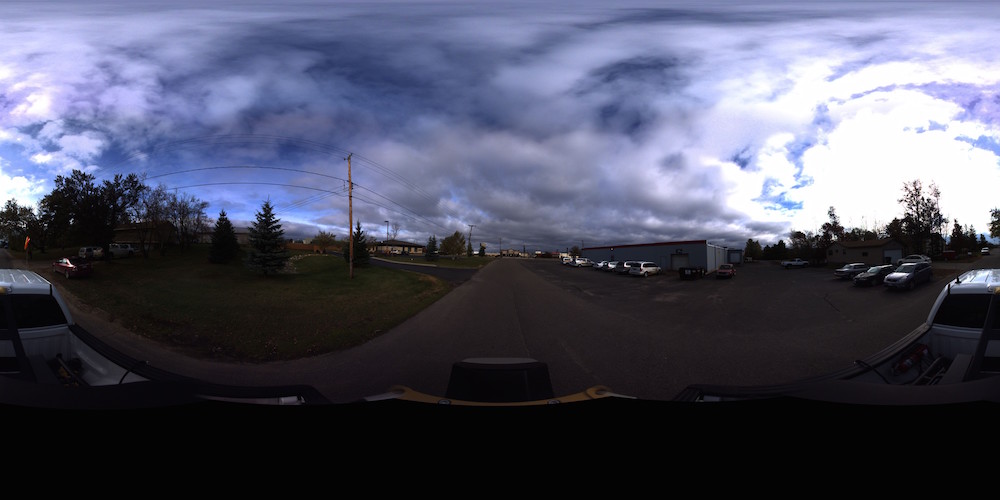}};

		\node [below =-.12cm of 2D1] (2D2) {\includegraphics[height=2.8cm,width=0.33\textwidth,trim={0 4.8cm 0 3.7cm},clip]{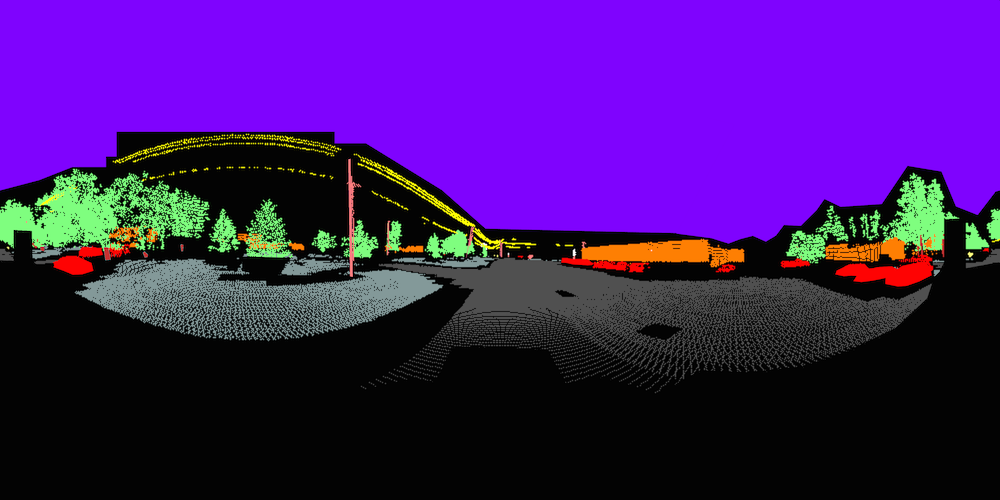}};

		\node [below =-.12cm of 2D2] (2D3) {\includegraphics[height=2.8cm,width=0.33\textwidth,trim={0 4.8cm 0 3.7cm},clip]{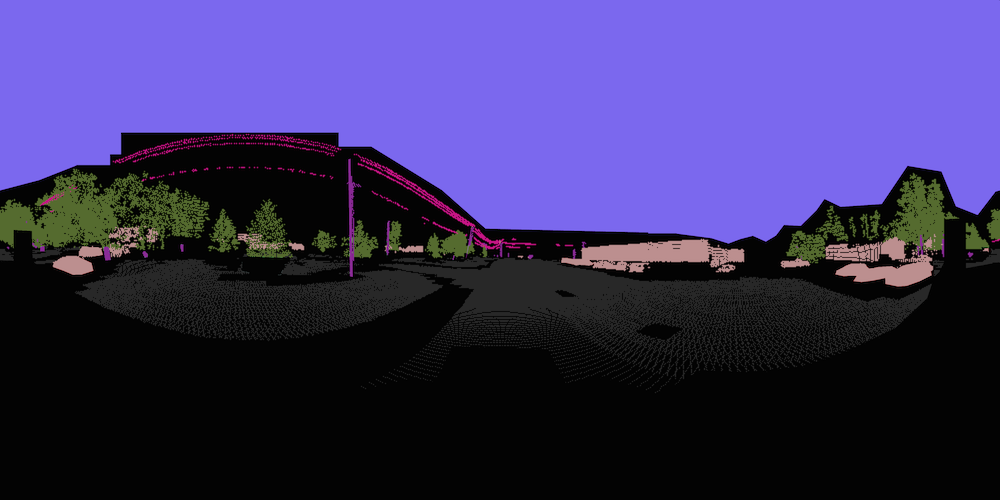}};

		\node [below =-.12cm of 2D3] (2D4) {\includegraphics[height=2.8cm,width=0.33\textwidth,trim={0 18cm 0 14cm},clip]{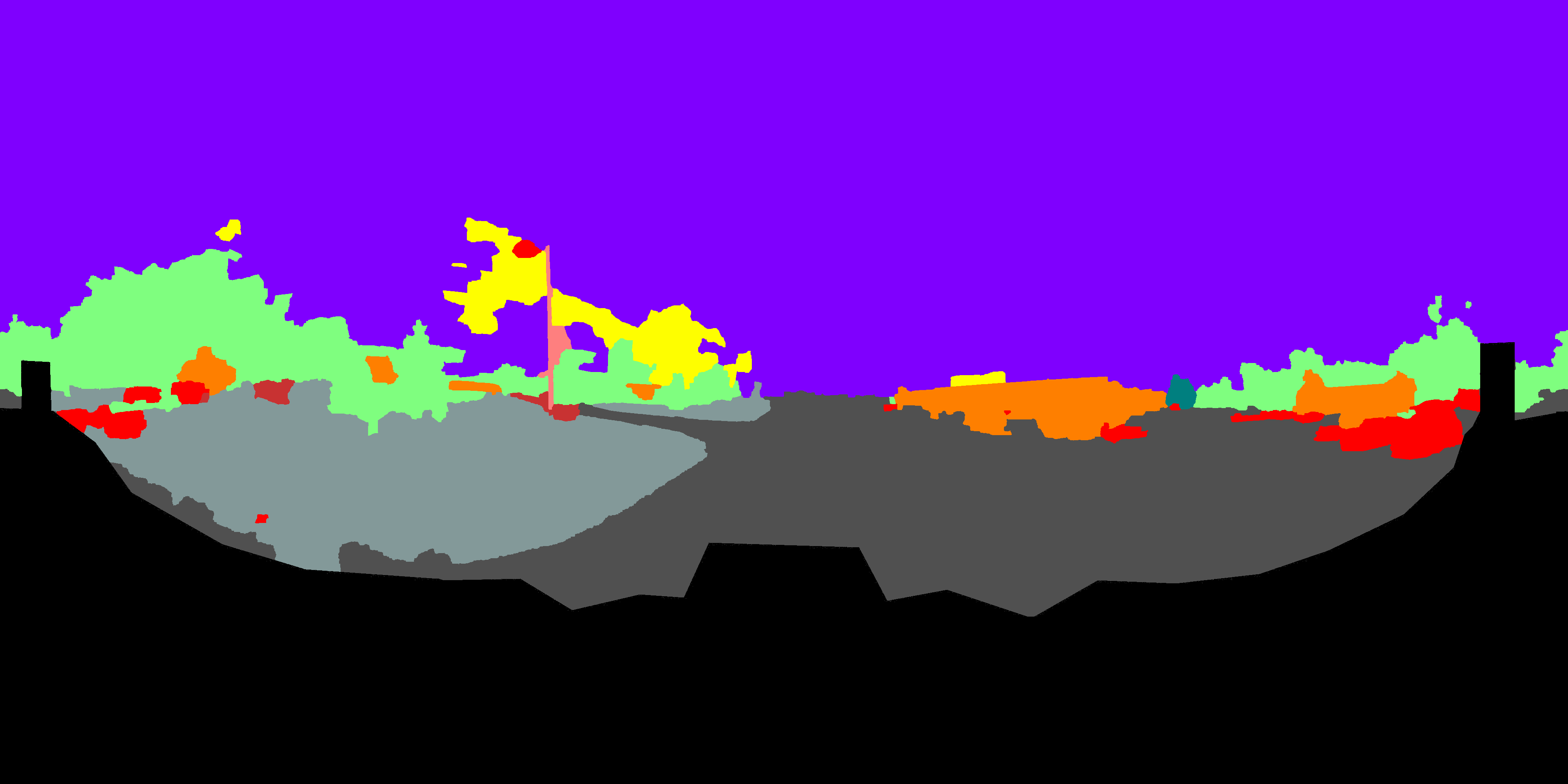}};

		\node [below =-.12cm of 2D4] (2D5) {\includegraphics[height=2.8cm,width=0.33\textwidth,trim={0 18cm 0 14cm},clip]{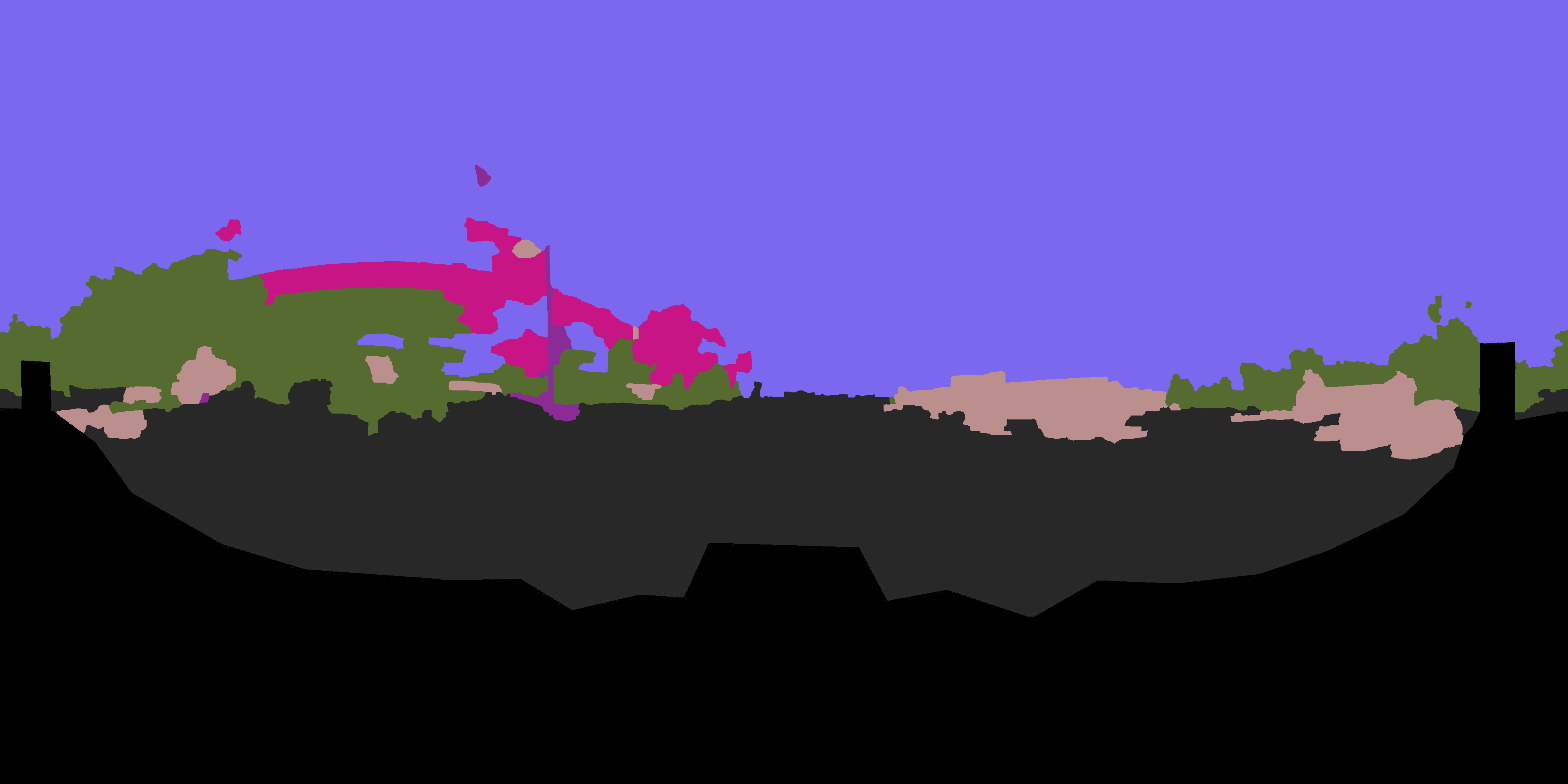}};

		\node [below =-.12cm of 2D5] (2D6) {\includegraphics[height=2.8cm,width=0.33\textwidth,trim={0 4.8cm 0 3.7cm},clip]{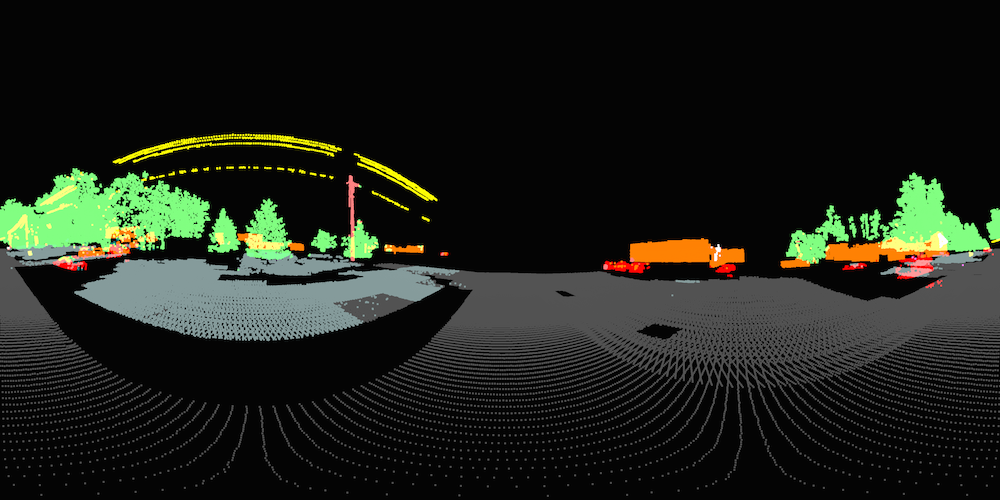}};

		\node [below =-.12cm of 2D6] (2D7) {\includegraphics[height=2.8cm,width=0.33\textwidth,trim={0 4.8cm 0 3.7cm},clip]{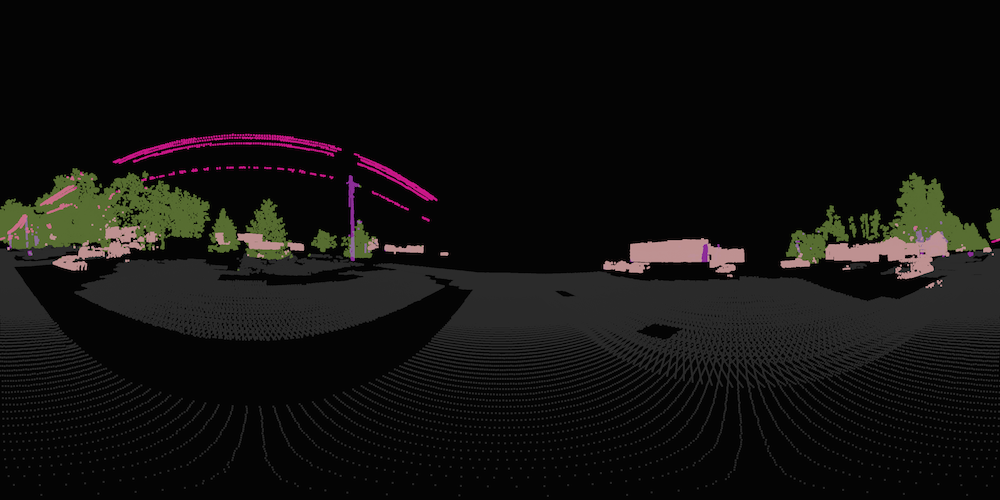}};
		
		\node [right =-.23cm of 2D1] (2D11) {\includegraphics[height=2.8cm,width=0.33\textwidth,trim={0 4.8cm 0 3.7cm},clip]{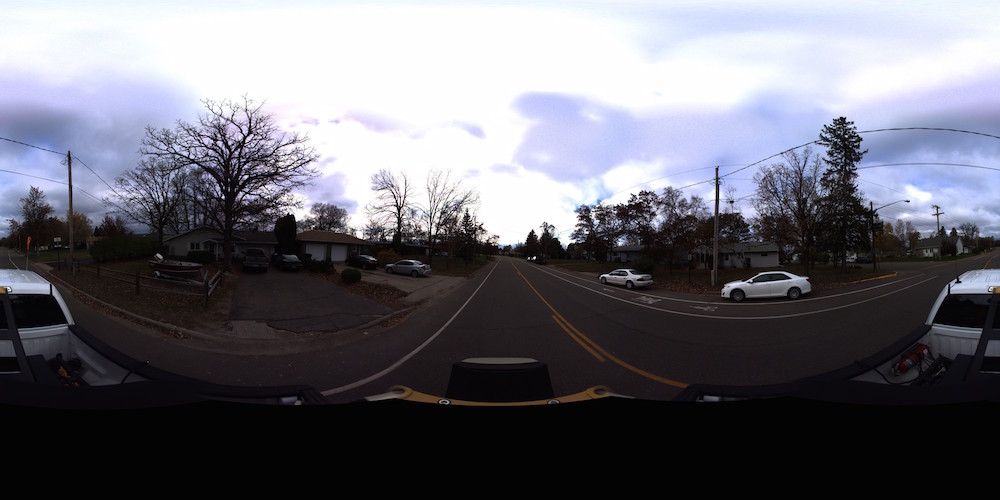}};
		
		\node [below =-.12cm of 2D11] (2D12) {\includegraphics[height=2.8cm,width=0.33\textwidth,trim={0 4.8cm 0 3.7cm},clip]{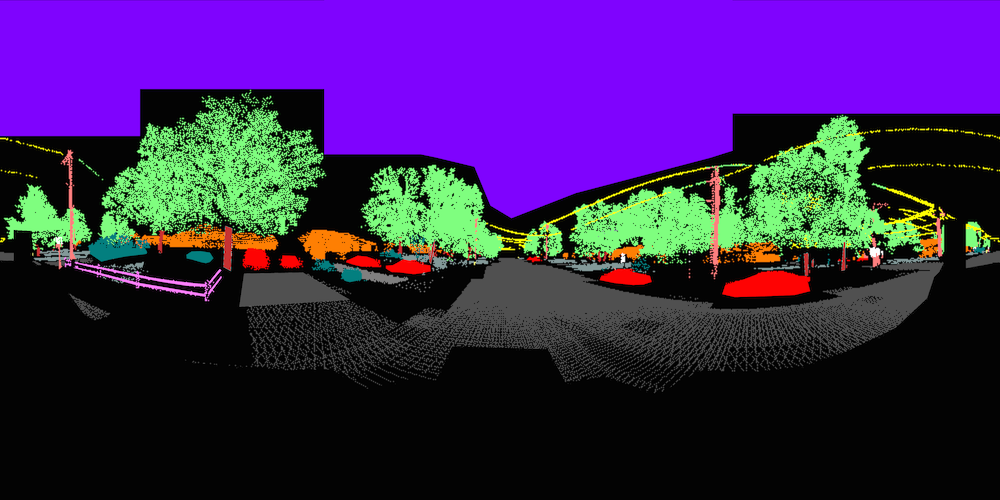}};

		\node [below =-.12cm of 2D12] (2D13) {\includegraphics[height=2.8cm,width=0.33\textwidth,trim={0 4.8cm 0 3.7cm},clip]{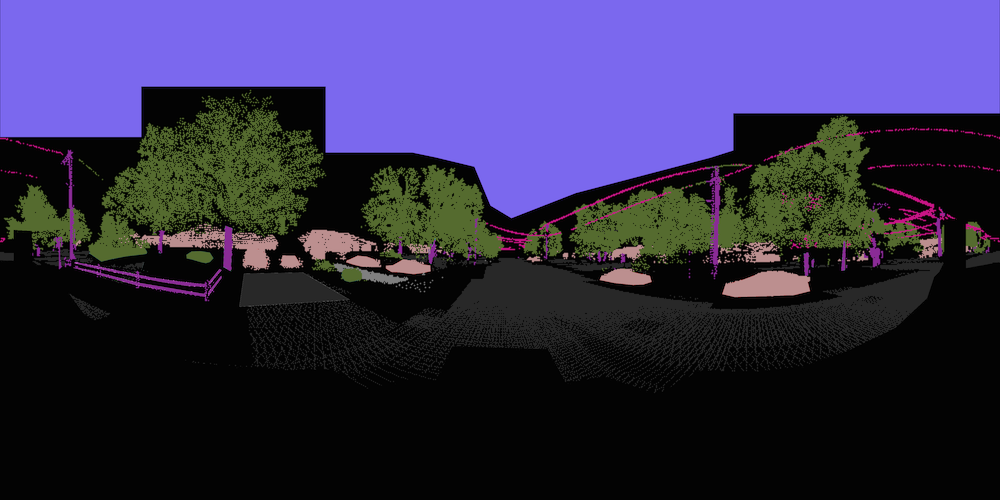}};

		\node [below =-.12cm of 2D13] (2D14) {\includegraphics[height=2.8cm,width=0.33\textwidth,trim={0 18cm 0 14cm},clip]{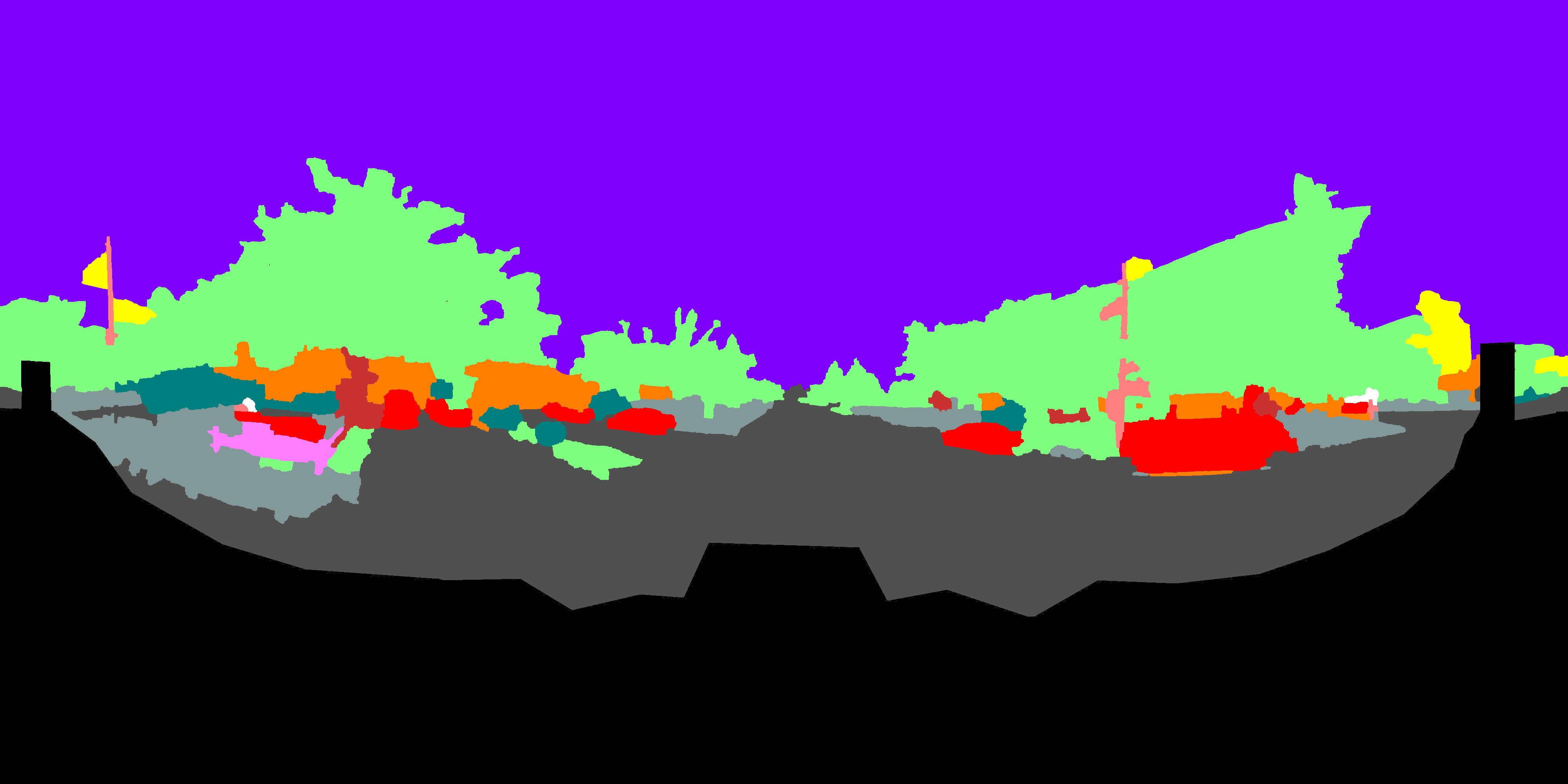}};

		\node [below =-.12cm of 2D14] (2D15) {\includegraphics[height=2.8cm,width=0.33\textwidth,trim={0 18cm 0 14cm},clip]{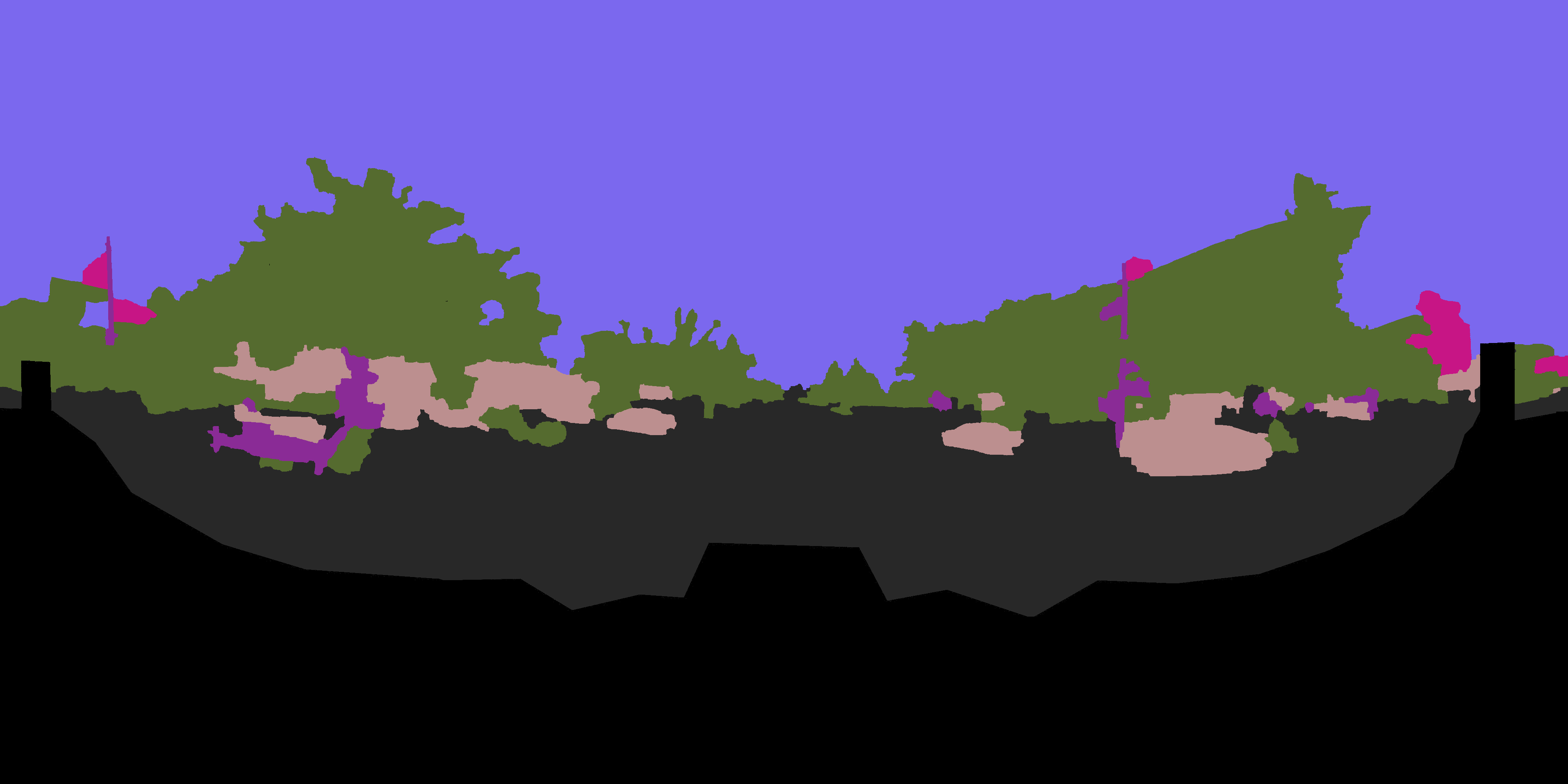}};

		\node [below =-.12cm of 2D15] (2D16) {\includegraphics[height=2.8cm,width=0.33\textwidth,trim={0 4.8cm 0 3.7cm},clip]{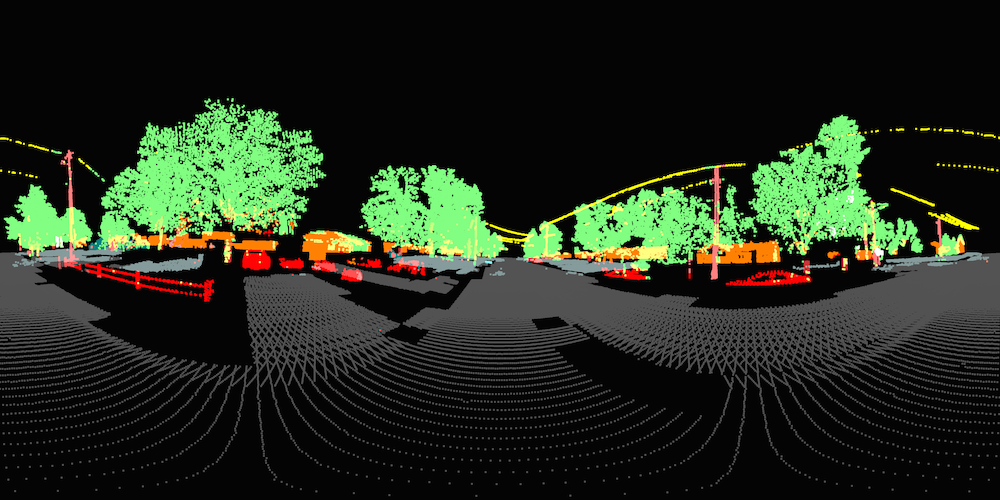}};

		\node [below =-.12cm of 2D16] (2D17) {\includegraphics[height=2.8cm,width=0.33\textwidth,trim={0 4.8cm 0 3.7cm},clip]{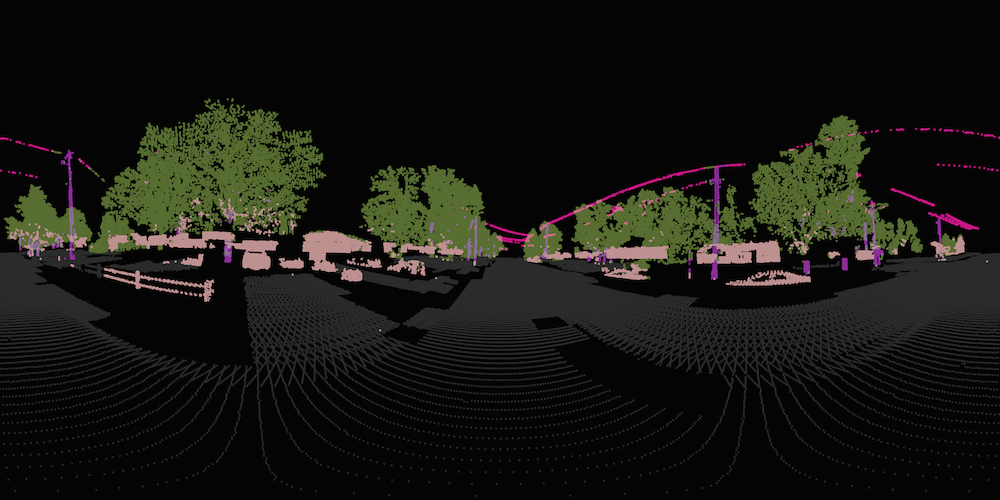}};
		
		\node [right =-.23cm of 2D11] (2D21) {\includegraphics[height=2.8cm,width=0.33\textwidth,trim={0 4.8cm 0 3.7cm},clip]{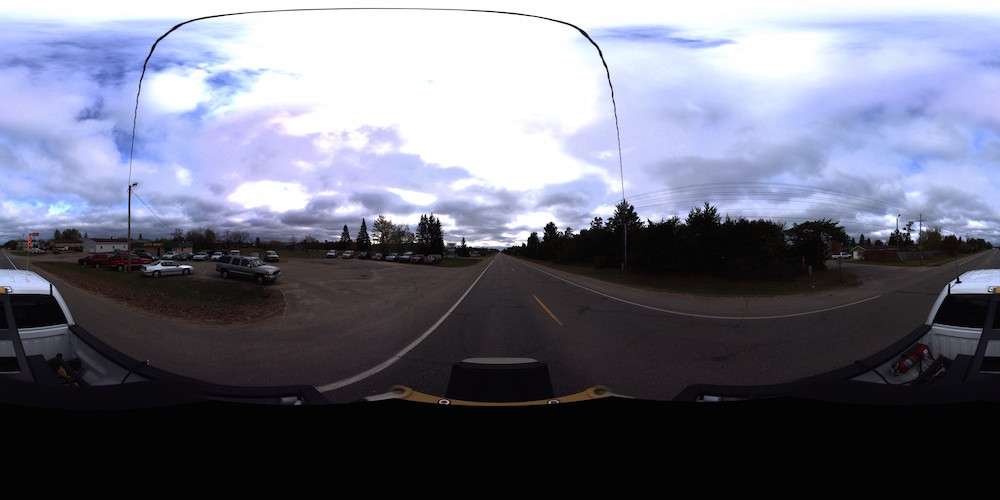}};
		
		\node [below =-.12cm of 2D21] (2D22) {\includegraphics[height=2.8cm,width=0.33\textwidth,trim={0 4.8cm 0 3.7cm},clip]{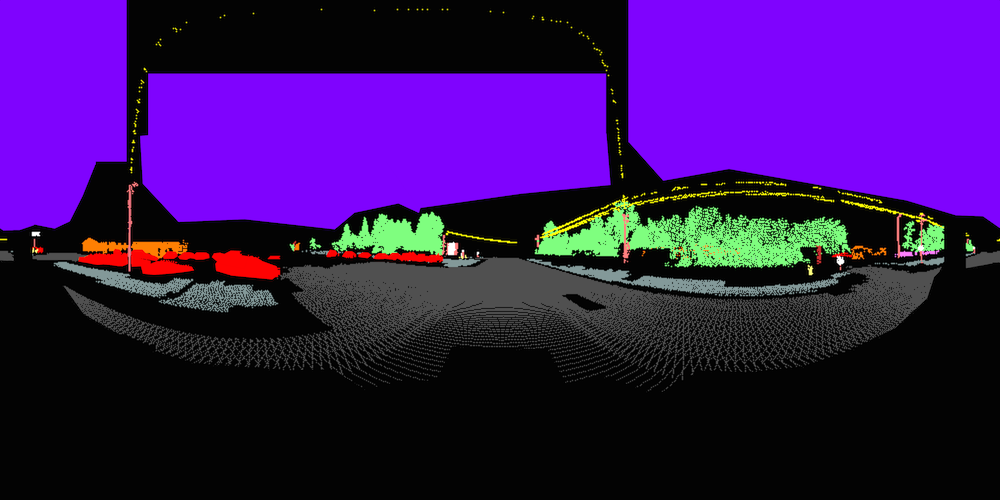}};

		\node [below =-.12cm of 2D22] (2D23) {\includegraphics[height=2.8cm,width=0.33\textwidth,trim={0 4.8cm 0 3.7cm},clip]{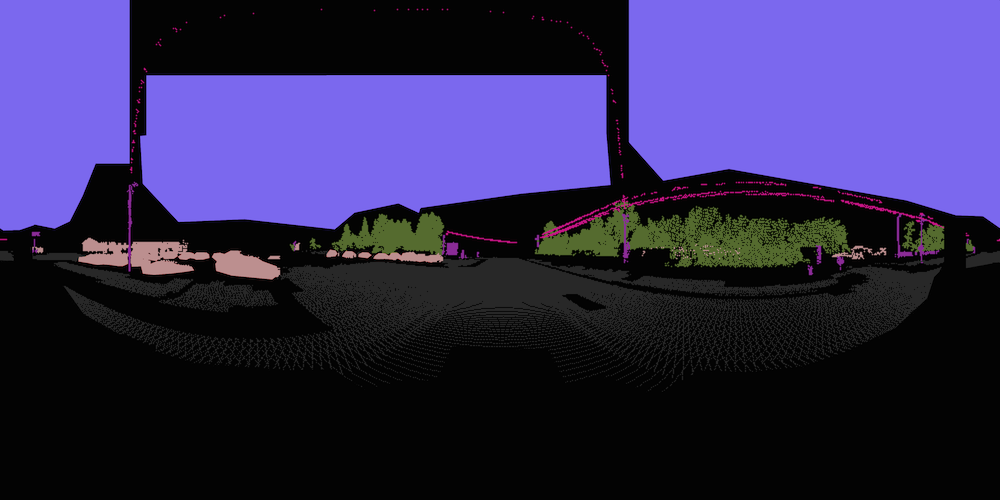}};

		\node [below =-.12cm of 2D23] (2D24) {\includegraphics[height=2.8cm,width=0.33\textwidth,trim={0 18cm 0 14cm},clip]{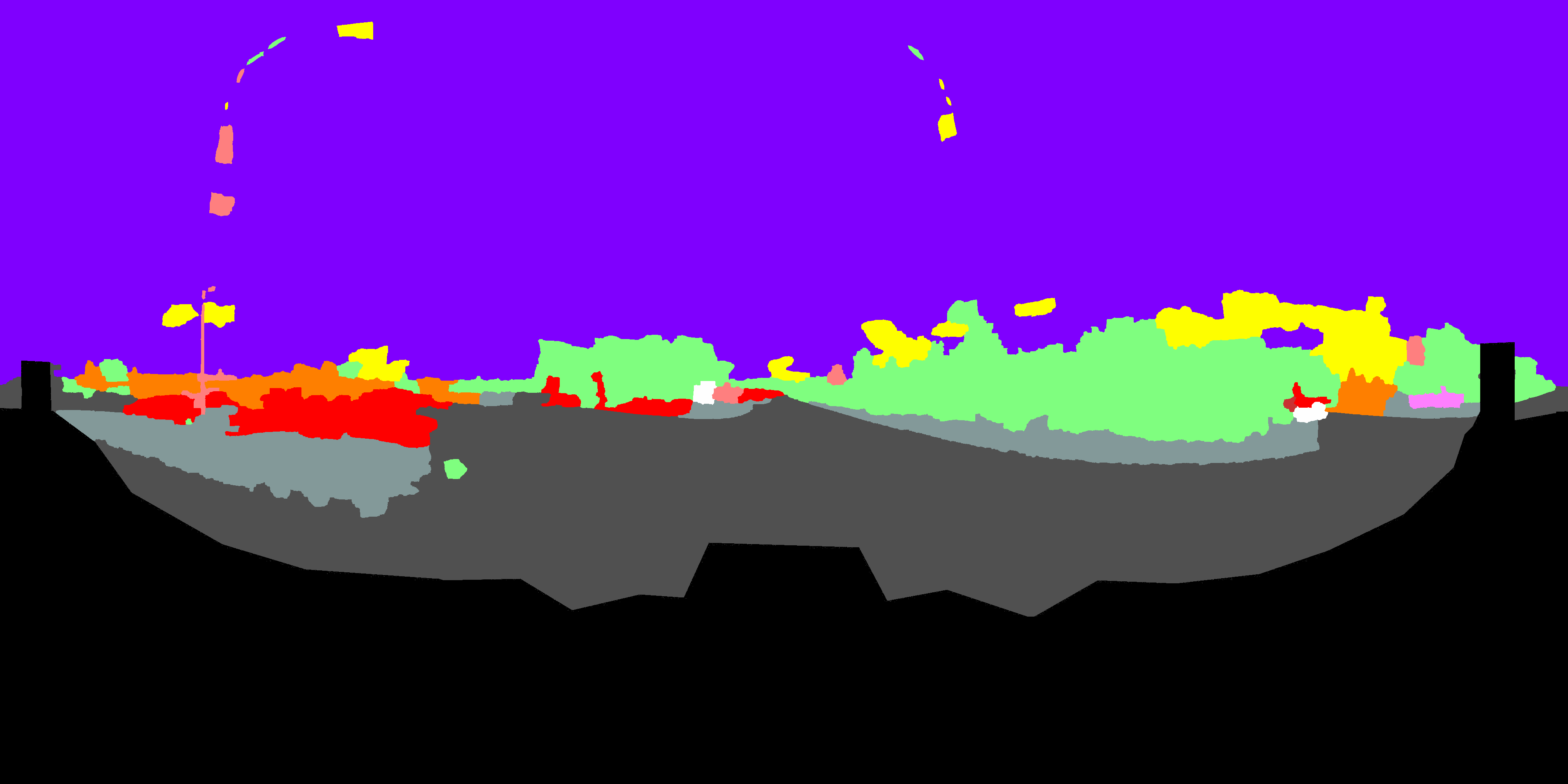}};

		\node [below =-.12cm of 2D24] (2D25) {\includegraphics[height=2.8cm,width=0.33\textwidth,trim={0 18cm 0 14cm},clip]{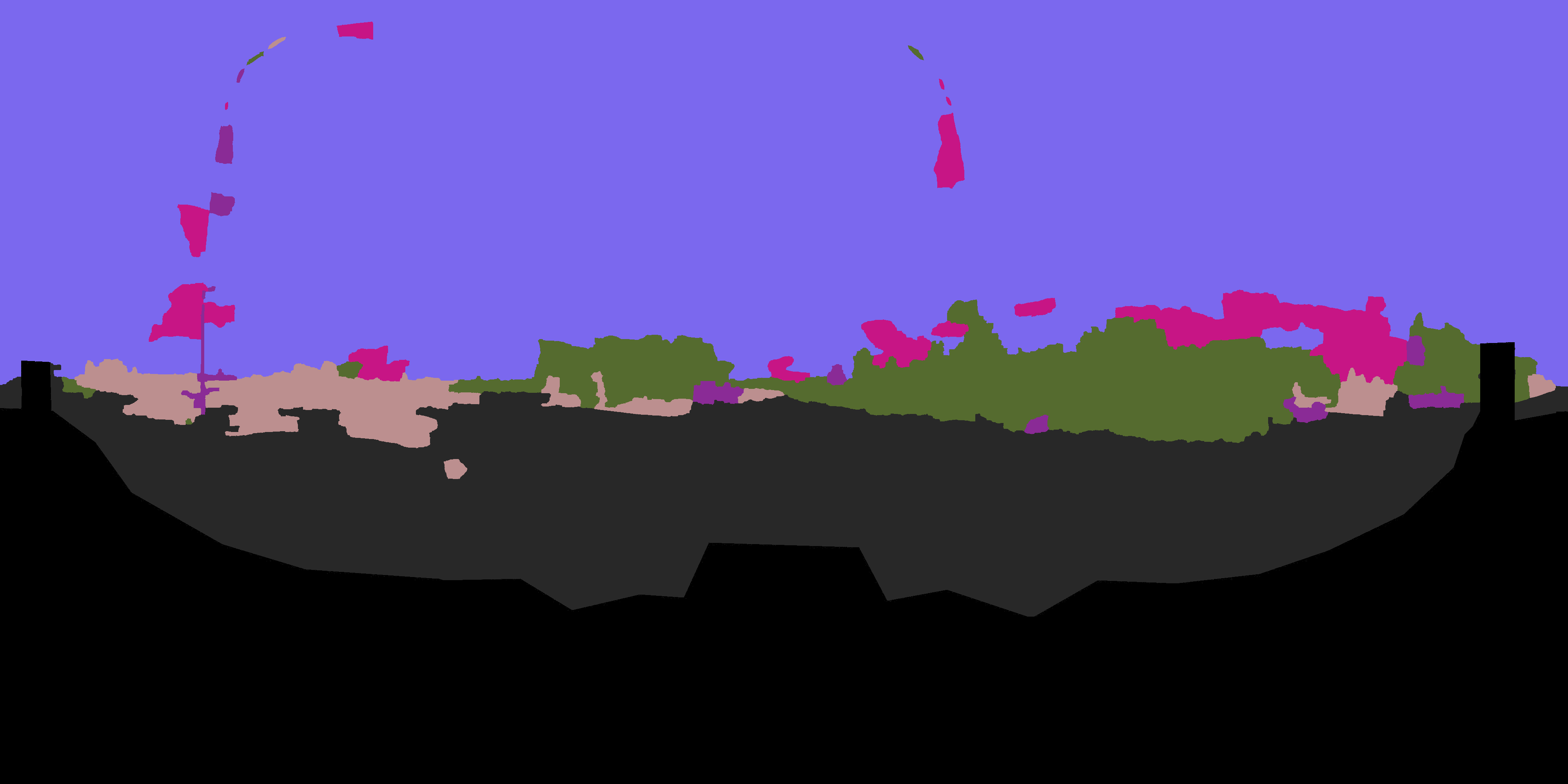}};

		\node [below =-.12cm of 2D25] (2D26) {\includegraphics[height=2.8cm,width=0.33\textwidth,trim={0 4.8cm 0 3.7cm},clip]{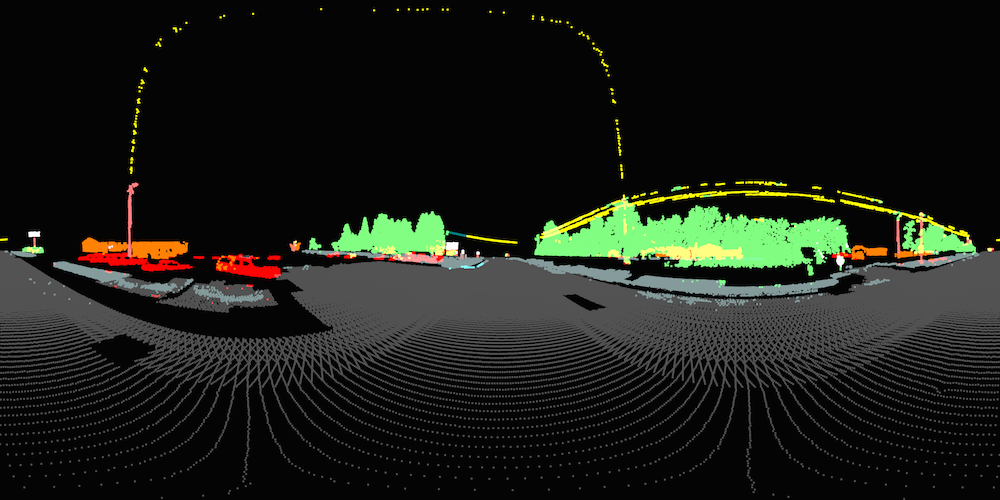}};

		\node [below =-.12cm of 2D26] (2D27) {\includegraphics[height=2.8cm,width=0.33\textwidth,trim={0 4.8cm 0 3.7cm},clip]{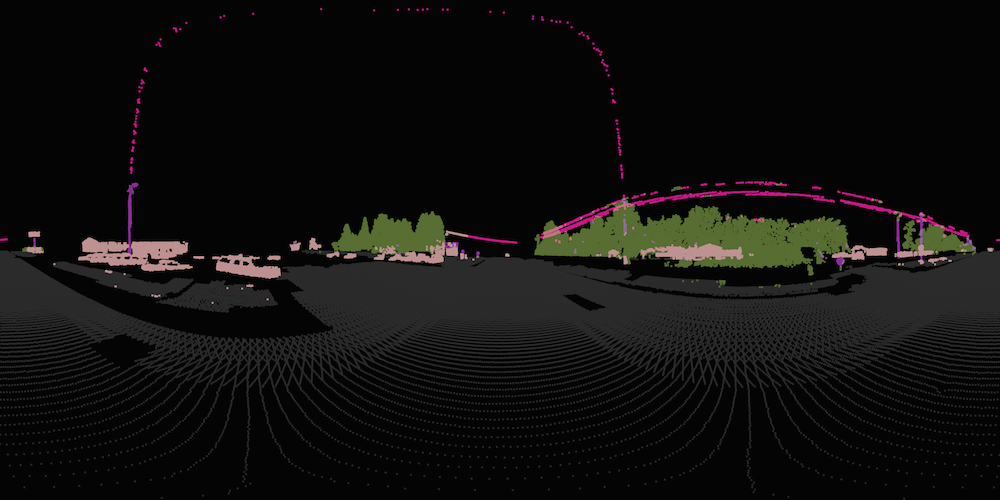}};

		\coordinate [] (ref1) at (.5cm,-18.5cm);
		\node [below =.75cm of ref1,scale=\sc] (b1) [rect1] {Grass};
		\node [right =\dc cm of b1,scale=\sc] (b2) [rect2] {Building};
		\node [right =\dc cm of b2,scale=\sc] (b3) [rect3] {Tree trunk};
		\node [right =\dc cm of b3,scale=\sc] (b4) [rect4] {Tree leaves};
		\node [right =\dc cm of b4,scale=\sc] (b5) [rect5] {Vehicle};
		\node [right =\dc cm of b5,scale=\sc] (b6) [rect6] {Road};
		\node [right =\dc cm of b6,scale=\sc] (b7) [rect7] {Bush};
		\node [right =\dc cm of b7,scale=\sc] (b8) [rect8] {Pole};
		\node [right =\dc cm of b8,scale=\sc] (b9) [rect9] {Sign};
		\node [right =\dc cm of b9,scale=\sc] (b10)  [rect10] {Post};
		\node [right =\dc cm of b10,scale=\sc] (b11) [rect11] {Barrier};
		\node [right =\dc cm of b11,scale=\sc] (b12) [rect12] {Wire};
		\node [right =\dc cm of b12,scale=\sc] (b13) [rect13] {Sidewalk};
		\node [right =\dc cm of b13,scale=\sc] (b14) [rect14] {Sky};
		
		\coordinate [] (ref2) at (.5cm,-2cm);
		\node [below =.25cm of b5,scale=\sc] (c1) [rect21] {\textcolor{white}{Horizontal}};
		\node [right =\dc cm of c1,scale=\sc] (c2) [rect16] {Vertical};
		\node [right =\dc cm of c2,scale=\sc] (c3) [rect17] {Cylindrical};
		\node [right =\dc cm of c3,scale=\sc] (c4) [rect18] {Scattered};
		\node [right =\dc cm of c4,scale=\sc] (c5) [rect19] {Wire};
		\node [right =\dc cm of c5,scale=\sc] (c6) [rect20] {Sky};
		
		\end{tikzpicture}

		\vspace{-.0cm}
		
		\centering
		\caption{Example results of semantic and geometric labeling in the DATA61/2D3D dataset. {\bf1st row:} image, {\bf 2nd row:} 2D semantic ground-truth, {\bf 3rd row:} 2D geometric ground-truth, {\bf4th row:} 2D semantic results, {\bf 5th row:} 2D geometric results, {\bf 6th row:} 3D semantic results, {\bf 7th row:} 3D geometric results.}
		\label{DATA61DatasetResultsFig1}
	\end{figure*}

	Furthermore, the panoramic images in the DATA61/2D3D provide the opportunity of observing an object in successive image frames and as a result, multiple 2D features for each object can be recorded. We linked these corresponding 2D nodes together with latent nodes in each connection to provide more information in the labeling process and gained a 2\% improvement on the 2D performance, as shown in Table~\ref{New_2d_table}. Figure~\ref{DATA61DatasetResultsFig1} shows some sample results of our semantic and geometric labeling on the DATA61/2D3D dataset.
	
	
	\subsection{Results on CMU/VMR}
	The CMU/VMR dataset is comprised of 372 pairs of urban images and corresponding 3D point cloud data, on average 31,000 3D points per image. Importantly, the ground-truth of this data is such that the labels of corresponding 2D and 3D nodes are always the same\footnote{Note that by examining the dataset, one can easily verify that its ground-truth is often erroneous, due to the inaccurate projection and misalignment problem.}. In other words, this dataset is not particularly well-suited to our approach. However, it remains a standard benchmark, and no other dataset, except the DATA61/2D3D dataset, explicitly evidencing the misalignment problem is available. The CMU/VMR dataset contains 19 classes, which yields  the following sizes for the parameter matrices for 2D-3D CRF with latent nodes: $\mathbf{A}_{[19\times28]}^{2D}$, $\mathbf{A}_{[19\times23]}^{3D}$, $\mathbf{A}_{[20\times52]}^{\Delta}$, $\mathbf{B}_{[361\times1]}^{2D}$, $\mathbf{B}_{[361\times1]}^{3D}$, $\mathbf{B}_{[380\times1]}^{2D-\Delta}$ and $\mathbf{B}_{[380\times1]}^{3D-\Delta}$, with alternative matrices for the 2D-3D CRF with no latent nodes of the form $\mathbf{B}_{[361\times1]}^{2D-3D}$, $\mathbf{B}_{[361\times8]}^{2D-3D}$ and $\mathbf{B}_{[361\times52]}^{2D-3D}$. Table~\ref{matrices_table} lists these parameters matrices and describes how their size relates with the selected set of features and class labels in the experiments.
	\vspace{2pt} 
	
\begin{figure*}[t]
		\centering
		\resizebox{18cm}{!}{  
			\subfigure{
				\begin{tikzpicture}
				
				\coordinate [] (ref) at (0,0);
				\node [right =-2cm of ref] (M2true) {\includegraphics[height=2.7cm,width=.16\textwidth]{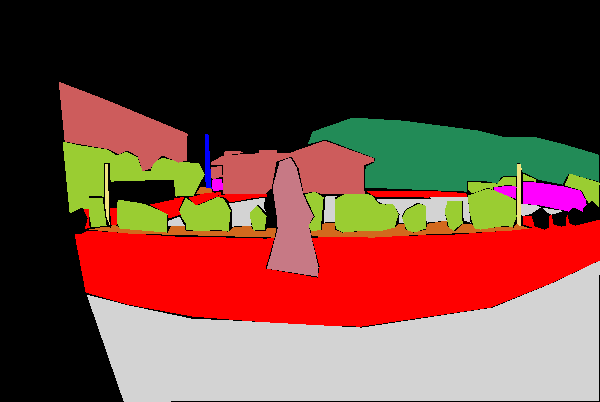}};
				\node [right =-.2cm of M2true] (M2orig) {\includegraphics[height=2.7cm,width=.16\textwidth]{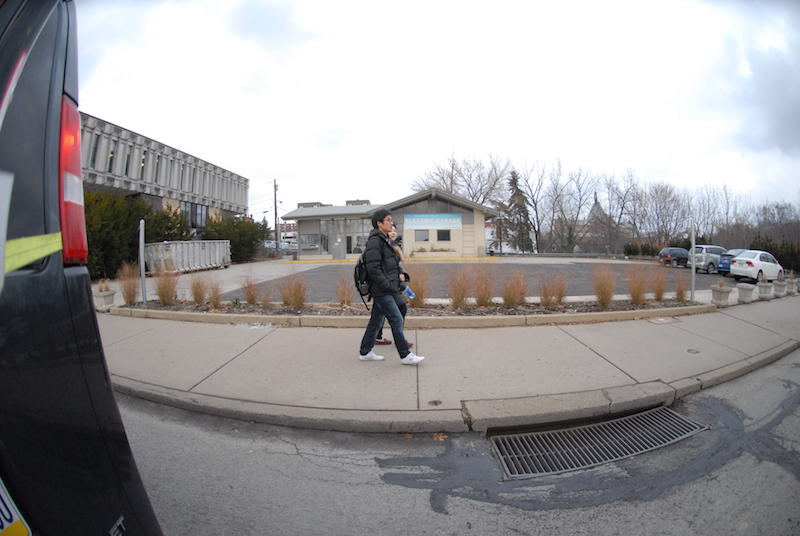}};
				\node [opacity=0.7, right =-.2cm of M2true] (M2result) {\includegraphics[height=2.7cm,width=.16\textwidth]{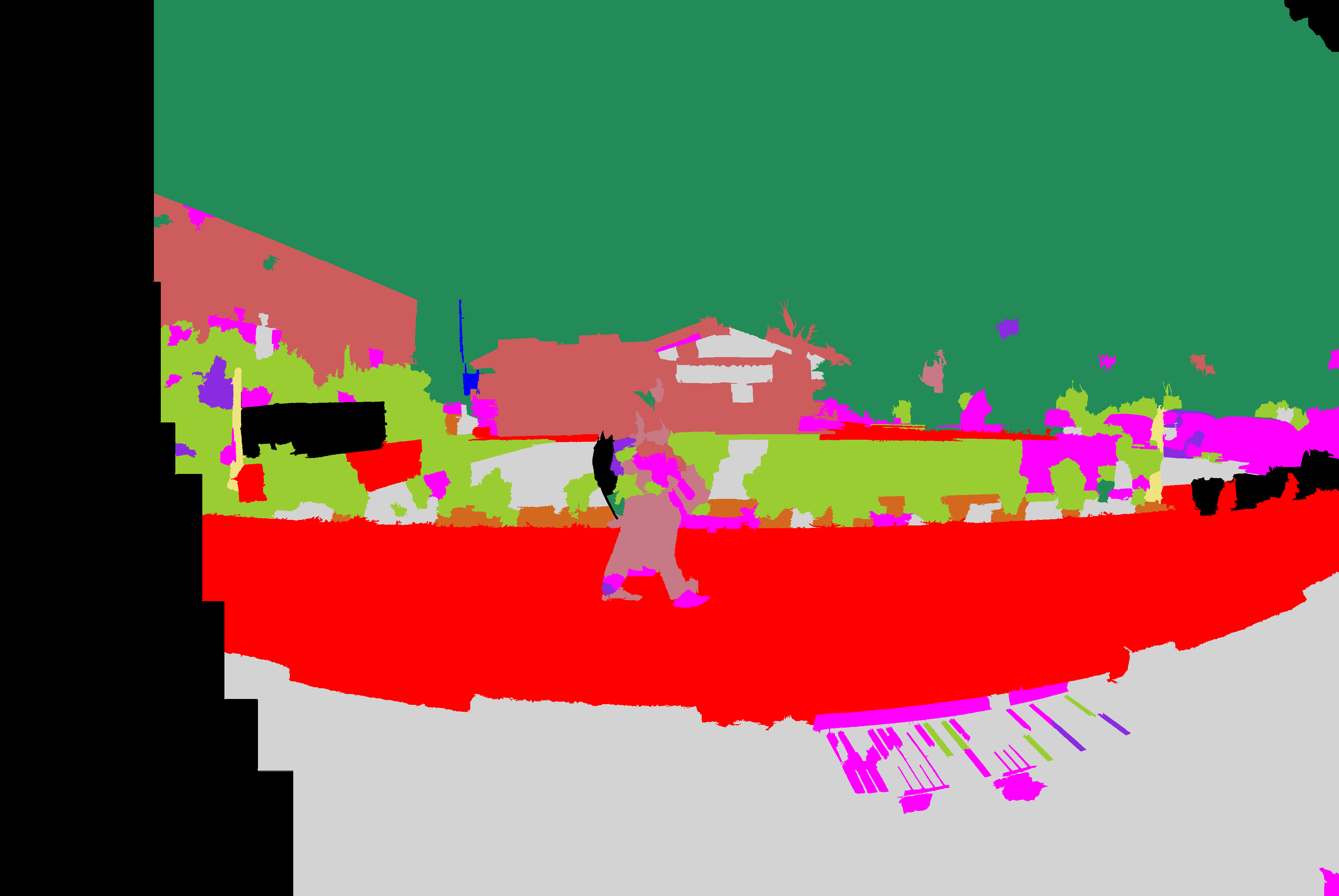}};
				\node [right =-.2cm of M2result] (M2or) {\includegraphics[height=2.7cm,width=.16\textwidth]{result1/Img6062.jpg}};
				\node  [opacity=0.7, right =-.2cm of M2result]  (M2jj) {\includegraphics[height=2.7cm,width=.16\textwidth]{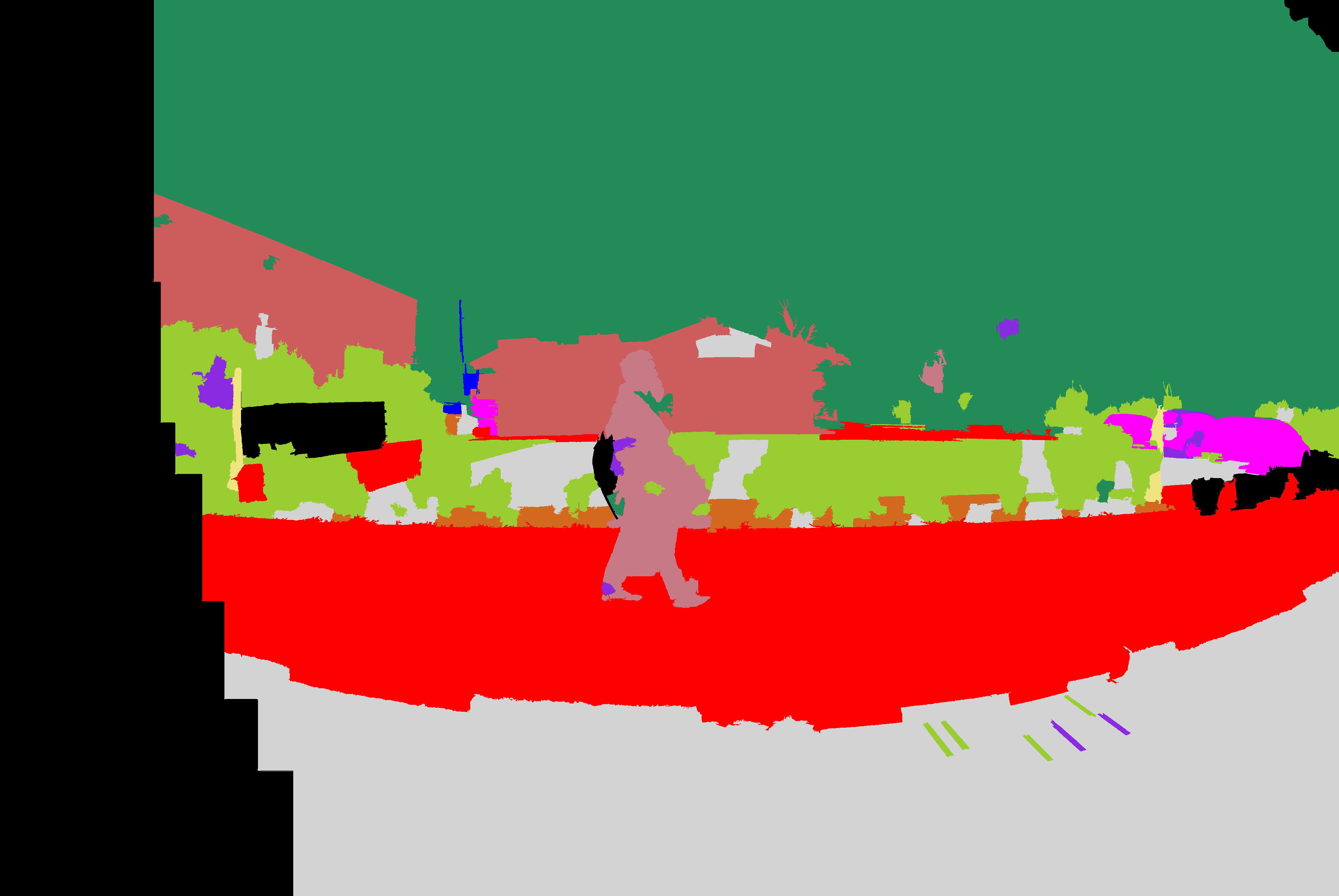}};
				
				\node [below = -.2cm of M2orig] (M1pcdtrue) {\includegraphics[height=2.5cm,width=0.51\textwidth]{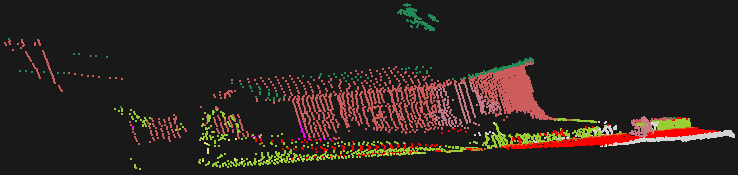}};
				\node [below =-.25cm of M1pcdtrue ] (M1pcdresult) {\includegraphics[height=2.5cm,width=0.51\textwidth]{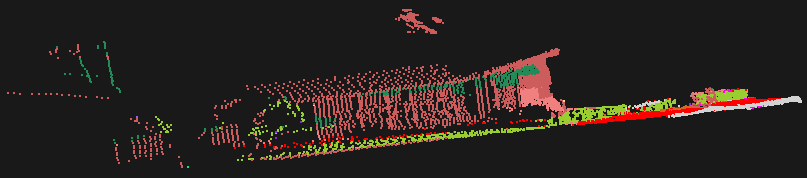}};
				\node [below =-.25cm of M1pcdresult ] (M1pcdresult1) {\includegraphics[height=2.5cm,width=0.51\textwidth]{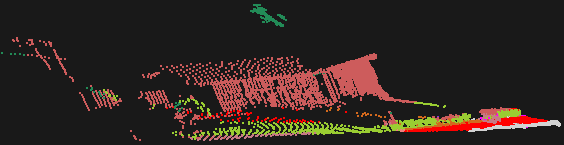}};
				
				\node  [draw,white,thick,circle,minimum size=1cm, right=-2cm of M2orig, below=-1.9cm of M2orig] (D1) {};
				\node  [draw,white,thick,circle,minimum size=1cm, right=-2cm of M2jj, below=-1.9cm of M2jj] (D2) {};
				\node  [draw,white,thick,circle,minimum size=1cm] (D3) at (3.6,-3.0){};
				\node  [draw,white,thick,circle,minimum size=1cm] (D4) at (3.8,-3.0-2.3){};
				\node  [draw,white,thick,circle,minimum size=1cm] (D5) at (3.6,-3.0-2.5-2.5){};

				\node [right =0cm of M2jj] (M2jj2) {\includegraphics[height=2.7cm,width=.16\textwidth]{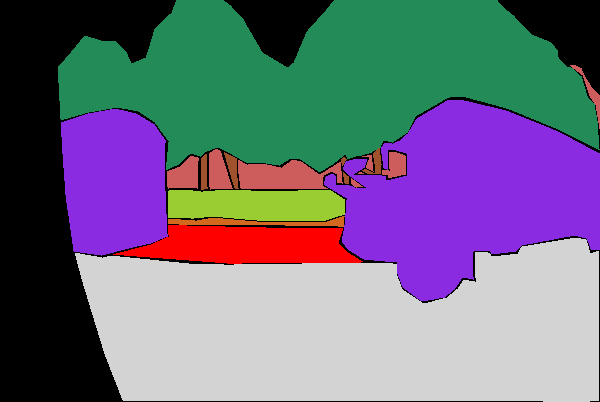}};
				\node [right =-.2cm of M2jj2] (M2orig2) {\includegraphics[height=2.7cm,width=.16\textwidth]{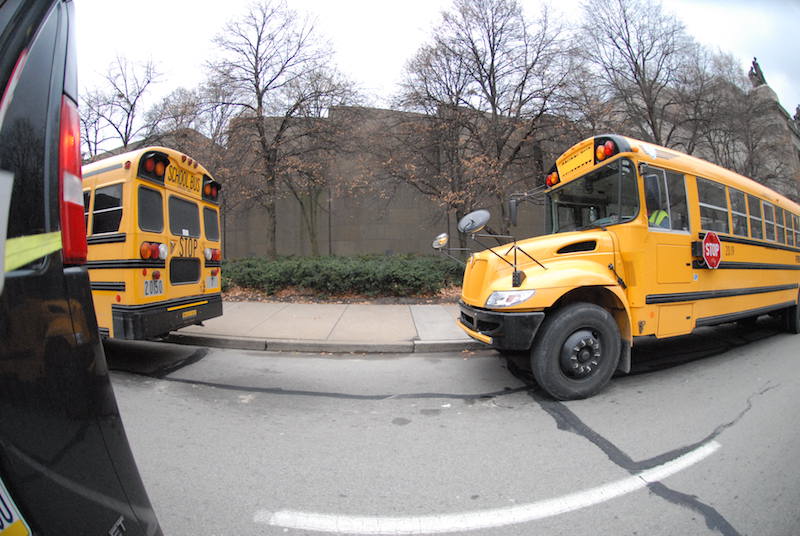}};
				\node [opacity=0.7, right =-.2cm of M2jj2] (M2result2) {\includegraphics[height=2.7cm,width=.16\textwidth]{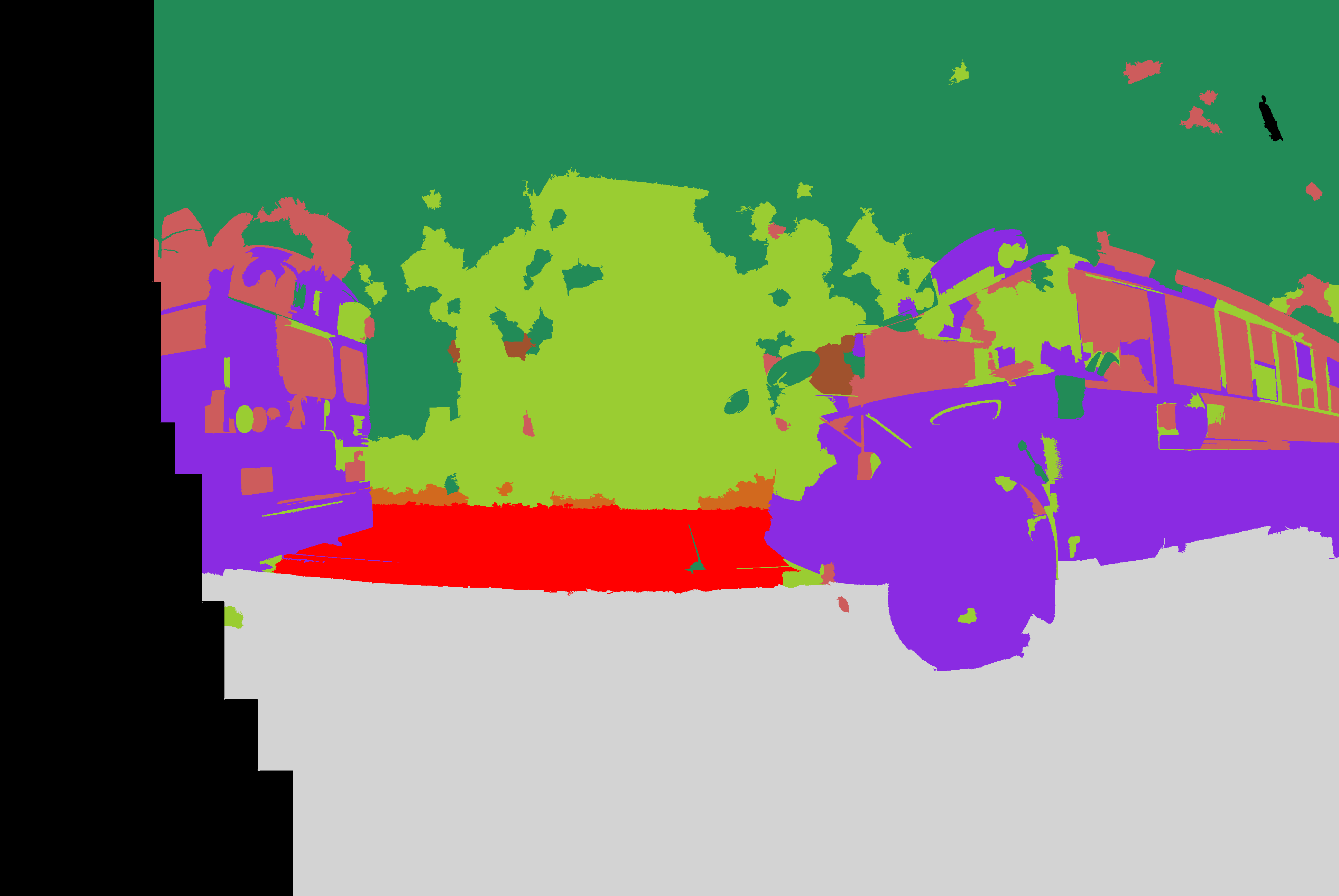}};
				\node [right =-.2cm of M2result2] (M2or2) {\includegraphics[height=2.7cm,width=.16\textwidth]{result1/Img6220.jpg}};
				\node  [opacity=0.7, right =-.2cm of M2result2]  (M2jj2) {\includegraphics[height=2.7cm,width=.16\textwidth]{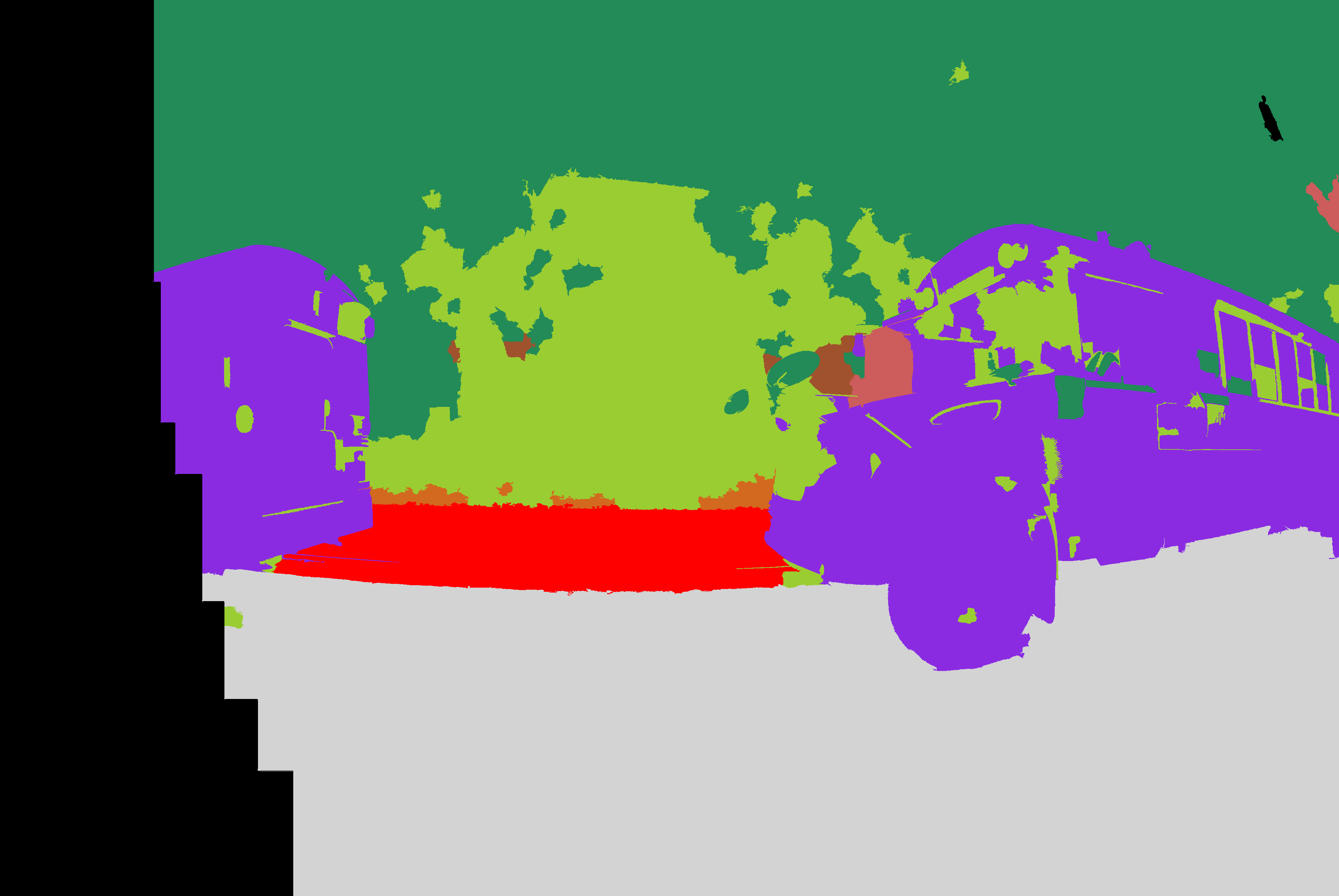}};
				
				\node [right =-.4cm of M1pcdtrue ] (M1pcdtrue2) {\includegraphics[height=2.5cm,width=0.51\textwidth]{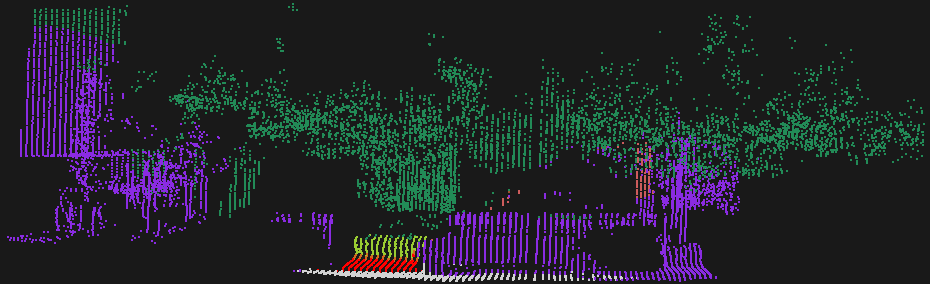}};
				\node [right =-.4cm of M1pcdresult ] (M1pcdresult2) {\includegraphics[height=2.5cm,width=0.51\textwidth]{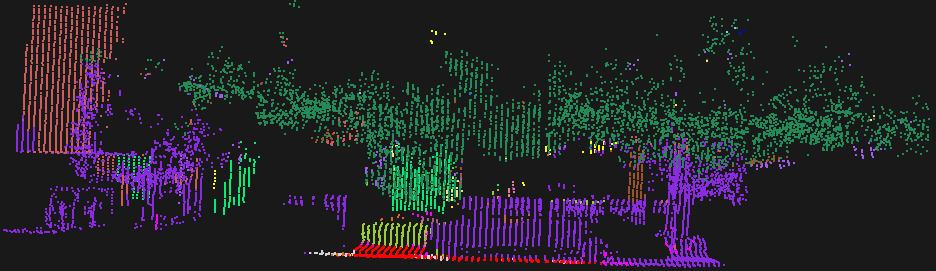}};
				\node [right =-.4cm of M1pcdresult1] (M1pcdresult12) {\includegraphics[height=2.5cm,width=0.51\textwidth]{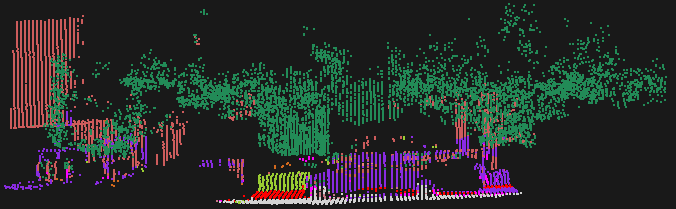}};
				
				\node  [draw,yellow,thick,circle,minimum size=1cm] (D11) at (10.7,.4){};
				\node  [draw,yellow,thick,circle,minimum size=1cm] (D12) at (13.7,.2){};
				\node  [draw,yellow,thick,circle,minimum size=2cm] (D13) at (7.6,-2.8){};
				\node  [draw,yellow,thick,circle,minimum size=2cm] (D14) at (7.6,-2.8-2.3){};
				\node  [draw,yellow,thick,circle,minimum size=2cm] (D15) at (7.6,-2.8-2.5-2.5){};
				

				\end{tikzpicture}
			}
		}
		\scalebox{.7}{
			\subfigure{
				\begin{tikzpicture}[baseline,decoration=brace,scale=.8]
				
				\node [fill=cole1] (hmnn1) at (0,.8) [rect15] {Road};
				\node [fill=cole2] (hmnn2) at (2.6,.8) [rect15] {Sidewalk};
				\node [fill=cole3] (hmnn3) at (5.2,.8) [rect15] {Ground};
				\node [fill=cole4] (hmnn4) at (7.8,.8) [rect15] {Building};
				\node [fill=cole5] (hmnn5) at (10.4,.8) [rect15] {Barrier};
				\node [fill=cole6] (hmnn6) at (13,.8) [rect15] {Bus stop};
				\node [fill=cole7] (hmnn7) at (15.6,.8) [rect15] {Stairs};
				\node [fill=cole8] (hmnn8) at (18.2,.8) [rect15] {Shrub};
				\node [fill=cole9] (hmnn9) at (20.8,.8) [rect15] {Tree trunk};
				\node [fill=cole10] (hmnn10) at (23.4,.8) [rect15] {Tree top};
				\node [fill=cole11] (hmnn11) at (0,0) [rect15] {Small vehicle};
				\node [fill=cole12] (hmnn12) at (2.6,0) [rect15] {Big vehicle};
				\node [fill=cole13] (hmnn13) at (5.2,0) [rect15] {Person};
				\node [fill=cole14] (hmnn14) at (7.8,0) [rect15] {Tall light};
				\node [fill=cole15] (hmnn15) at (10.4,0) [rect15] {Post};
				\node [fill=cole16] (hmnn16) at (13,0) [rect15] {Sign};
				\node [fill=cole17] (hmnn17) at (15.6,0) [rect15] {Utility pole};
				\node [fill=cole18] (hmnn18) at (18.2,0) [rect15] {Wire};
				\node [fill=cole19] (hmnn19) at (20.8,0) [rect15] {Traffic signal};
				
				\end{tikzpicture}
			}
		}
		
		\centering
		\caption{Sample results of two scenes in the CMU/VMR dataset. {\bf1st row in each scene:} {\bf Left:} 2D ground-truth; {\bf Middle:} the results of~\cite{wacv2015}; {\bf Right:} our 2D results. {\bf2nd row:} ground-truth of the 3D data; {\bf3rd row:} the results of~\cite{wacv2015}; {\bf4th row:} our 3D results. The circles highlight mislabeling in the 3D ground-truth of this dataset, which happened because of misalignments between 2D images and 3D data, and indicate how our approach has enhanced the results in those challenging parts compared to~\cite{wacv2015}.}
		
		\label{MunozFig}
	\end{figure*}

	We compare the results of the 2D-3D CRF model with handcrafted and learned potentials and also with latent nodes and no latent nodes in Table~\ref{Munoz_2d_table} and Table~\ref{Munoz_3d_table} for the 2D and 3D domains, respectively. In this case, while our approach still yields the best F1-scores on average, there is less difference between our results with latent nodes and the no latent method. This can easily be explained by the fact that, as mentioned above, the ground-truth labels of corresponding nodes in 2D and 3D are always the same. In addition, it can be seen that our method does not perform very well on the rare categories with insufficient number of training samples, e.g. the last five classes in the tables. This outcome is not surprising though, since our training strategy heavily relies on the training data. Figure~\ref{MunozFig} demonstrates a qualitative comparison.
	
\begin{figure*}[b]
		\centering
		\def\sc{.7}
		\def\dc{-.05}
		\begin{tikzpicture}
		\coordinate [] (ref33) at (-.1cm,0);
		\node [right =-.52cm of ref33] (2D1) {\includegraphics[height=2.5cm,width=0.25\textwidth]{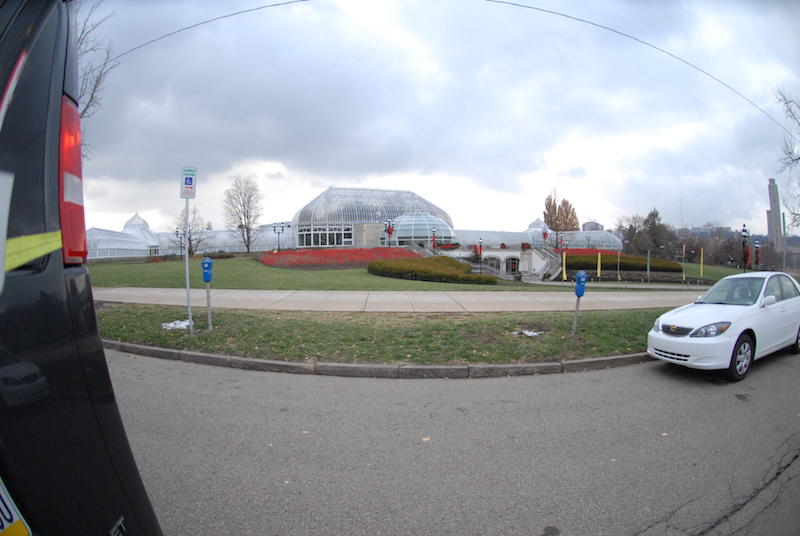}};

		\node [below =-.16cm of 2D1] (2D2) {\includegraphics[height=2.5cm,width=0.25\textwidth]{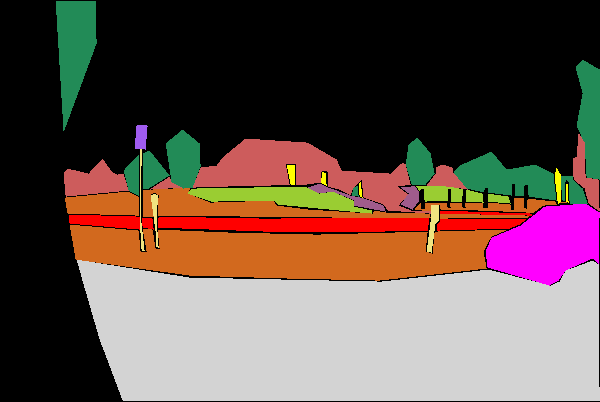}};
		
		\node [below =-.16cm of 2D2] (2D3) {\includegraphics[height=2.5cm,width=0.25\textwidth]{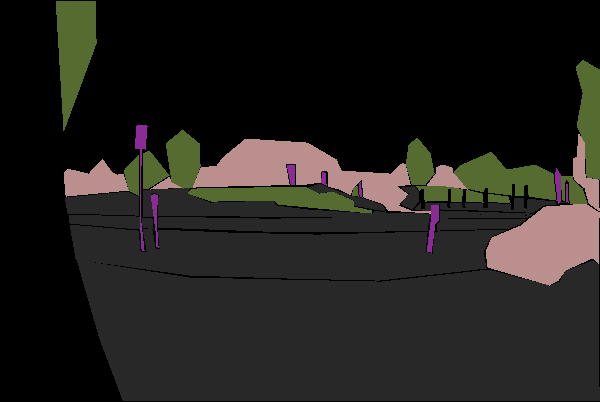}};
		
		\node [below =-.16cm of 2D3] (2D4) {\includegraphics[height=2.5cm,width=0.25\textwidth]{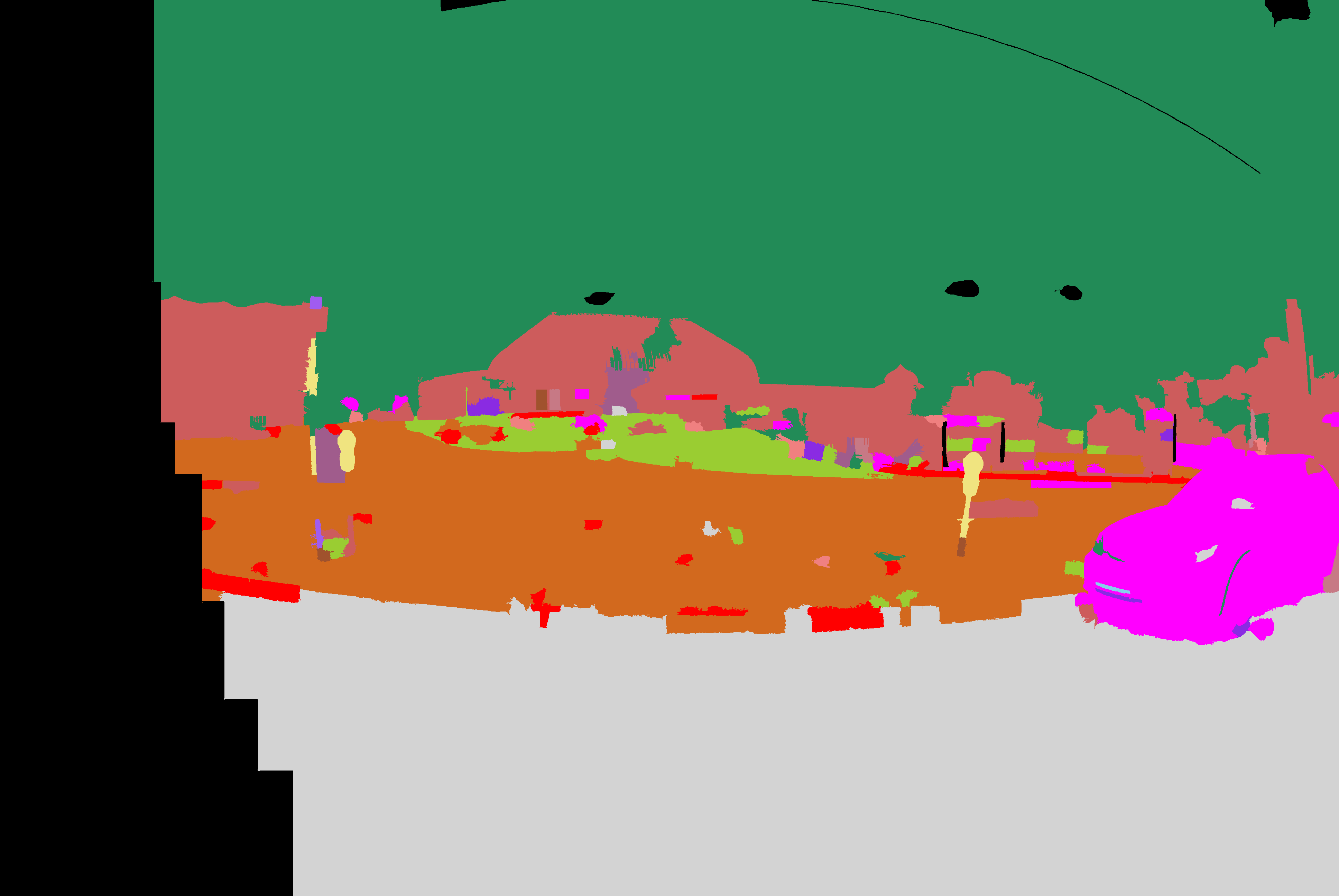}};
		
		\node [below =-.16cm of 2D4] (2D5) {\includegraphics[height=2.5cm,width=0.25\textwidth]{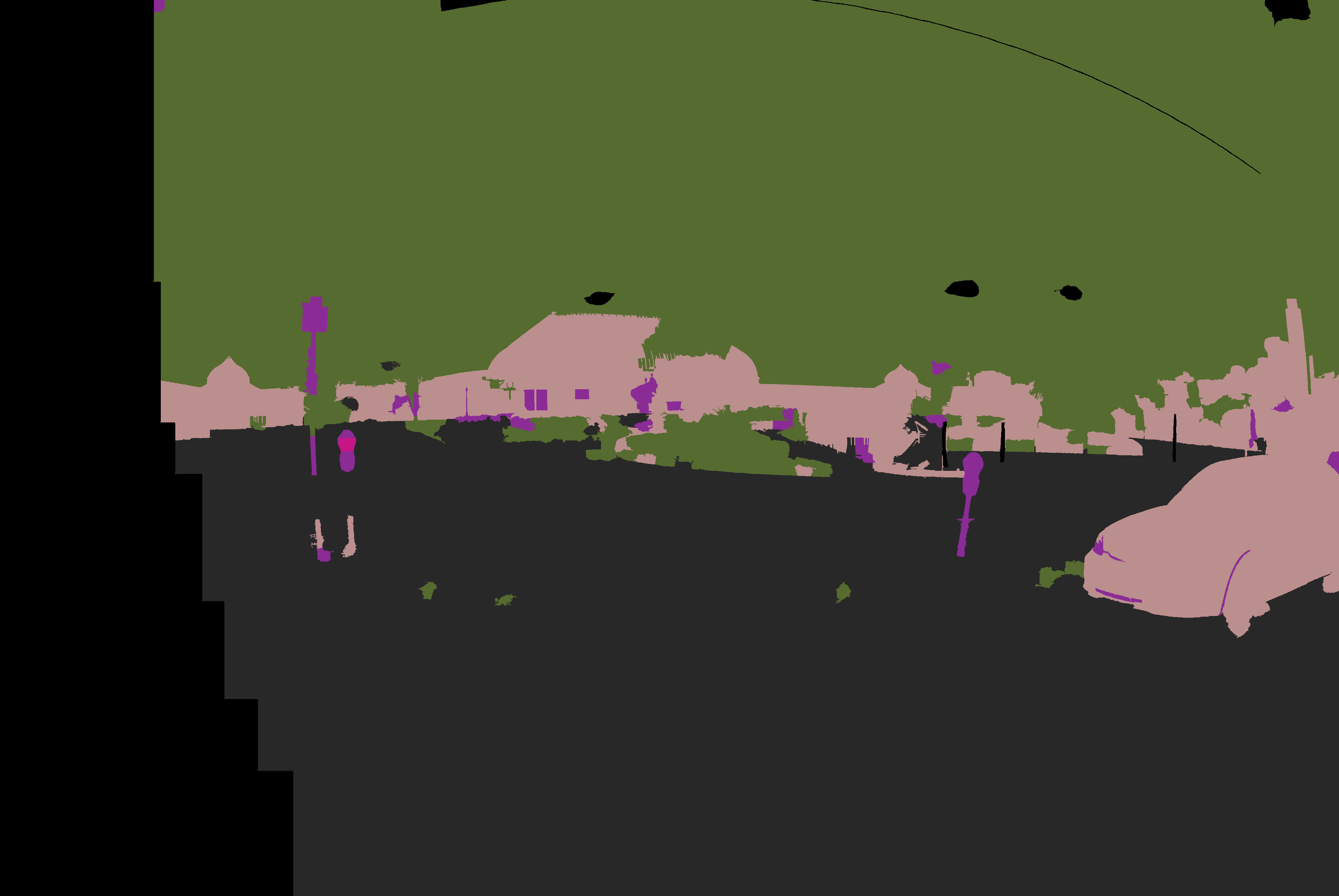}};
		
		\node [below =-.16cm of 2D5] (2D6) {\includegraphics[height=2.5cm,width=0.25\textwidth]{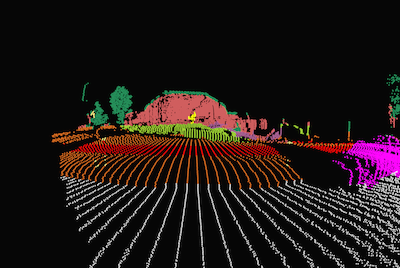}};
		
		\node [below =-.16cm of 2D6] (2D7) {\includegraphics[height=2.5cm,width=0.25\textwidth]{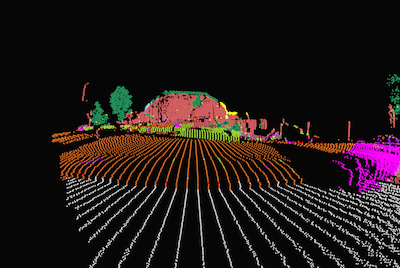}};
		
		\node [below =-.16cm of 2D7] (2D8) {\includegraphics[height=2.5cm,width=0.25\textwidth]{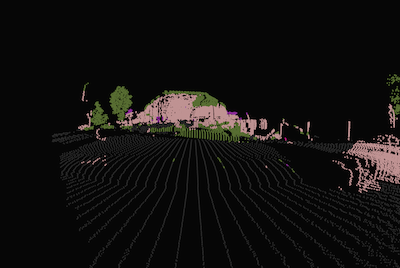}};

		\node [right =-.13cm of 2D1] (2D11) {\includegraphics[height=2.5cm,width=0.25\textwidth]{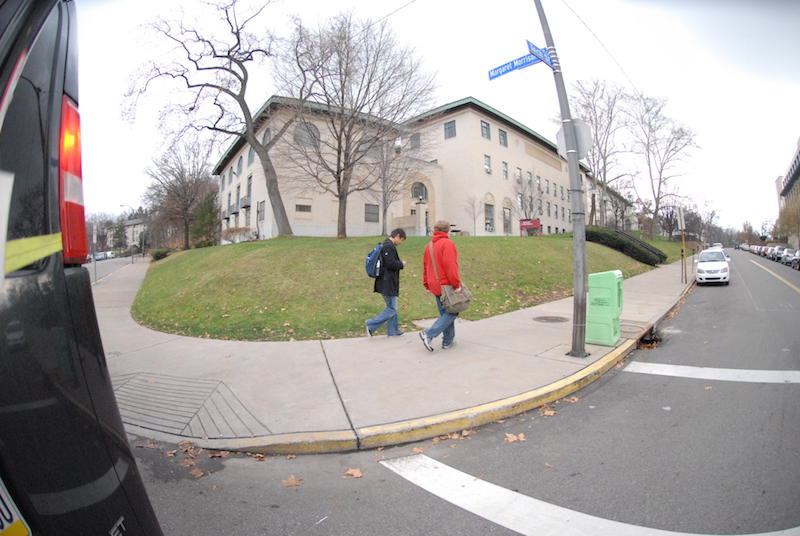}};
		\node [below =-.16cm of 2D11] (2D12) {\includegraphics[height=2.5cm,width=0.25\textwidth]{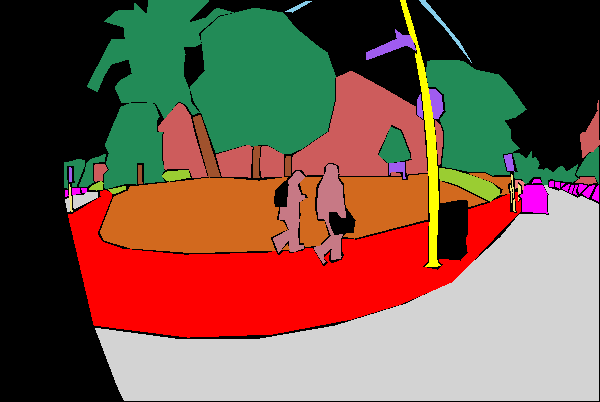}};
		
		\node [below =-.16cm of 2D12] (2D13) {\includegraphics[height=2.5cm,width=0.25\textwidth]{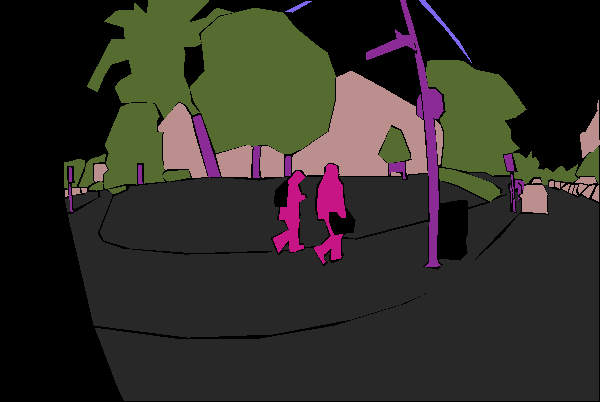}};
		
		\node [below =-.16cm of 2D13] (2D14) {\includegraphics[height=2.5cm,width=0.25\textwidth]{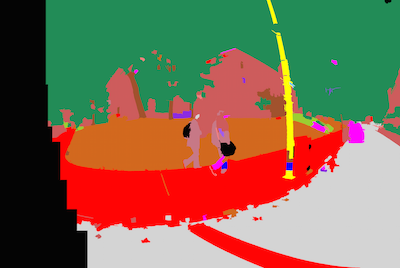}};
		
		\node [below =-.16cm of 2D14] (2D15) {\includegraphics[height=2.5cm,width=0.25\textwidth]{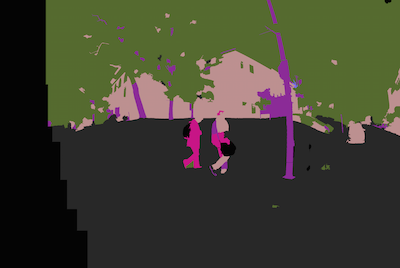}};

		\node [below =-.16cm of 2D15] (2D16) {\includegraphics[height=2.5cm,width=0.25\textwidth]{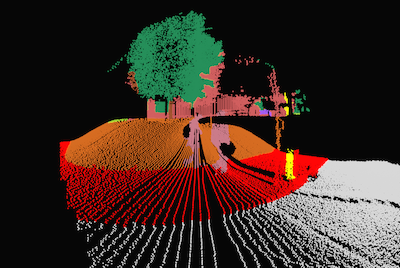}};
		
		\node [below =-.16cm of 2D16] (2D17) {\includegraphics[height=2.5cm,width=0.25\textwidth]{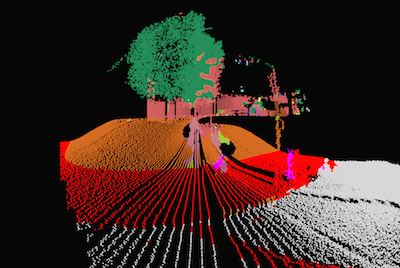}};
		
		\node [below =-.16cm of 2D17] (2D18) {\includegraphics[height=2.5cm,width=0.25\textwidth]{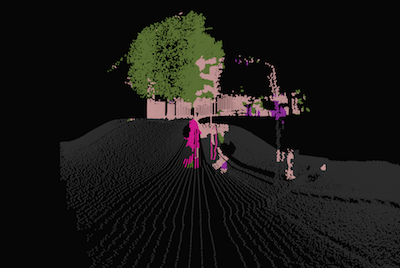}};

		\node [right =-.13cm of 2D11] (2D21) {\includegraphics[height=2.5cm,width=0.25\textwidth]{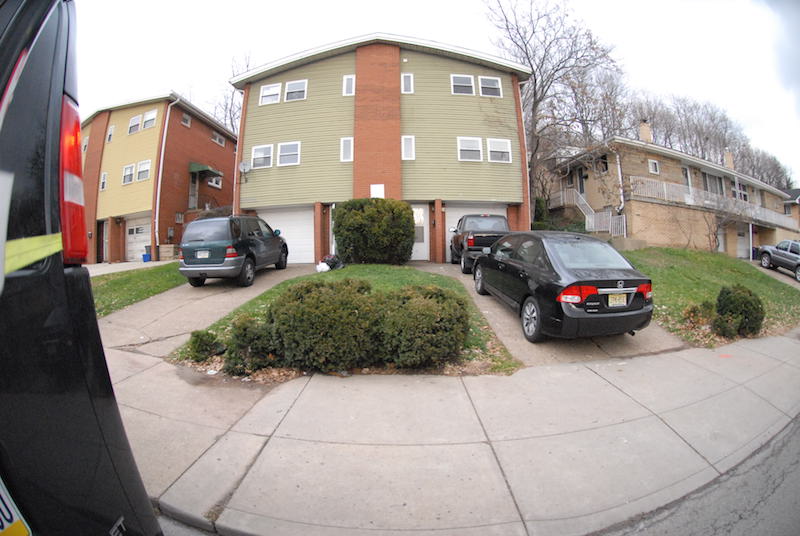}};
		\node [below =-.16cm of 2D21] (2D22) {\includegraphics[height=2.5cm,width=0.25\textwidth]{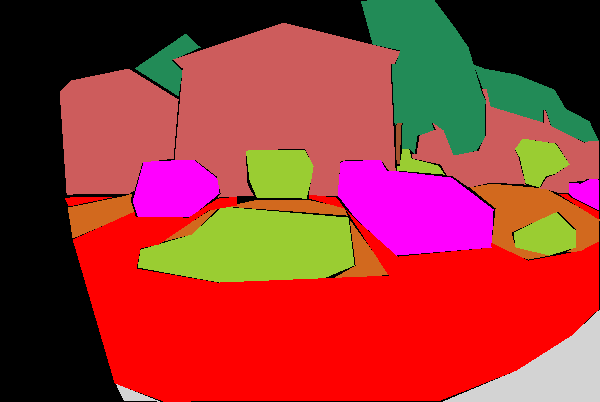}};
		
		\node [below =-.16cm of 2D22] (2D23) {\includegraphics[height=2.5cm,width=0.25\textwidth]{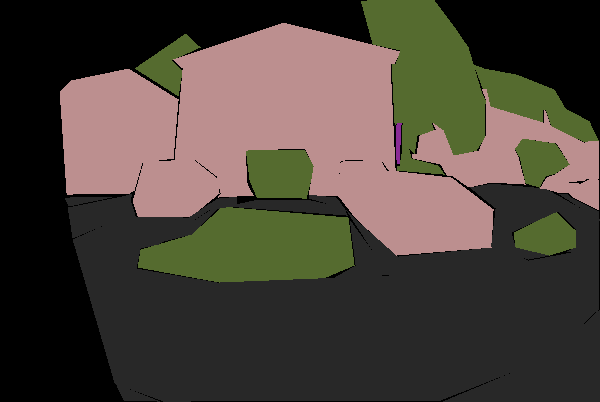}};
		
		\node [below =-.16cm of 2D23] (2D24) {\includegraphics[height=2.5cm,width=0.25\textwidth]{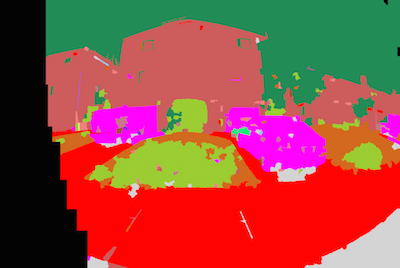}};
		
		\node [below =-.16cm of 2D24] (2D25) {\includegraphics[height=2.5cm,width=0.25\textwidth]{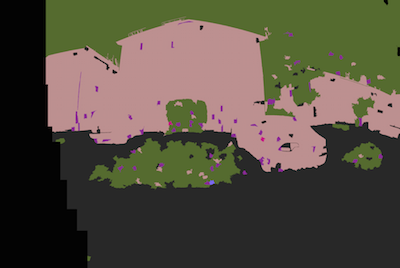}};
		
		\node [below =-.16cm of 2D25] (2D26) {\includegraphics[height=2.5cm,width=0.25\textwidth]{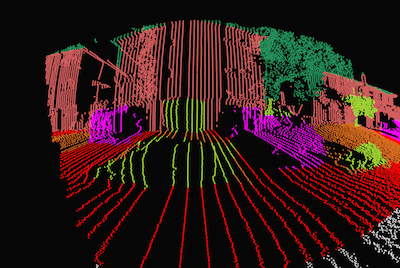}};
		
		\node [below =-.16cm of 2D26] (2D27) {\includegraphics[height=2.5cm,width=0.25\textwidth]{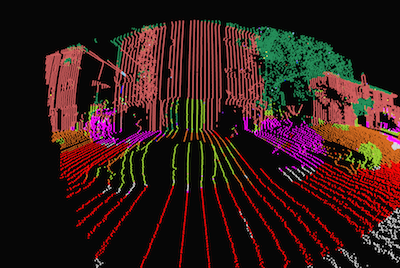}};
		
		\node [below =-.16cm of 2D27] (2D28) {\includegraphics[height=2.5cm,width=0.25\textwidth]{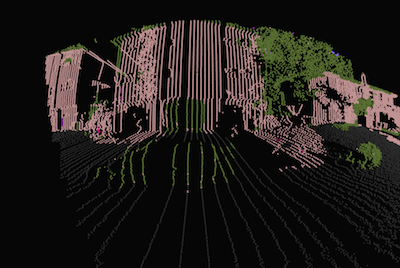}};

		\node [right =-.13cm of 2D21] (2D31) {\includegraphics[height=2.5cm,width=0.25\textwidth]{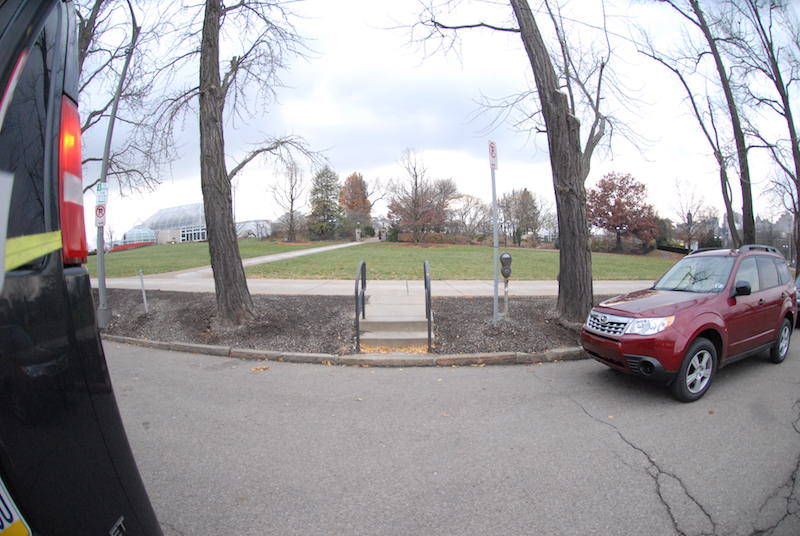}};
		\node [below =-.16cm of 2D31] (2D32) {\includegraphics[height=2.5cm,width=0.25\textwidth]{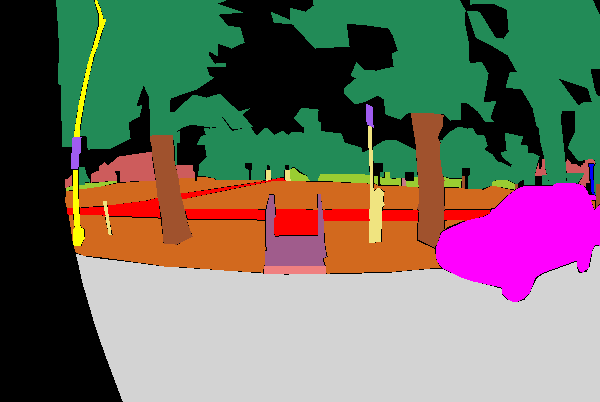}};
		
		\node [below =-.16cm of 2D32] (2D33) {\includegraphics[height=2.5cm,width=0.25\textwidth]{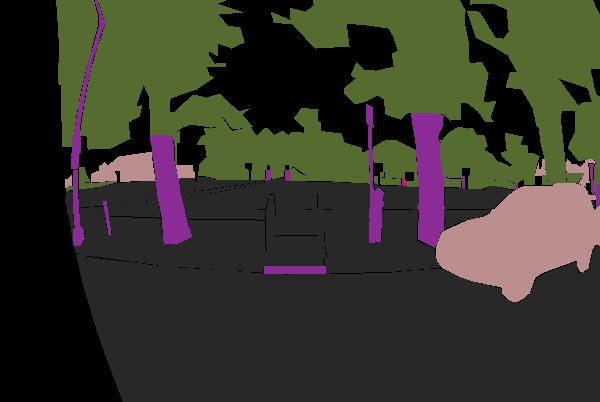}};
		
		\node [below =-.16cm of 2D33] (2D34) {\includegraphics[height=2.5cm,width=0.25\textwidth]{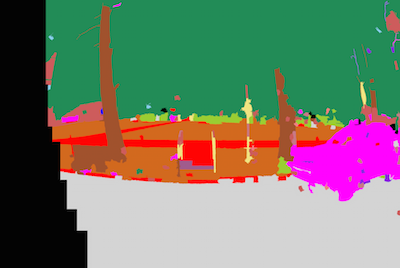}};
		
		\node [below =-.16cm of 2D34] (2D35) {\includegraphics[height=2.5cm,width=0.25\textwidth]{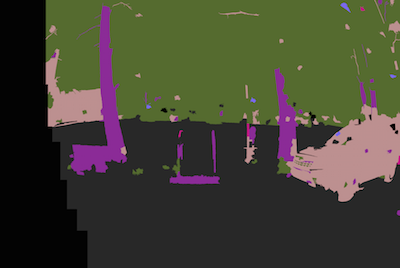}};
		
		\node [below =-.16cm of 2D35] (2D36) {\includegraphics[height=2.5cm,width=0.25\textwidth]{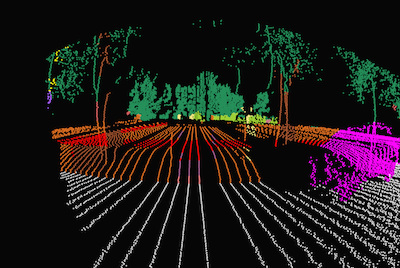}};
		
		\node [below =-.16cm of 2D36] (2D37) {\includegraphics[height=2.5cm,width=0.25\textwidth]{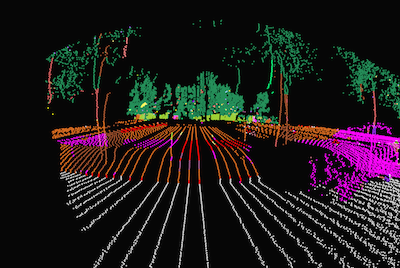}};
		
		\node [below =-.16cm of 2D37] (2D38) {\includegraphics[height=2.5cm,width=0.25\textwidth]{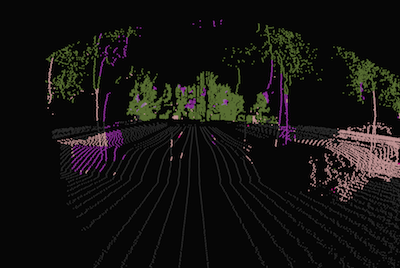}};

		\end{tikzpicture}
		
		\scalebox{.7}{
			\subfigure{
				\begin{tikzpicture}[baseline,decoration=brace,scale=.8]
				
				\node [fill=cole1] (hmnn1) at (0,.8) [rect15] {Road};
				\node [fill=cole2] (hmnn2) at (2.6,.8) [rect15] {Sidewalk};
				\node [fill=cole3] (hmnn3) at (5.2,.8) [rect15] {Ground};
				\node [fill=cole4] (hmnn4) at (7.8,.8) [rect15] {Building};
				\node [fill=cole5] (hmnn5) at (10.4,.8) [rect15] {Barrier};
				\node [fill=cole6] (hmnn6) at (13,.8) [rect15] {Bus stop};
				\node [fill=cole7] (hmnn7) at (15.6,.8) [rect15] {Stairs};
				\node [fill=cole8] (hmnn8) at (18.2,.8) [rect15] {Shrub};
				\node [fill=cole9] (hmnn9) at (20.8,.8) [rect15] {Tree trunk};
				\node [fill=cole10] (hmnn10) at (23.4,.8) [rect15] {Tree top};
				\node [fill=cole11] (hmnn11) at (0,0) [rect15] {Small vehicle};
				\node [fill=cole12] (hmnn12) at (2.6,0) [rect15] {Big vehicle};
				\node [fill=cole13] (hmnn13) at (5.2,0) [rect15] {Person};
				\node [fill=cole14] (hmnn14) at (7.8,0) [rect15] {Tall light};
				\node [fill=cole15] (hmnn15) at (10.4,0) [rect15] {Post};
				\node [fill=cole16] (hmnn16) at (13,0) [rect15] {Sign};
				\node [fill=cole17] (hmnn17) at (15.6,0) [rect15] {Utility pole};
				\node [fill=cole18] (hmnn18) at (18.2,0) [rect15] {Wire};
				\node [fill=cole19] (hmnn19) at (20.8,0) [rect15] {Traffic signal};

				\end{tikzpicture}
			}
		}
		\begin{tikzpicture}
		\coordinate [] (ref2) at (.5cm,-2cm);
		\node [below =.25cm of ref2,scale=\sc] (c11) [rect21] {\textcolor{white}{Horizontal}};
		\node [right =\dc cm of c11,scale=\sc] (c12) [rect16] {Vertical};
		\node [right =\dc cm of c12,scale=\sc] (c13) [rect17] {Cylindrical};
		\node [right =\dc cm of c13,scale=\sc] (c14) [rect18] {Scattered};
		\node [right =\dc cm of c14,scale=\sc] (c15) [rect19] {Person};
		\node [right =\dc cm of c15,scale=\sc] (c16) [rect20] {Wire};
		\end{tikzpicture}
		\centering
		\caption{Example results of semantic and geometric labeling in the CMU/VMR dataset. {\bf1st row:} image, {\bf 2nd row:} 2D semantic ground-truth, {\bf 3rd row:} 2D geometric ground-truth, {\bf 4th row:} 2D semantic results, {\bf 5th row:} 2D geometric results, {\bf 6th row:} 3D semantic ground-truth, {\bf 7th row:} 3D semantic results, {\bf 8th row:} 3D geometric results.}
		
		\label{MunozFig22}
	\end{figure*}

	Six geometric classes are considered in the CMU/VMR dataset (Table~\ref{Semantic_geometric}).
	Similarly to the DATA61/2D3D dataset, the 2D and 3D geometric data are augmented to the semantic model as two separate data modalities and their simultaneous inference is carried out given the semantic and geometric cues of the 2D and 3D data. Tables~\ref{Munoz_2d_table} and \ref{Munoz_3d_table} demonstrate the results of the 2D and 3D semantic scene parsing using the proposed semantic and geometric 2D/3D multimodal model. It improves the F1-scores of the 2D and 3D data.  The results of the geometric labeling of the 2D and 3D data are shown in Table~\ref{DATA61_geometry_table1}. Figure~\ref{MunozFig22} shows some sample results of our semantic and geometric labeling on the CMU/VMR dataset.
	%
	%
	
	\begin{table*}[hp]\centering\renewcommand{\arraystretch}{1.2}
		\caption{ Training and inference time for DATA61/2D3D and CMU/VMR datasets. }
		\vspace{-15pt}
		\def\d{4cm}
		\def\r{30}
		
		{
			\begin{tabular}{c+c+c+c+c}\\ 
				{}&{\bf \scriptsize Training time}&{\bf \scriptsize Inference time}&{\bf \scriptsize Training time}&{\bf \scriptsize Inference time}\\ 
				
				{}&{\bf \scriptsize (DATA61/2D3D dataset)}&{\bf \scriptsize (DATA61/2D3D dataset)}&{\bf \scriptsize (CMU/VMR dataset)}&{\bf \scriptsize (CMU/VMR dataset)}\\ \thickhline
				
				{2D-3D CRF with latent nodes}&\multicolumn{1}{|c}{6hr45min}&\multicolumn{1}{|c}{0.85s}&\multicolumn{1}{|c}{4hr40min}&\multicolumn{1}{|c}{0.47s}\\
				\thickhline
				
				{Simultaneous Inference of Semantic and }&\multicolumn{1}{|c}{19hr20min}&\multicolumn{1}{|c}{2.3s}&\multicolumn{1}{|c}{24hr15min}&\multicolumn{1}{|c}{1.2s}\\
				
				{Geometric Classes both in 2D and 3D}&\multicolumn{1}{|c}{}&\multicolumn{1}{|c}{}&\multicolumn{1}{|c}{}&\multicolumn{1}{|c}{}\\
				\thickhline
				
			\end{tabular}
		}
		
		\label{time1}
	\end{table*}

	\begin{table*}[hp]\centering\renewcommand{\arraystretch}{1.2}
		\caption{ Mapping table between the geometric and semantic classes for DATA61/2D3D dataset and CMU/VMR dataset. }
		\vspace{-15pt}
		\def\d{4cm}
		\def\r{30}
		{
			\begin{tabular}{c+c+c}\\ 
				{\bf \scriptsize Geometric Classes}&{\bf \scriptsize Semantic Classes  (DATA61/2D3D dataset)}&{\bf \scriptsize Semantic Classes  (CMU/VMR dataset)}\\ \thickhline
				
				{Horizontal Plane}&\multicolumn{1}{|c}{Grass - Road - Sidewalk}&\multicolumn{1}{|c}{Road - Sidewalk - Ground - Stairs}\\
				\thickhline
				
				{Vertical Plane}&\multicolumn{1}{|c}{Building - Vehicle}&\multicolumn{1}{|c}{Building - Small Vehicle - Big Vehicle}\\
				\thickhline
				
				{Cylindrical}&\multicolumn{1}{|c}{Tree Trunk - Pole- Sign - Post-Barrier}&\multicolumn{1}{|c}{Barrier - Bus Stop - Tree Trunk- Tall Light - Post - Sign - Utility Pole- Traffic Signal}\\
				\thickhline
				
				{Scattered}&\multicolumn{1}{|c}{Tree Leaves - Bush}&\multicolumn{1}{|c}{Shrub - Tree Top}\\
				\thickhline
				
				{Sky}&\multicolumn{1}{|c}{Sky}&\multicolumn{1}{|c}{---}\\
				\thickhline
				
				{Person}&\multicolumn{1}{|c}{---}&\multicolumn{1}{|c}{Person}\\
				\thickhline

				{Wire}&\multicolumn{1}{|c}{Wire}&\multicolumn{1}{|c}{Wire}\\
				\thickhline
				
			\end{tabular}
		}
		
		\label{Semantic_geometric}
	\end{table*}
	\begin{table*}[h]\centering\renewcommand{\arraystretch}{1.15}
		\caption{Per class F1-scores for the 2D domain in the DATA61/2D3D dataset. We present the results for unary, pairwise model learned on the 2D domain only,  the method of~\cite{wacv2015} with handcrafted potentials, the 2D-3D learned potentials, the 2D-3D learned potentials with latent nodes, semantic results with semantic - geometric model with and without latent nodes.}
		\vspace{-20pt}
		\def\d{.1cm}
		\def\tt{1cm}
		\def\r{30}
		\small\addtolength{\tabcolsep}{3pt}
		{
			\resizebox{\textwidth}{!}{ 
				\begin{tabular}{l+p{\d}+p{\d}+p{\d}+p{\d}+p{\d}+p{\d}+p{\d}+p{\d}+p{\d}+p{\d}+p{\d}+p{\d}+p{\d}+p{\d}ccc}\\ 
					{}&\rotatebox{\r}{\bf \scriptsize Grass}&\rotatebox{\r}{\bf \scriptsize Building}&\rotatebox{\r}{\bf \scriptsize Tree trunk}&\rotatebox{\r}{\bf \scriptsize Tree leaves}&\rotatebox{\r}{\bf \scriptsize Vehicle}&\rotatebox{\r}{\bf \scriptsize Road}&\rotatebox{\r}{\bf \scriptsize Bush}&\rotatebox{\r}{\bf \scriptsize Pole}&\rotatebox{\r}{\bf \scriptsize Sign}&\rotatebox{\r}{\bf \scriptsize Post}&\rotatebox{\r}{\bf \scriptsize Barrier}&\rotatebox{\r}{\bf \scriptsize Wire}&\rotatebox{\r}{\bf \scriptsize Sidewalk}&\rotatebox{\r}{\bf \scriptsize Sky}&{\bf \scriptsize avg}\\ \thickhline
					
					{Unary}&{80}&33&14&80&49&{95}&16&28&3&0&0&29&15&{98}&\multicolumn{1}{|c}{38}\\
					\thickhline
					
					{Pairwise 2D (learned)}&85&57&17&85&55&95&18&30&0&0&3&{34}&{20}&{\bf99}&\multicolumn{1}{|c}{43}\\
					\thickhline
					
					{2D-3D handcrafted potentials, Namin \cite{wacv2015}}&74&{56}&{21}&{82}&{58}&92&{23}&{33}&{19}&{8}&{5}&{32}&29&{97}&\multicolumn{1}{|c}{45}\\
					\thickhline
					
					{2D-3D learned potentials (no feature)}&{94}&{58}&{12}&{83}&{72}&{64}&{31}&{34}&{6}&{0}&{13}&{37}&{48}&{97}&\multicolumn{1}{|c}{46}\\
					\thickhline
					
					{2D-3D learned potentials (full features)}&{90}&{63}&{10}&{91}&{68}&{96}&{31}&{43}&{1}&{0}&{0}&{44}&{53}&{\bf99}&\multicolumn{1}{|c}{49}\\
					\thickhline
					
					{2D-3D learned potentials (selected features)}&{92}&{64}&{18}&{92}&{69}&{98}&{36}&{34}&{3}&{0}&{\bf28}&{40}&{60}&{\bf99}&\multicolumn{1}{|c}{52}\\
					\thickhline
					
					{2D-3D learned potentials with latent nodes}&{\bf95}&{71}&{28}&{93}&{76}&{97}&{44}&{44}&{10}&{5}&{21}&{38}&{68}&{\bf99}&\multicolumn{1}{|c}{56}\\
					\thickhline
					
					{Semantic results with semantic - geometric model }&{92}&{70}&{26}&{93}&{72}&{97}&{32}&{49}&{17}&{0}&{0}&{\bf63}&{65}&{\bf99}&\multicolumn{1}{|c}{55}\\
					
					{(selected features) }&{}&{}&{}&{}&{}&{}&{}&{}&{}&{}&{}&{}&{}&{}&\multicolumn{1}{|c}{}\\
					\thickhline
					
					{Semantic results with semantic - geometric model }&{93}&{79}&{45}&{\bf95}&{77}&{98}&{34}&{55}&{22}&{0}&{0}&{\bf63}&{83}&{\bf99}&\multicolumn{1}{|c}{60}\\
					
					{and latent nodes}&{}&{}&{}&{}&{}&{}&{}&{}&{}&{}&{}&{}&{}&{}&\multicolumn{1}{|c}{}\\
					\thickhline
					
					{Semantic results with semantic - geometric model }&{\bf95}&{\bf82}&{\bf52}&{90}&{\bf78}&{\bf99}&{\bf78}&{\bf99}&{\bf33}&{\bf60}&{20}&{61}&{\bf92}&{\bf99}&\multicolumn{1}{|c}{\bf62}\\
					
					{ (Connected 2D frames) }&{}&{}&{}&{}&{}&{}&{}&{}&{}&{}&{}&{}&{}&{}&\multicolumn{1}{|c}{}\\
					\thickhline
					
				\end{tabular}
			}
		}
		
		\label{New_2d_table}
	\end{table*}
	\begin{table*}[h]\centering\renewcommand{\arraystretch}{1.15}
		\caption{Per class F1-scores for the 3D domain in the DATA61/2D3D dataset. We present the results for unary, pairwise model learned on the 2D domain only,  the method of~\cite{wacv2015} with handcrafted potentials, the 2D-3D learned potentials, the 2D-3D learned potentials with latent nodes, semantic results with semantic - geometric model with and without latent nodes.}
		\vspace{-20pt}
		\def\d{.1cm}
		\def\r{30}
		\small\addtolength{\tabcolsep}{2.8pt}
		{
			\resizebox{\textwidth}{!}{  
				\begin{tabular}{l+p{\d}+p{\d}+p{\d}+p{\d}+p{\d}+p{\d}+p{\d}+p{\d}+p{\d}+p{\d}+p{\d}+p{\d}+p{\d}+p{\d}ccc}\\ 
					{}&\rotatebox{\r}{\bf \scriptsize Grass}&\rotatebox{\r}{\bf \scriptsize Building}&\rotatebox{\r}{\bf \scriptsize Tree trunk}&\rotatebox{\r}{\bf \scriptsize Tree leaves}&\rotatebox{\r}{\bf \scriptsize Vehicle}&\rotatebox{\r}{\bf \scriptsize Road}&\rotatebox{\r}{\bf \scriptsize Bush}&\rotatebox{\r}{\bf \scriptsize Pole}&\rotatebox{\r}{\bf \scriptsize Sign}&\rotatebox{\r}{\bf \scriptsize Post}&\rotatebox{\r}{\bf \scriptsize Barrier}&\rotatebox{\r}{\bf \scriptsize Wire}&\rotatebox{\r}{\bf \scriptsize Sidewalk}&\rotatebox{\r}{\bf \scriptsize Sky}&{\bf \scriptsize avg}\\ \thickhline
					
					{Unary}&52&61&27&87&58&{82}&10&24&19&43&19&74&0&\#&\multicolumn{1}{|c}{43}\\
					\thickhline
					
					{Pairwise 3D (learned)}&58&{80}&{50}&{97}&56&76&16&{\bf62}&{32}&40&0&{89}&0&\#&\multicolumn{1}{|c}{50}\\
					\thickhline
					
					{2D-3D handcrafted potentials, Namin \cite{wacv2015}}&{63}&{81}&{41}&{96}&{70}&76&{21}&38&28&{47}&{23}&{87}&0&\#&\multicolumn{1}{|c}{52}\\
					\thickhline
					
					{2D-3D learned potentials (no feature)}&{68}&{81}&{31}&{92}&{67}&{83}&{\bf69}&{43}&{37}&{25}&{16}&{75}&10&\#&\multicolumn{1}{|c}{54}\\
					\thickhline
					
					{2D-3D learned potentials (full features)}&{72}&{75}&{27}&{95}&{77}&{90}&{42}&{\bf62}&{31}&{9}&{0}&{89}&0&\#&\multicolumn{1}{|c}{52}\\
					\thickhline
					
					{2D-3D learned potentials (selected features)}&{60}&{92}&{45}&{97}&{75}&79&{61}&{58}&{49}&{29}&{\bf27}&{82}&0&\#&\multicolumn{1}{|c}{58}\\
					\thickhline
					
					{2D-3D learned potentials with latent nodes}&{66}&{\bf94}&{49}&{95}&{\bf79}&{83}&{51}&{\bf62}&{\bf54}&{43}&{25}&{89}&{8}&\#&\multicolumn{1}{|c}{61}\\
					\thickhline
					
					{Semantic results with semantic - geometric model }&{71}&{88}&{51}&{97}&{76}&{84}&{56}&{60}&{51}&{49}&{6}&{92}&{21}&\#&\multicolumn{1}{|c}{62}\\
					
					{(selected features) }&{}&{}&{}&{}&{}&{}&{}&{}&{}&{}&{}&{}&{}&{}&\multicolumn{1}{|c}{}\\
					\thickhline
					
					{Semantic results with semantic - geometric model }&{79}&{91}&{64}&{\bf99}&{77}&{\bf93}&{60}&{61}&{50}&{58}&{0}&{\bf96}&{\bf34}&\#&\multicolumn{1}{|c}{\bf66}\\
					
					{and latent nodes }&{}&{}&{}&{}&{}&{}&{}&{}&{}&{}&{}&{}&{}&{}&\multicolumn{1}{|c}{}\\
					\thickhline
					
					{Semantic results with semantic - geometric model }&{\bf80}&{92}&{\bf65}&{98}&{75}&{\bf93}&{65}&{59}&{49}&{\bf62}&{0}&{93}&{32}&\#&\multicolumn{1}{|c}{\bf66}\\
					
					{ (Connected 2D frames) }&{}&{}&{}&{}&{}&{}&{}&{}&{}&{}&{}&{}&{}&{}&\multicolumn{1}{|c}{}\\
					\thickhline
					
				\end{tabular}
			}
		}
		
		\label{New_3d_table}
	\end{table*}
	\begin{table*}[h]\centering\renewcommand{\arraystretch}{1.2}
		\caption{Per class F1-scores for geometric results with semantic - geometric model and latent nodes in the DATA61/2D3D dataset.}
		\vspace{-20pt}
		
		\def\r{30}
		\small\addtolength{\tabcolsep}{2.8pt}
		{
			\resizebox{\textwidth}{!}{  
				\begin{tabular}{l+c+c+c+c+c+cccc}\\ 
					{}&{\bf \scriptsize Horizontal plane}&{\bf \scriptsize Vertical plane}&{\bf \scriptsize Cylindrical}&{\bf \scriptsize Scattered}&{\bf \scriptsize Wire}&{\bf \scriptsize Sky}&{\bf \scriptsize avg}\\ \thickhline
					
					{2D geometric results with semantic - geometric model}&{98}&76&25&94&43&{99}&\multicolumn{1}{|c}{72}\\
					\thickhline
					
					{3D geometric results with semantic - geometric model}&99&91&62&99&95&\#&\multicolumn{1}{|c}{89}\\
					\thickhline
					
				\end{tabular}
			}
		}
		\label{DATA61_geometry_table}
	\end{table*}
	
	\begin{table*}[t]\centering\renewcommand{\arraystretch}{1.2}
		\caption{Per class F1-scores for the 2D domain in the CMU/VMR dataset. We present the results for unary, pairwise model learned on the 2D domain only,  the method of~\cite{Munoz:2012:eccv}, the method of~\cite{wacv2015} with handcrafted potentials, the 2D-3D learned potentials, the 2D-3D learned potentials with latent nodes, semantic results with semantic - geometric model with and without latent nodes.}
		\vspace{-20pt}
		\def\d{.1cm}
		\def\r{30}
		\small\addtolength{\tabcolsep}{2.8pt}
		{
			\resizebox{\textwidth}{!}{  
				\begin{tabular}{l+p{\d}+p{\d}+p{\d}+p{\d}+p{\d}+p{\d}+p{\d}+p{\d}+p{\d}+p{\d}+p{\d}+p{\d}+p{\d}+p{\d}+p{\d}+p{\d}+p{\d}+p{\d}+p{\d}ccc}\\ 
					{}&\rotatebox{\r}{\bf \scriptsize Road}&\rotatebox{\r}{\bf \scriptsize Sidewalk}&\rotatebox{\r}{\bf \scriptsize Ground}&\rotatebox{\r}{\bf \scriptsize Building}&\rotatebox{\r}{\bf \scriptsize Barrier}&\rotatebox{\r}{\bf \scriptsize Bus stop}&\rotatebox{\r}{\bf \scriptsize Stairs}&\rotatebox{\r}{\bf \scriptsize Shrub}&\rotatebox{\r}{\bf \scriptsize Tree trunk}&\rotatebox{\r}{\bf \scriptsize Tree top}&\rotatebox{\r}{\bf \scriptsize Small Vehicle}&\rotatebox{\r}{\bf \scriptsize Big vehicle}&\rotatebox{\r}{\bf \scriptsize Person}&\rotatebox{\r}{\bf \scriptsize Tall light}&\rotatebox{\r}{\bf \scriptsize Post}&\rotatebox{\r}{\bf \scriptsize Sign}&\rotatebox{\r}{\bf \scriptsize Utility pole}&\rotatebox{\r}{\bf \scriptsize Wire}&\rotatebox{\r}{\bf \scriptsize Traffic Signal}&{\bf \scriptsize avg}\\ \thickhline
					
					{Unary}&95&81&75&56&29&17&32&50&31&53&32&49&29&{\bf16}&{\bf15}&{\bf 16}&33&{\bf41}&29&\multicolumn{1}{|c}{41}\\
					\thickhline
					
					{Pairwise 2D (learned)}&89&77&74&84&25&17&40&62&{37}&89&78&{57}&38&{1}&{5}&3&16&{12}&9&\multicolumn{1}{|c}{43}\\
					\thickhline
					
					{Munoz \cite{Munoz:2012:eccv}}&{\bf96}&{\bf90}&70&{83}&{50}&16&33&{62}&30&{86}&{\bf84}&50&{47}&2&9&{\bf 16}&14&2&17&\multicolumn{1}{|c}{45}\\
					\thickhline
					
					{2D-3D handcrafted potentials, Namin \cite{wacv2015}}&94&87&{79}&74&45&{22}&{40}&54&27&84&67&24&38&13&2&10&{37}&35&{\bf40}&\multicolumn{1}{|c}{46}\\
					\thickhline
					
					{2D-3D learned potentials (no feature)}&95&84&{78}&{70}&{58}&{18}&{57}&{68}&{43}&{84}&{81}&{52}&{55}&9&3&2&{15}&5&{8}&\multicolumn{1}{|c}{47}\\
					\thickhline
					
					{2D-3D learned potentials (full features)}&93&85&{83}&{\bf88}&{60}&{4}&{61}&{\bf67}&{41}&{87}&{79}&{61}&{45}&0&3&2&{12}&9&{2}&\multicolumn{1}{|c}{46}\\
					\thickhline
					
					{2D-3D learned potentials (selected features)}&93&80&{80}&{87}&{60}&{1}&{70}&{\bf67}&{37}&{\bf90}&{\bf84}&{\bf67}&{54}&7&4&4&{21}&15&{3}&\multicolumn{1}{|c}{49}\\
					\thickhline
					
					{2D-3D learned potentials with latent nodes}&94&84&{\bf84}&{84}&{\bf65}&{4}&{\bf75}&{64}&{43}&{89}&{\bf84}&{58}&{52}&11&6&2&{25}&18&{3}&\multicolumn{1}{|c}{\bf50}\\
					\thickhline
					
					{Semantic results with semantic - geometric model}&94&87&{82}&{82}&{61}&{26}&{59}&{68}&{43}&{89}&{74}&{60}&{55}&0&4&4&{27}&15&{8}&\multicolumn{1}{|c}{49}\\
					
					{(selected features)}&{}&{}&{}&{}&{}&{}&{}&{}&{}&{}&{}&{}&{}&{}&{}&{}&{}&{}&{}&\multicolumn{1}{|c}{}\\
					\thickhline
					
					{Semantic results with semantic - geometric model}&94&87&{\bf84}&{81}&{58}&{\bf28}&{63}&{66}&{\bf47}&{87}&{78}&{64}&{\bf56}&0&6&5&{\bf38}&17&{10}&\multicolumn{1}{|c}{\bf51}\\
					
					{and latent nodes}&{}&{}&{}&{}&{}&{}&{}&{}&{}&{}&{}&{}&{}&{}&{}&{}&{}&{}&{}&\multicolumn{1}{|c}{}\\
					\thickhline
					
				\end{tabular}
			}
		}
		
		\label{Munoz_2d_table}
	\end{table*}
	
	\begin{table*}[t]\centering\renewcommand{\arraystretch}{1.2}
		\caption{Per class F1-scores for the 3D domain in the CMU/VMR dataset. We present the results for unary, pairwise model learned on the 2D domain only,  the method of~\cite{Munoz:2012:eccv}, the method of~\cite{wacv2015} with handcrafted potentials, the 2D-3D learned potentials, the 2D-3D learned potentials with latent nodes, semantic results with semantic - geometric model with and without latent nodes.} 
		\vspace{-20pt}
		\def\d{.1cm}
		\def\r{30}
		\small\addtolength{\tabcolsep}{2.8pt}
		{
			\resizebox{\textwidth}{!}{  
				\begin{tabular}{l+p{\d}+p{\d}+p{\d}+p{\d}+p{\d}+p{\d}+p{\d}+p{\d}+p{\d}+p{\d}+p{\d}+p{\d}+p{\d}+p{\d}+p{\d}+p{\d}+p{\d}+p{\d}+p{\d}ccc}\\ 
					{}&\rotatebox{\r}{\bf \scriptsize Road}&\rotatebox{\r}{\bf \scriptsize Sidewalk}&\rotatebox{\r}{\bf \scriptsize Ground}&\rotatebox{\r}{\bf \scriptsize Building}&\rotatebox{\r}{\bf \scriptsize Barrier}&\rotatebox{\r}{\bf \scriptsize Bus stop}&\rotatebox{\r}{\bf \scriptsize Stairs}&\rotatebox{\r}{\bf \scriptsize Shrub}&\rotatebox{\r}{\bf \scriptsize Tree trunk}&\rotatebox{\r}{\bf \scriptsize Tree top}&\rotatebox{\r}{\bf \scriptsize Small Vehicle}&\rotatebox{\r}{\bf \scriptsize Big vehicle}&\rotatebox{\r}{\bf \scriptsize Person}&\rotatebox{\r}{\bf \scriptsize Tall light}&\rotatebox{\r}{\bf \scriptsize Post}&\rotatebox{\r}{\bf \scriptsize Sign}&\rotatebox{\r}{\bf \scriptsize Utility pole}&\rotatebox{\r}{\bf \scriptsize Wire}&\rotatebox{\r}{\bf \scriptsize Traffic Signal}&{\bf \scriptsize avg}\\ \thickhline
					
					{Unary}&70&49&62&67&34&2&19&26&11&67&34&4&13&2&0&1&2&0&0&\multicolumn{1}{|c}{24}\\
					\thickhline
					
					{Pairwise 3D (learned)}&78&52&67&78&15&1&32&31&1&73&44&14&9&1&0&0&0&0&0&\multicolumn{1}{|c}{26}\\
					\thickhline
					
					{Munoz \cite{Munoz:2012:eccv}}&82&73&68&{87}&46&11&38&{63}&28&{\bf88}&{73}&{\bf56}&26&{\bf10}&0&0&0&0&0&\multicolumn{1}{|c}{39}\\
					\thickhline
					
					{2D-3D handcrafted potentials, Namin \cite{wacv2015}}&{92}&{85}&{81}&85&{50}&{16}&{42}&55&{29}&82&70&16&{43}&6&{2}&{\bf7}&{\bf29}&{9}&{\bf23}&\multicolumn{1}{|c}{43}\\
					\thickhline
					
					{2D-3D learned potentials (no feature)}&{92}&{84}&{85}&{87}&{64}&{3}&{59}&{64}&{32}&77&{70}&19&{42}&5&{2}&{3}&{7}&{3}&{9}&\multicolumn{1}{|c}{42}\\
					\thickhline
					
					{2D-3D learned potentials (full features)}&{90}&{86}&{\bf87}&{90}&{59}&{2}&{64}&{69}&{31}&79&{70}&29&{47}&1&{1}&{0}&{5}&{0}&{0}&\multicolumn{1}{|c}{43}\\
					\thickhline
					
					{2D-3D learned potentials (selected features)}&{90}&{85}&{85}&{89}&{62}&{2}&{63}&{68}&{29}&86&{78}&46&{53}&3&{1}&{0}&{15}&{0}&{0}&\multicolumn{1}{|c}{45}\\
					\thickhline
					
					{2D-3D learned potentials with latent nodes}&{92}&{\bf88}&{84}&{88}&{64}&{7}&{\bf66}&{66}&{31}&86&{75}&42&{53}&8&{\bf7}&{0}&{17}&{10}&{0}&\multicolumn{1}{|c}{47}\\
					\thickhline
					
					{Semantic results with semantic - geometric model}&{93}&86&{85}&{\bf92}&{66}&{12}&{62}&{68}&{39}&{86}&{\bf80}&{47}&{56}&0&2&2&{21}&{10}&{0}&\multicolumn{1}{|c}{48}\\
					
					{(selected features)}&{}&{}&{}&{}&{}&{}&{}&{}&{}&{}&{}&{}&{}&{}&{}&{}&{}&{}&{}&\multicolumn{1}{|c}{}\\
					\thickhline
					
					{Semantic results with semantic - geometric model}&{\bf94}&86&{\bf87}&{90}&{\bf71}&{\bf18}&{60}&{\bf70}&{\bf44}&{87}&{78}&{43}&{\bf58}&0&2&2&{28}&{\bf13}&{0}&\multicolumn{1}{|c}{\bf50}\\
					
					{and latent nodes}&{}&{}&{}&{}&{}&{}&{}&{}&{}&{}&{}&{}&{}&{}&{}&{}&{}&{}&{}&\multicolumn{1}{|c}{}\\
					\thickhline

				\end{tabular}
			}
		}
		
		\label{Munoz_3d_table}
	\end{table*}
	\begin{table*}[h]\centering\renewcommand{\arraystretch}{1.2}
		\caption{Per class F1-scores for geometric results with semantic-geometric model and latent nodes in the CMU/VMR dataset.}
		\vspace{-20pt}
		
		\def\r{30}
		\small\addtolength{\tabcolsep}{3pt}
		{
			\resizebox{\textwidth}{!}{ 
				\begin{tabular}{l+c+c+c+c+c+cccc}\\ 
					{}&{\bf \scriptsize Horizontal plane}&{\bf \scriptsize Vertical plane}&{\bf \scriptsize Cylindrical}&{\bf \scriptsize Scattered}&{\bf \scriptsize Person}&{\bf \scriptsize Wire}&{\bf \scriptsize avg}\\ \thickhline
					
					{2D geometric results with semantic-geometric model}&{97}&85&44&88&56&{52}&\multicolumn{1}{|c}{70}\\
					\thickhline
					
					{3D geometric results with semantic-geometric model}&96&91&60&87&56&19&\multicolumn{1}{|c}{68}\\
					\thickhline
					
				\end{tabular}
			}
		}
		\label{DATA61_geometry_table1}
	\end{table*}
	
	

\subsection{Scalability}
In this section we discuss the scalability of our model.
In the proposed model with multiple modalities, each node has correspondence with only a few nodes in other modalities and hence, the number of latent nodes in the graph grows linearly with the total number of nodes in all modalities.

Augmenting our model with latent nodes introduces new potential functions between latent nodes and 2D/3D nodes, with parameter matrices of size $L \times (L+1)$. However, as explained in Sec.~\ref{sec:method2}, only $2L$ parameters are required to be trained for each potential function.

In order to show the scalability of the proposed method and since no dataset with multiple (more than two) visual modalities were publicly available, we considered geometric class labels to represent another form of visual modality. Note that using a real visual sensor as another modality might impose some challenges on finding node correspondences between modalities. Nonetheless, in terms of computational complexity, the problem is not different from our case where geometric classes are taken as a domain.

The training and inference times reported in Table~\ref{time1} demonstrate that even though the training time prolongs to some extent as the number of modalities grows, the inference times remain quite short.
	
\section{Conclusion}
	In this paper, we have presented a general multimodal model that could simultaneously accommodate multiple modalities.
	We have also addressed the problem of domain inconsistencies in multimodal semantic labeling, which is an important issue when multimodal data is concerned.
	Such inconsistencies typically cause undesirable connections between two modalities, which in turn lead to poor labeling performance. We have, therefore, proposed a latent CRF model, in which latent nodes supervise the pairwise edges between each two domains. Having access to the information of both modalities, these nodes can either improve the labeling in both domains or cut the links between inconsistent regions. Furthermore, we presented a new set of data-driven learned potentials, which can model complex relationships between the latent nodes and the modalities.
	In addition, our general model enables us to jointly consider the geometric and semantic classes for both 2D and 3D data and perform a concurrent inference on them to further improve the 2D and 3D semantic labeling results. Thanks to our general model, latent nodes and our learned potentials, our model achieved state-of-the-art results on two publicly available datasets.

	\section*{Acknowledgment}
	
	The authors would like to thank Justin Domke for his assistance in implementing the learned potentials.

	\ifCLASSOPTIONcaptionsoff
	\newpage
	\fi

	
	
	\bibliographystyle{IEEEtran}
	\bibliography{IEEEabrv,egbib}
	%

	%
	\vspace{-37pt}
	\begin{IEEEbiographynophoto}{Sarah Taghavi Namin}
		is a postdoctoral computer vision researcher at the Australian National University (ANU). She obtained her PhD in 2016 from ANU and NICTA(Data61) computer vision research group. She received her BSc degree in Electrical engineering from the University of Tehran in 2007 and her MSc degree in electrical engineering from K.N. Toosi University of Technology in 2010. 
	\end{IEEEbiographynophoto}
	\vspace{-37pt}
	\begin{IEEEbiographynophoto}{Mohammad Najafi}
		is a postdoctoral computer vision researcher at the University of Oxford. He obtained his PhD in 2016 from the Australian National University (ANU) and NICTA(Data61) computer vision research group. He received his BSc and MSc degrees in electrical engineering in 2007 and 2010 from the University of Tehran. 
	\end{IEEEbiographynophoto}
	
	\vspace{-37pt}
	\begin{IEEEbiographynophoto}{Mathieu Salzmann}
		is a Senior Researcher at EPFL-CVLab. Previously, he was a Senior Researcher and Research Leader in NICTA€™'s computer vision research group. Prior to this, from Sept. 2010 to Jan. 2012, he was a Research Assistant Professor at TTI-Chicago, and, from Feb. 2009 to Aug. 2010, a postdoctoral fellow at ICSI and EECS at UC Berkeley under the supervision of Prof. Trevor Darrell. He obtained his PhD in Jan. 2009 from EPFL under the supervision of Prof. Pascal Fua. His research interests lie at the intersection of machine learning and geometry for computer vision.
	\end{IEEEbiographynophoto}
	\vspace{-37pt}
	\begin{IEEEbiographynophoto}{Lars Petersson}
		is a Principal Research Scientist within the Smart Vision System's Group, Data61, CSIRO, Australia. There, he is leading a team specialising in resource constrained computer vision. Previously, he was a Principal Researcher and Research Leader in NICTA's computer vision research group where, from 2003 until 2016, he was leading projects such as Smart Cars, AutoMap, and Distributed Large Scale Vision. Before joining NICTA, he did one year of postdoctoral research at the Australian National University working with Dr Alexander Zelinsky.  He received his PhD in March 2002 from KTH, Stockholm, Sweden, where he also received his Master's degree in Engineering Physics.
	\end{IEEEbiographynophoto}
	
	
	

\end{document}